\documentclass[acmsmall,screen,nonacm]{acmart}
\usepackage{multirow}
\usepackage{amsmath,amsfonts}
\usepackage{adjustbox}
\usepackage{colortbl}

\definecolor{Gray}{gray}{0.9}
\definecolor{mintbg}{rgb}{.63,.79,.95}
\definecolor{darkbrown}{RGB}{180, 50, 33}
\colorlet{lightmintbg}{mintbg!20}
\definecolor{darkgreen}{rgb}{0.0, 0.5, 0.0}
\definecolor{lightgreen}{rgb}{0.6, 0.95, 0.7}
\usepackage{tikz}
\usepackage{textcomp}
\usepackage{hyphenat}
\newcommand{\tablistcommand}{\leavevmode\par\vspace{-\baselineskip}}
\usepackage{enumitem}
\usepackage{makecell,multirow}
\usepackage{pbox}
\usepackage{algorithm}
\usepackage{algpseudocode} 
\usepackage{algorithmicx}
\usepackage{xcolor} 
\usepackage{longtable}
\usepackage{tabularray}
\usepackage{printlen}
\usepackage[utf8]{inputenc}
\usepackage[T1]{fontenc}
\usepackage{ltablex}
\usepackage{caption, booktabs}
\usepackage{afterpage}
\usepackage{pifont}

\begin{document}

\title{Beyond Few-shot Object Detection: A Detailed Survey}
\author{Vishal Chudasama}
\authornotemark[1]
\email{vishal.chudasama1@sony.com}
\orcid{0000-0002-3727-5484}
\affiliation{%
  \institution{Sony Research India}
  \city{Bangalore}
  \state{Karnataka}
  \country{India}
}
\author{Hiran Sarkar}
\authornote{Both authors contributed equally to this research.}
\affiliation{%
  \institution{Sony Research India}
  \city{Bangalore}
  \state{Karnataka}
  \country{India}
}
\email{hiran.sarkar@sony.org}
\orcid{0000-0002-9232-8198}

\author{Pankaj Wasnik}
\authornote{Corresponding author}
\affiliation{%
  \institution{Sony Research India}
  \city{Bangalore}
  \state{Karnataka}
  \country{India}
}
\email{pankaj.wasnik@sony.org}
\orcid{0000-0001-5602-2901}

\author{Vineeth N Balasubramanian}
\affiliation{%
  \institution{Indian Institute of Technology}
  \city{Hyderabad}
  \country{India}
}
\email{vineethnb@cse.iith.ac.in}
\orcid{0000-0003-2656-0375}

\author{Jayateja Kalla}
\authornote{Contributed during his internship at Sony Research India}
\affiliation{%
  \institution{Indian Institute of Science}
  \city{Bangalore}
  \country{India}
}
\email{jayatejak@iisc.ac.in}
\orcid{0000-0002-0093-3607}

\renewcommand{\shortauthors}{Vishal \textit{et al.}}





\maketitle

\section*{Abstract}
Object detection is a critical field in computer vision focusing on accurately identifying and locating specific objects in images or videos. Traditional methods for object detection rely on large labeled training datasets for each object category, which can be time-consuming and expensive to collect and annotate. To address this issue, researchers have introduced few-shot object detection (FSOD) approaches that merge few-shot learning and object detection principles. These approaches allow models to quickly adapt to new object categories with only a few annotated samples. While traditional FSOD methods have been studied before, this survey paper comprehensively reviews FSOD research with a specific focus on covering different FSOD settings such as standard FSOD, generalized FSOD, incremental FSOD, open-set FSOD, and domain adaptive FSOD. These approaches play a vital role in reducing the reliance on extensive labeled datasets, particularly as the need for efficient machine learning models continues to rise. 
  This survey paper aims to provide a comprehensive understanding of the above-mentioned few-shot settings and explore the methodologies for each FSOD task. It thoroughly compares state-of-the-art methods across different FSOD settings, analyzing them in detail based on their evaluation protocols. Additionally, it offers insights into their applications, challenges, and potential future directions in the evolving field of object detection with limited data.  
\section{Introduction}
\label{sec:introduction}
Object detection has experienced remarkable advancements in recent years due to significant progress in deep learning techniques, as demonstrated by methods like Faster R-CNN~\cite{faster_rcnn}, YOLO~\cite{yolo_v2}, and DETR~\cite{detr}. The primary goal of object detection is to accurately identify and locate objects within an image while also categorizing these objects into specific predefined classes. However, traditional deep learning approaches for object detection heavily depend on large-scale labeled training datasets~\cite{imagenet_dataset}. This dependence poses significant challenges in real-world scenarios where collecting large amounts of data is often impractical~\cite{few_shot_lrn}. Acquiring a sufficient number of images can be infeasible, and annotating these images for object detection is both expensive and time-consuming. Additionally, training complex deep learning models with limited data frequently results in overfitting issues, where the model performs well on the training data but fails to generalize to unseen data.
\begin{figure}[t!]
    \centering
    \includegraphics[width=\linewidth]{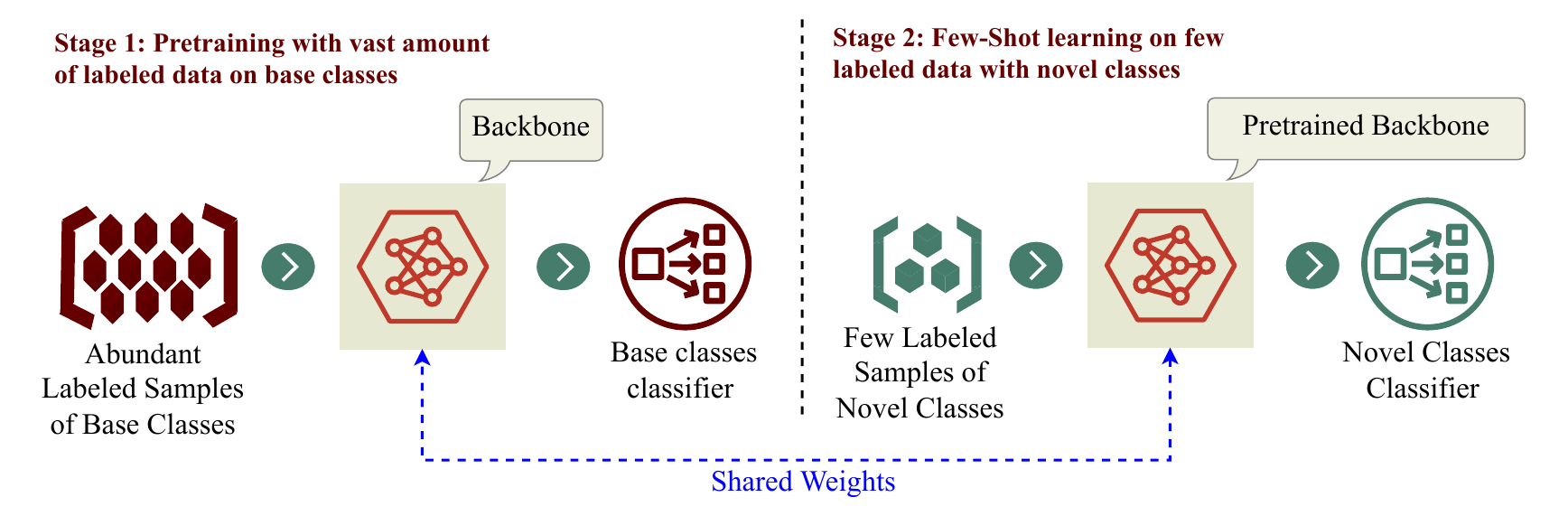}  
    \vspace{-2em}
    \caption{Illustration of traditional few-shot learning setting: pre-trained model adapts to new classes with minimal data samples.}
    \vspace{-1em}
    \label{intro_figure}
\end{figure}
In contrast, humans possess an exceptional ability to learn new concepts with minimal data, especially during the early stages of development. For example, children can quickly identify and differentiate new objects after encountering them only a few times. Inspired by this impressive human capacity, a setting called few-shot learning~\cite{few_shot_lrn, few_shot_lrn_survey} has emerged, where the model is trained to learn from a few number of samples. 

Fig.~\ref{intro_figure} illustrates the standard few-shot learning regime, where a model is initially trained on a large amount of labeled data and adapts to new classes with a significantly smaller number of samples. In the field of few-shot object detection (FSOD), the objective is to detect certain objects using only a limited number of annotated instances, thus eliminating the need for extensive annotated data, which is a primary limitation of state-of-the-art object detection approaches. The first stage in FSOD is to pre-train the model using a large amount of data from known classes, called \textit{base classes}. In the second stage, this knowledge enables the model to recognize new classes, termed \textit{novel classes}, with only a few examples. 

FSOD is important in various real-world applications where obtaining a large amount of annotated data is challenging, expensive, or time-consuming. In medical imaging \cite{fsod_medical_imaging}, FSOD methods can help identify rare new diseases from a limited number of labeled examples, enabling quicker diagnosis and treatment. In wildlife conservation, FSOD methods can help monitor endangered species with minimal data, supporting conservation efforts. FSOD is also valuable in industrial inspection \cite{defrcn_mam, xray_prohibited}, where defects or anomalies in manufacturing processes can be detected with limited training samples, enhancing quality control. In security and surveillance applications, FSOD can detect suspicious activities or objects with minimal labeled data, improving safety and response times. FSOD methods also have a vital role to play in other domains such as remote sensing or multispectral imaging \cite{retentive_com_remote_sensing, fsod_multispectral},  as well as other settings such as cross-domain generalization \cite{asyfod, cutmix}, further broadening its applicability and impact across diverse fields. Given the increased significance, newer variants of the traditional FSOD setting have emerged in recent years. Exploring this expansion of the FSOD setting in recent literature is the key objective of this survey. 

\begin{figure}[t!]
    \centering
    \includegraphics[width=\linewidth]{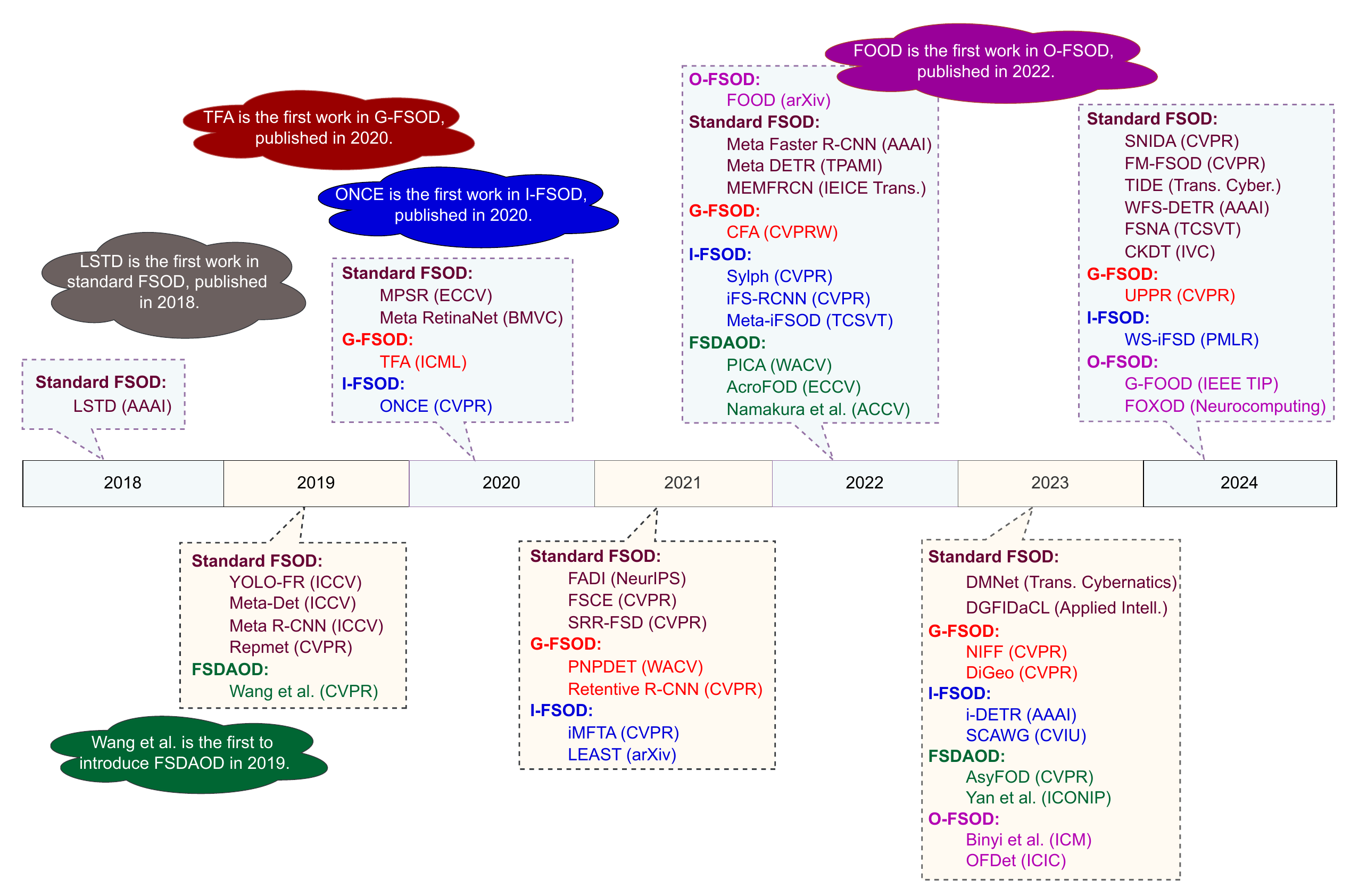}  
    \caption{Timeline of FSOD efforts: (i) Standard FSOD works are highlighted in {\color{darkbrown}{brown color}}; (ii) Generalized FSOD (G-FSOD) works are highlighted in \textcolor{red} {red color}; (iii) Incremental FSOD (I-FSOD) works are highlighted in \textcolor{blue} {blue color}; (iv) Open-set FSOD (O-FSOD) works are highlighted in \textcolor{magenta} {magenta color}; and (v) Domain adaptation FSOD (FSDAOD) works are highlighted in {\color{darkgreen}{green color}}.}
    \label{intro_timeline}
\end{figure}
Fig.~\ref{intro_timeline} outlines the evolution of the few-shot object detection task in recent years. While the initial efforts focused on standard FSOD, extensions and variants have emerged as listed below:
\begin{itemize}
    \item Standard FSOD
    \item Generalized FSOD (G-FSOD) 
    \item Incremental FSOD (I-FSOD)
    \item Open-set FSOD (O-FSOD)
    \item Domain Adaptation based FSOD (FSDAOD)
\end{itemize}

Standard FSOD eliminates the dependency on vast amounts of labeled training data, and its primary focus lies in enhancing performance in novel classes rather than maintaining performance in base classes. Chen \textit{et al.} provide the first paper in this direction in their work LSTD \cite{lstd}. Nevertheless, learning new classes while maintaining performance in base classes is often crucial in real-world applications. To address this challenge, two other tasks, generalized few-shot object detection (G-FSOD) and incremental few-shot object detection (I-FSOD), aim to perform strongly on both base and novel classes. G-FSOD tackles the challenge of proper knowledge retention of the base classes while learning the new classes. TFA \cite{tfa} is the first work in the task of G-FSOD that provides results on both the base and novel classes. Both standard FSOD and G-FSOD rely on the availability of the base classes while learning the new classes. However, it is not feasible in the real-world scenario to avail the old classes while learning the new classes. ONCE \cite{once} is the first to address this issue and introduce the task of I-FSOD. I-FSOD, unlike the previous tasks, does not require the old classes to adapt to new ones.

Open-set few-shot object detection (O-FSOD) is another FSOD sub-category that focuses on not only detecting objects of trained classes but also detecting objects of unseen classes. In many real-world scenarios, it is impractical to pre-define or pre-label all possible object categories, making it essential for systems to recognize and handle new or rare objects that weren't part of the initial training set. FOOD \cite{food} is the first work proposed in this direction by Su \textit{et al.}. Another sub-category of FSOD, called Few-shot domain adaptive object detection (FSDAOD), involves adapting a detector to a new domain. FSDAOD is trained on a source domain with abundant data, which is then adapted to a new domain with only a small amount of data. This is particularly useful in real-world applications where collecting extensive labeled data for every possible domain is impractical. Wang \textit{et al.} \cite{fs_adaptive_frcnn} proposed the first work in this direction, which can generalize to a target domain given a few samples. All of these different approaches aim to tackle the difficulties associated with limited data in real-world scenarios, striving to balance the need for recognizing new classes and maintaining the accuracy of known classes. 

The above-mentioned variants of standard FSOD, G-FSOD, I-FSOD, O-FSOD and FSDAOD tasks differ primarily based on the availability of training data and the classes on which a model is evaluated. This survey seeks to comprehensively study these variants and analyze the developments beyond the standard FSOD setting in recent years. We begin with a comparison of existing FSOD-based survey papers in Section~\ref{sec:comp_related_works}, and clarify the need for this survey. We then briefly cover the background of object detection in Section~\ref{sec:OD-background}. The problem statement with detailed notations and differences between few shot tasks are provided in Section~\ref{sec:prob_def}. Then, we comprehensively review research works related to each few-shot task in Subsections \ref{subsec:fsod_survey} to \ref{subsec:fsdaod_survey}. This thorough review offers an overview of the recent state-of-the-art research in all these approaches. Benchmark datasets and evaluation protocols are discussed in Section~\ref{sec:datasets-evaluation}. The result analysis of all these methods is discussed in Section~\ref{sec: Discussion}. Finally, various research directions, applications and challenges in this field are explored in Section~\ref{sec:challenges-application-future}.

\begin{table}
\caption{Summary of survey papers on few-shot object detection. This highlights the contributions made by each survey paper.}
\label{Sup-tab:survey}
\vspace{-1em}
\begin{adjustbox}{width=.99\linewidth,center}
{\footnotesize
\begin{tabular}{p{2.2cm}p{12.0cm}}
\hline
\textbf{Survey Paper}  & \textbf{Keypoints}   \\ \hline
\rule{-2pt}{10pt}Jiaxu \textit{et al.} \cite{FSOD_survey2} (arXiv 2021) & 
\begin{itemize}[leftmargin=0.2cm,noitemsep,topsep=0pt,
                    before = \tablistcommand,
                    after  = \tablistcommand]
    \item Focuses on survey of few-shot object detection \cite{Attention-RPN, Meta-RCNN, meta_faster_rcnn, adf_net, BMM-CME, tip, RNI, second_order_pooling, mpsr, YOLO-FR, remote_sensing, mm_fsod, srr_fsd, meta_detr,mutual_guide_and_hybrid, RepMet, meta_retina_net, Meta-Det, fsce, context_transformer, AttFDNet, Halluc, cdgp}
    along with few shot learning\cite{siamese_fsl, closer_look_fsl}, semi supervised learning \cite{pseudo_label, stac}, weakly supervised learning \cite{grad_cam}. It divides the FSOD work into limited supervised \cite{meta_retina_net, Meta-Det, least, fsce, tip}, semi-supervised \cite{cdgp, tip} and weakly supervised \cite{vMF-MIL} based FSOD methods.
    \item Also discusses some challenges and future trends including domain transfer, mixed supervised learning, unsupervised learning, data augmentations etc.
    \end{itemize}
\\ \hline
\rowcolor{lightmintbg}\rule{-2pt}{10pt}Huang \textit{et al.} \cite{FSOD_survey4} (arXiv 2021) &
\begin{itemize}[leftmargin=0.2cm,noitemsep,topsep=0pt,
                    before = \tablistcommand,
                    after  = \tablistcommand]
    \item Focuses on few-shot object detection \cite{lstd, YOLO-FR, fsod_viewpoint_estimation, SQMGH, adf_net, drl, Attention-RPN, qa_fewdet, kernelized_fsod, fct, mm_fsod, icpe, tenet, hierarchical_attention_net, meta_faster_rcnn, apsp, second_order_pooling, meta_detr, dense_relation_distill, Meta-Det, meta_retina_net, cared, ir_fsod, tfa, defrcn, cdgp, fsce, srr_fsd, kr_fsod, CoRPN, Halluc, pdc, fsod_dis, fsod_kd}, one-shot object localization \cite{siamfc, siamrpn}, zero-shot object detection \cite{zs_yolo, zsod} and extensional zero-shot object detection \cite{zsod_td}. 
    \item The FSOD methods are divided into meta-learning \cite{meta_faster_rcnn}, transfer learning \cite{defrcn} and finetune-free \cite{AirDet} methods.
    \item The paper discusses popular benchmarks of fsod, performances of the cited papers and promising future directions.
\end{itemize}
\\ \hline
\rule{-2pt}{10pt}Antonelli \textit{et al.} \cite{FSOD_survey1} (ACM Comput. Surv. 2022)  &  \begin{itemize}[leftmargin=0.2cm,noitemsep,topsep=2pt,
                    before = \tablistcommand,
                    after  = \tablistcommand]
    \item This survey focuses on few-shot object detection. \cite{mpsr, lstd, tfa, srr_fsd, universal_prototype, RepMet, Attention-RPN, os2d, coae, BMM-CME, qa_fewdet, Meta-Det, Meta-RCNN, fsod_viewpoint_estimation}. It further divides the FSOD approaches into Data Augmentation \cite{mpsr}, Transfer Learning \cite{lstd, tfa, srr_fsd}, Distance Metric Learning \cite{RepMet} and Meta Learning \cite{Meta-Det}.
    \item It also covers zero-shot object detection \cite{zero_shot_1, zero_shot_4} approaches; discusses open issues and future research directions of this field.
\end{itemize} 
\\ \hline
\rowcolor{lightmintbg} \rule{-2pt}{7pt} Zhang \textit{et al.} \cite{FSCIL_survey1} (arXiv 2023)  & 
\begin{itemize}[leftmargin=0.2cm,noitemsep,topsep=2pt,
                    before = \tablistcommand,
                    after  = \tablistcommand]
\item This paper focuses on few-shot class incremental classification \cite{gen_iFSL, distillation_based_FSCIL} and few-shot class incremental object detection \cite{once, sylph, ifsrcnn, idetr, meta_ifsod, bpmch, imfta}.
\item The few-shot class incremental object detection is further divided into anchor-free \cite{once, sylph} and anchor-based approaches \cite{ifsrcnn}.
\end{itemize} 
\\ \hline
\rule{-2pt}{10pt}Liu \textit{et al.} \cite{FSOD_survey3} (ACM TIST 2023)  & 
\begin{itemize}[leftmargin=0.2cm,noitemsep,topsep=0pt,
                    before = \tablistcommand,
                    after  = \tablistcommand]
    \item Focuses on few-shot object detection \cite{tip, Halluc, srr_fsd, fadi, silco, mini, lvc, RepMet, dmnet, Meta-RCNN, universal_prototype, apsp, meta_faster_rcnn, BMM-CME, svd, extreme_dist_bias, cdgp, hiclpl, context_transformer, fsce, mpsr, YOLO-FR, Attention-RPN, DAna, SQMGH, kernelized_fsod, hierarchical_attention_net, meta_detr, fs_detr, fct, drl, qa_fewdet, gcn_fsod, unit, fsod_viewpoint_estimation, lstd, tfa, retentive_rcnn, knowledge_inheritance, once, sylph, dbf, idetr, defrcn} which are divided into transfer learning \cite{tfa}, meta-learning \cite{YOLO-FR}, data oriented \cite{tip, srr_fsd}, model oriented \cite{RepMet}, algorithm oriented \cite{lstd} methods.
    \item It also discusses the technical challenges associated with the FSOD task.
\end{itemize}
\\ \hline
\rowcolor{lightmintbg}\rule{-2pt}{10pt}Huang \textit{et al.} \cite{FSOD_survey6} (T-PAMI 2023)  & 
\begin{itemize}[leftmargin=0.2cm,noitemsep,topsep=0pt,
                    before = \tablistcommand,
                    after  = \tablistcommand]
    \item This survey focuses on few-shot object detection methods \cite{YOLO-FR, retentive_rcnn, lstd, tfa, mpsr, Meta-RCNN, fsod_viewpoint_estimation, Attention-RPN, meta_detr, retentive_rcnn, deformable_detr, RepMet, tip} and divided these methods into finetuning-only methods \cite{lstd}, prototype based \cite{RepMet} and modulation based \cite{tip, YOLO-FR} methods.
    \item It also briefly covers the detection tasks including weakly-supervised object detection \cite{weakly_supervised_MIL}, self-supervision using other modalities \cite{weakly_supervised_COVID}, low-data and semi-supervised object detection \cite{acrst},  few-shot semantic segmentation \cite{tosfl}, and zero-shot object detection \cite{zero_shot_2}.
\end{itemize}
\\ \hline
\rule{-2pt}{10pt}Kohler \textit{et al.} \cite{FSOD_survey7} (T-NNLS 2023)  & 
\begin{itemize}[leftmargin=0.2cm,noitemsep,topsep=0pt,
                    before = \tablistcommand,
                    after  = \tablistcommand]
    \item This survey focuses on few-shot object detection \cite{Attention-RPN, second_order_pooling, kernelized_fsod, fsod_viewpoint_estimation, SQMGH, DAna, ifc, arrm, fct, cared, apsp, qa_fewdet, FSOD-KT, meta_detr, coae, gendet, spcd, meta_faster_rcnn,sylph, tip, mm_fsod, RepMet, Meta-Det, meta_retina_net, meta_ssd, fadi, tfa, CoRPN, ford_bl, mpsr, Halluc, lvc, fsce, lstd, retentive_rcnn, bpmch, memfrcn, cfa, fssp, cdgp, dmnet, td_sampler} along with related concepts including cross-domain \cite{mtor, gpa}, zero shot \cite{zero_shot_1}, and weakly supervised \cite{weakly_sup_od} based object detection methods. It divides the FSOD techniques into meta learning and transfer learning \cite{tfa, fsce}. 
    \item It also discusses current trends which also includes the open challenges of this task.
\end{itemize} 
\\ \hline
\rowcolor{lightmintbg}\rule{-2pt}{10pt}Sa \textit{et al.} \cite{FSOD_survey10} (Applied Intelligence 2023) & \begin{itemize}[leftmargin=0.2cm,noitemsep,topsep=0pt,
                    before = \tablistcommand,
                    after  = \tablistcommand]
\item This paper performs survey of few-shot object detection \cite{YOLO-FR, Meta-RCNN, tfa, fsod_viewpoint_estimation, context_transformer, mpsr, Attention-RPN, RNI, Halluc, dense_relation_distill, defrcn, BMM-CME, fsce, meta_detr, digeo, once_shot_cross_dom} mainly in cross domain setting.
\item It focuses on approaches based on transfer learning, and several datasets including ExDark \cite{exdark_dataset}, Clipart1k \cite{clipart1k}, NEU-DET \cite{neu_det_dataset} and DOTA \cite{dota_dataset}.
\end{itemize}
\\ \hline
\rule{-2pt}{10pt}Xin \textit{et al.} \cite{FSOD_survey9} (Information Fusion 2024)  & \begin{itemize}[leftmargin=0.2cm,noitemsep,topsep=0pt,
                    before = \tablistcommand,
                    after  = \tablistcommand]
\item This work focuses on survey of few-shot object detection \cite{Meta-RCNN, YOLO-FR, Attention-RPN, Meta-Det, qa_fewdet, reinforce, fct, meta_detr, vfa, lstd, mpsr, tfa, fsce, coco_rcnn, lvc, model_calibration, fsod_dis, kd_fsod, disentangle_and_remerge, mutually_distill_sparse_rcnn, context_transformer, srr_fsd, spatial_reasoning, defrcn, weight_imprinting, retentive_rcnn, niff, bpmch, ifsod_sft, pickandplace, AirDet, ecea}. It further divides the work into single task based \cite{lstd, context_transformer, mpsr, tfa, fsce, coco_rcnn, lvc, fsod_dis, niff} and episodic task based methods \cite{Meta-RCNN, YOLO-FR, Meta-Det, qa_fewdet, reinforce, meta_detr, fct, vfa}.
\item It also compares FSOD methods on various metrics, including generalization ability, adaptive ability, mitigating domain shift, latency and discusses challenges and applications.
\end{itemize}
\\ \hline
\rowcolor{lightmintbg} \rule{-2pt}{7pt}Ours & 
\begin{itemize}[leftmargin=0.2cm,noitemsep,topsep=2pt,
                    before = \tablistcommand,
                    after  = \tablistcommand]
\item This work performs survey on FSOD in different settings: standard FSOD \cite{aht, meta_detr, mpsr, norm_vae}, generalized FSOD \cite{YOLO-FR, tfa, niff, retentive_rcnn}, incremental FSOD \cite{ws_ifsd, ifsod_sft}, open-set FSOD \cite{foxod, ofdet} and domain adaptation based FSOD \cite{pica, asyfod, cd-vito}. 
\item The standard FSOD works are categorized into meta-learning \cite{aht, meta_detr}, data sampling \cite{mpsr, td_sampler}, metric learning \cite{memfrcn, RepMet}, attention mechanism \cite{fssp, fsce}, proposal generation \cite{lvc,fct} and knowledge transfer \cite{ford_bl, data_augmentation_and_distribution_calibration} based approaches.
\item It also covers comparisons between different FSOD tasks, results discussions of all the tasks, along with several challenges, applications and future research directions.
\end{itemize}\\
\hline
\end{tabular}
}
\end{adjustbox}
\end{table}

\section{Comparison with related survey papers} \label{sec:comp_related_works}
In the field of FSOD learning, several surveys~\cite{FSOD_survey1, FSOD_survey2, FSOD_survey3, FSOD_survey4, FSOD_survey5, FSOD_survey6, FSOD_survey7, FSOD_survey9, FSOD_survey10} have been conducted to investigate and analyze various aspects of this domain. a detailed summary of the existing FSOD-based survey papers papers can be found in Table \ref{Sup-tab:survey}, while a comparison between our survey paper with the existing FSOD-based survey are depicted in Table \ref{tab:survey_compare}.

Notably, Huang \textit{et al.}~\cite{FSOD_survey6} focused on exploring the fusion of self-supervised representations with FSOD, emphasizing the importance of self-supervision pre-training in improving object detection tasks. Additionally, they discussed the challenges associated with integrating self-supervised representations with detection techniques. Another significant contribution was made by Kohler \textit{et al.}~\cite{FSOD_survey7}, who provided a comprehensive overview of the current state-of-the-art (SOTA) in FSOD methods, categorizing these approaches based on their training schemes and architectural layouts. Jiaxu \textit{et al.} \cite{FSOD_survey2} presented a data-driven taxonomy of the training data and the type of corresponding supervision utilized during the training phase. Huang \textit{et al.} \cite{FSOD_survey4} conducted a study on low-shot object detection, encompassing zero-shot, one-shot, and few-shot object detection. Antonelli \textit{et al.} \cite{FSOD_survey1} dissected FSOD methods into categories such as data augmentation, transfer learning, distance metric learning, and meta-learning-based approaches. Zhang \textit{et al.} \cite{FSCIL_survey1} delved into the realm of few-shot class incremental learning and object detection from both anchor-free and anchor-based perspectives. The authors in \cite{FSOD_survey3, FSOD_survey7} also examined the extensive field of FSOD, with \cite{FSOD_survey3} analyzing existing FSOD algorithms from a new viewpoint based on their contributions, and \cite{FSOD_survey7} categorizing approaches based on their training scheme and architectural layout, broadly classifying them into meta-learning and transfer-learning based methods. Sa \textit{et al.} \cite{FSOD_survey10} reviewed standard FSOD models focusing on few-shot object detection in a cross-domain setting. Xin \textit{et al.} \cite{FSOD_survey9} evaluated FSOD from episodic-task and single-task perspectives. However, it is worth noting that these recent surveys did not delve into the intricacies of the training mechanisms that distinguish between standard FSOD, G-FSOD, I-FSOD, O-FSOD, and FSDAOD tasks.

Analyzing these distinctions is essential for conducting fair comparisons between works and fostering a deeper understanding of these research fields. In this paper, we take a unique perspective by examining few-shot works through the lens of data availability, providing greater clarity for researchers in this domain. We divide the FSOD task into five different categories: i) standard FSOD, ii) generalized FSOD, iii) incremental FSOD, iv) open-set FSOD and v) few-shot domain adaptive object detection (FSDAOD), and extensively cover the works that tackle each of these objectives. Moreover, we segment the standard FSOD task into six sub-categories based on their proposed methodologies: i) meta-learning based, ii) data sampling and scale variation based, iii) class margin and knowledge transfer based, iv) metric learning and classification refinement based, v) attention mechanism and feature enhancement based, and vi) proposal generation and quality improvement based approaches. Additionally, we have integrated comprehensive summary tables for various FSOD approaches, improving the accessibility of this valuable information. Furthermore, to facilitate a thorough understanding of the training strategies utilized by these approaches, we have classified them based on their training schemes, providing valuable insights into the tactics employed to address their FSOD tasks. We also discuss the challenges associated with FSOD tasks in detail, along with their applications and potential directions for future research in the landscape of object detection with limited data.

\begin{table}[t!]
\caption{Comparison of our survey paper with existing FSOD-based survey papers on the variety of tasks covered and related analysis in the field.}
\label{tab:survey_compare}
\begin{adjustbox}{width=.99\linewidth,center}
\begin{tabular}{l|cccccc|ccccc}
\hline \hline
\multirow{2}{*}{\textbf{Survey Paper}} & \multicolumn{6}{c|}{Few Shot setting-based}                    & \multicolumn{5}{c}{Analysis-based} \\ \cline{2-12}
             & \multirow{2}{*}{\begin{tabular}[c]{@{}l@{}}Standard \\ FSOD \end{tabular}} & \multirow{2}{*}{\begin{tabular}[c]{@{}l@{}} G-FSOD \end{tabular}} & \multirow{2}{*}{\begin{tabular}[c]{@{}l@{}}I-FSOD \end{tabular}} & \multirow{2}{*}{\begin{tabular}[c]{@{}l@{}}O-FSOD \end{tabular}} & \multirow{2}{*}{\begin{tabular}[c]{@{}l@{}}FSDAOD \end{tabular}} &  \multirow{2}{*}{\begin{tabular}[c]{@{}l@{}}Discussion on  \\different settings \end{tabular}} & \multirow{2}{*}{\begin{tabular}[c]{@{}l@{}}Benchmark \\ Settings \end{tabular}} & \multirow{2}{*}{\begin{tabular}[c]{@{}l@{}}Code \\ Listing \end{tabular}} & \multirow{2}{*}{\begin{tabular}[c]{@{}l@{}}Application \end{tabular}}   & \multirow{2}{*}{\begin{tabular}[c]{@{}l@{}}Challenges \end{tabular}}   &  \multirow{2}{*}{\begin{tabular}[c]{@{}l@{}}Future \\ Research \end{tabular}}  \\ 
 &  &    &   &  &   &    &   &   & &  &  \\             
             \hline
Jiaxu \textit{et al.} \cite{FSOD_survey2}  &  \checkmark    & $\times$  &  $\times$  & $\times$ & $\times$  & $\times$& $\times$ & $\times$  & $\times$  & \checkmark  & \checkmark \\ \hline
\rowcolor{lightmintbg} Huang \textit{et al.} \cite{FSOD_survey4}  & \checkmark & $\times$ & $\times$ & $\times$  &  $\times$  & $\times$ & $\times$ &  $\times$ &  $\times$ & $\times$ & \checkmark \\ \hline
Antonelli \textit{et al.} \cite{FSOD_survey1}  & \checkmark & $\times$ & $\times$ & $\times$  &  $\times$  & $\times$ & $\times$  &  $\times$ &  $\times$ & \checkmark  & \checkmark \\ \hline
\rowcolor{lightmintbg} Zhang \textit{et al.} \cite{FSCIL_survey1}   & $\times$ &$\times$& \checkmark &  $\times$ &  $\times$  & $\times$ & $\times$  & $\times$  &  \checkmark &\begin{tabular}[c]{@{}c@{}} $\times$ \\ (uncertain) \end{tabular} & $\times$\\ \hline
Liu \textit{et al.} \cite{FSOD_survey3}   & \checkmark &$\times$& $\times$ & $\times$  &  $\times$  & $\times$ & $\times$  & $\times$  &  $\times$ & \checkmark & \checkmark \\ \hline
\rowcolor{lightmintbg} Huang \textit{et al.} \cite{FSOD_survey6}    & \checkmark & \checkmark & $\times$ &  $\times$ &  $\times$  & $\times$ & $\times$ & $\times$  & $\times$ &$\times$& \checkmark \\ \hline
Kohler \textit{et al.} \cite{FSOD_survey7}   & \checkmark & \checkmark & $\times$ & $\times$  &  $\times$  & $\times$ & $\times$ &$\times$ & $\times$&$\times$ & \begin{tabular}[c]{@{}c@{}} $\times$ \\ (uncertain) \end{tabular} \\ \hline
\rowcolor{lightmintbg} Sa \textit{et al.} \cite{FSOD_survey10}  & \checkmark &$\times$& $\times$ &  $\times$ &  $\times$  &$\times$ & $\times$&$\times$ & $\times$&$\times$ &$\times$\\ \hline
Xin \textit{et al.} \cite{FSOD_survey9}   & \checkmark & \checkmark & $\times$ &  $\times$ &  $\times$  & \begin{tabular}[c]{@{}c@{}}\checkmark \\ (FSOD \& G-FSOD) \end{tabular}  & $\times$ & $\times$  &  \checkmark & \checkmark & \checkmark \\ \hline
\rowcolor{lightgreen} \textbf{Ours}      &  \checkmark  &  \checkmark &  \checkmark &  \checkmark &  \checkmark & \checkmark & \checkmark & \checkmark & \checkmark &  \checkmark & \checkmark \\ \hline
\hline
\end{tabular}
\end{adjustbox}
\end{table}
\begin{figure}[t!]
    \centering
    \includegraphics[width=0.99\linewidth]{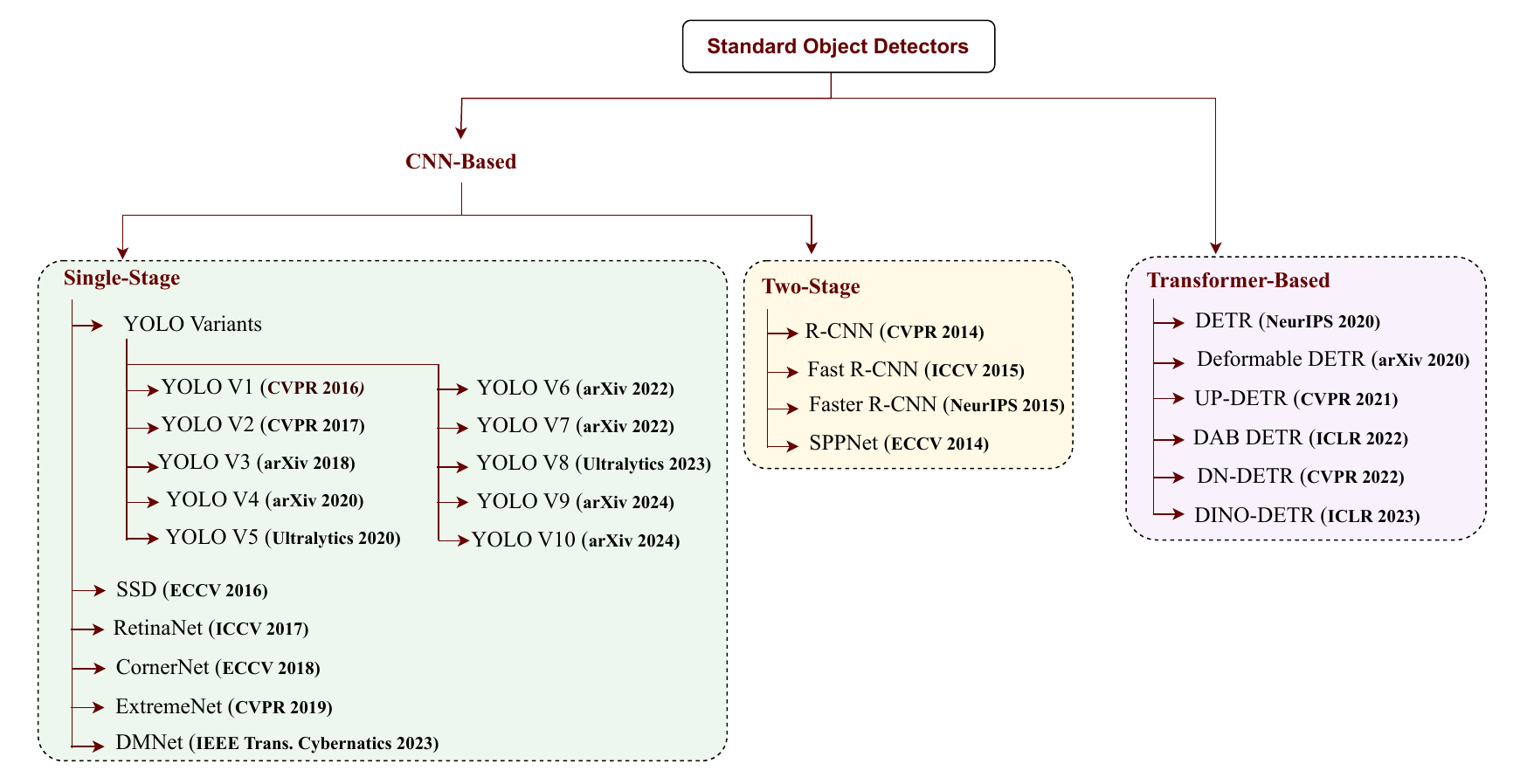}
    \caption{Taxonomy of standard object detection architectures.}
    \label{fig:summary_od_architectures}
\end{figure}
\section{Background on Object detection} \label{sec:OD-background}
In this section, we provide an overview of generic object detection. Object detection involves the tasks of localizing and recognizing objects within an image. Specifically, an object detector aims to predict bounding boxes around each object while correctly identifying their respective categories. For those new to this field, comprehensive survey papers~\cite{OD_survey3, OD_survey4, OD_survey2} offer valuable insights into object detection. The selection of model architectures significantly influences the performance of the object detection task. We categorize SOTA object detection model architectures into two primary categories:
\begin{itemize}
    \item Convolution Neural Network (CNN)-based object detectors and 
    \item Transformer-based object detectors.
\end{itemize} 
Fig.~\ref{fig:summary_od_architectures} shows the taxonomy summarizing standard object detection architectures and we also present a summary of these standard object detectors in Table \ref{tab:OD-summary}. In the following subsections, we will discuss each category in detail.

\subsection{CNN-based object detectors} \label{subsec:OD-cnn}
CNN shows impressive performance on image object classification tasks~\cite{imagenet_dataset} due to its capability of complex hierarchical features from the images. The research community has subsequently proposed leveraging these robust feature representations to enhance the performance of object detection tasks. These CNN-based object detectors can be classified into two categories:
\begin{itemize}
    \item Two-Stage detectors
    \item Single-Stage detectors
\end{itemize}
\begin{figure}[!t]
    \centering
    \begin{minipage}[t]{0.4\textwidth}
        \centering
        \includegraphics[width=\textwidth, height=0.25\textheight]{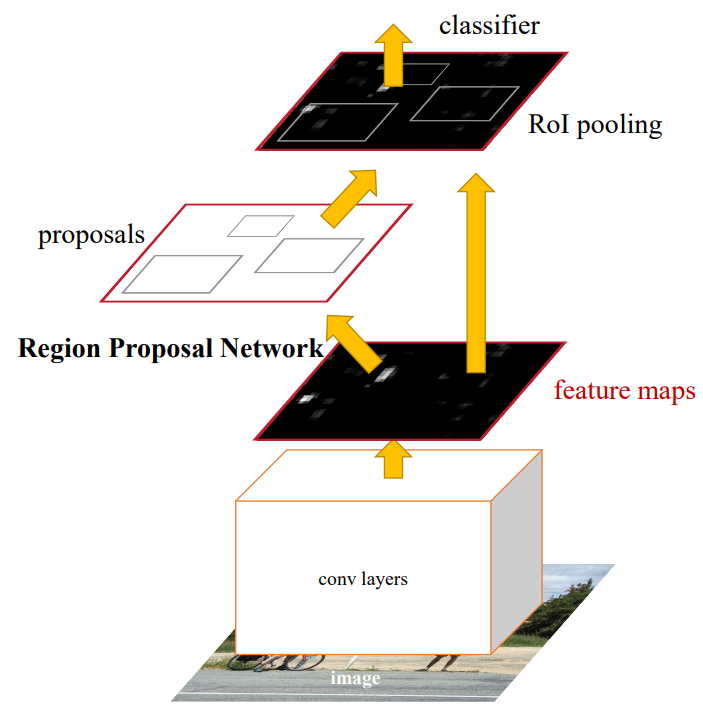}
        \caption{\textbf{Two-Stage Object Detector:} Architecture design of Faster R-CNN model. Image Courtesy from \cite{faster_rcnn}.}
        \label{fig:faster-rcnn}
    \end{minipage}
    \hfill
    \begin{minipage}[t]{0.55\textwidth}
        \centering
        \includegraphics[width=\textwidth]{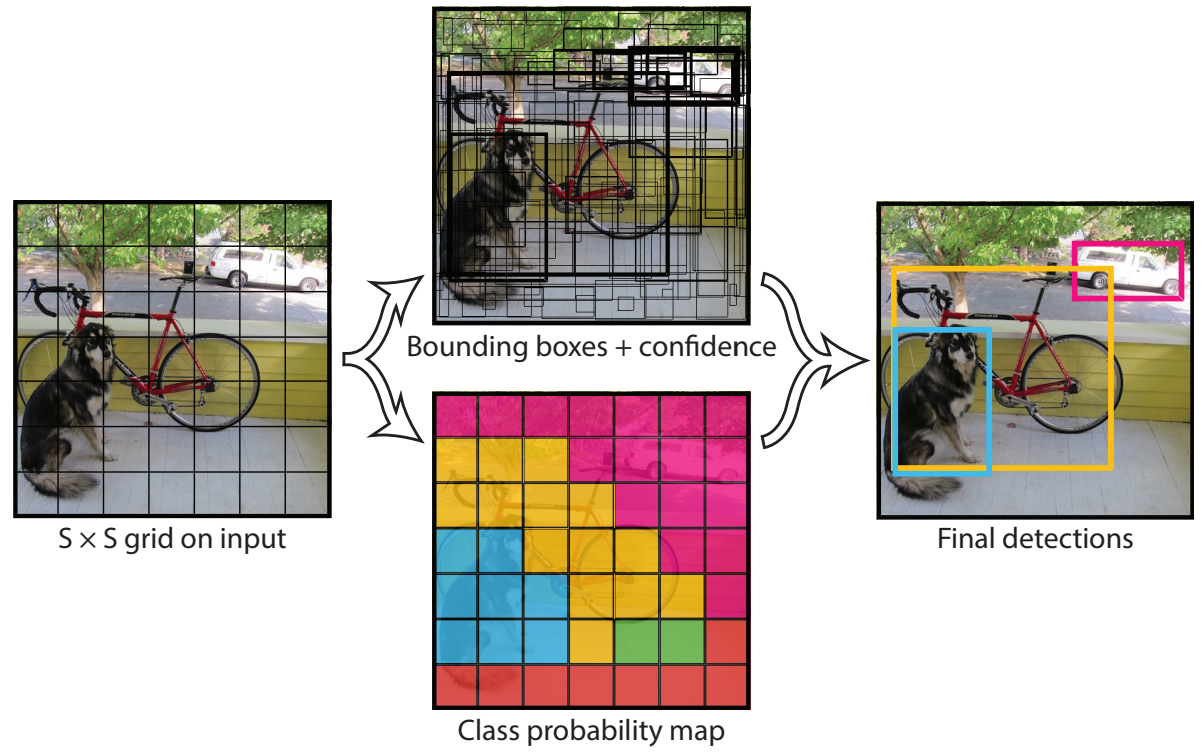}
        \caption{\textbf{Single-Stage Object Detector:} An illustration of the YOLO object detector pipeline. Image Courtesy from \cite{yolo_v1}.}
        \label{fig:yolo}
    \end{minipage}
    \vspace{-1em}
\end{figure}

\subsubsection{Two-Stage CNN-based Object Detectors:}
\label{subsec:OD-two-stage-cnn}
Faster R-CNN~\cite{faster_rcnn}, in conjunction with Feature Pyramid Networks (FPN)~\cite{fpn}, is one of the most popular two-stage architectures widely adopted in object detection. This approach is inspired from the Regions with CNN features (R-CNN)~\cite{rcnn} and Fast R-CNN~\cite{fast_rcnn} methods.

Fig.~\ref{fig:faster-rcnn} illustrates the network design of Faster R-CNN. In the initial stage, the object detector extracts features from the input image using a backbone network, resulting in single or multi-scale feature maps. These features are then input into the Region Proposal Network (RPN)~\cite{faster_rcnn}, which generates object proposals as bounding boxes. These proposals are predicted at predefined locations, scales, and aspect ratios, refined using regression, and scored for objectness. Following this, Non-Maximum Suppression (NMS)~\cite{fast_rcnn} is applied to eliminate redundant and low-quality proposals. In the second stage, each object proposal undergoes further processing. A pooled feature map is extracted by resampling the features within its bounding box to a fixed size using techniques like RoIAlign or RoIPool. This pooled feature is then passed through the Box Head or Region-of-Interest (RoI) head, which predicts the object's category and refines the bounding box using regression. NMS is applied once more to remove redundant and low-confidence predictions. The combination of the RPN and the box head is referred to as the object detector in two-stage approaches.
It is important to note that two-stage approaches like R-CNN~\cite{rcnn}, SPPNet~\cite{spp_net}, Fast R-CNN~\cite{fast_rcnn} and Faster R-CNN~\cite{faster_rcnn} require significant computational resources due to the NMS and RoI pooling processing steps, which increases the inference time and makes them highly sensitive to hyperparameters.

\subsubsection{Single-Stage CNN-based Object Detectors:}
\label{subsec:OD-one-stage-cnn}
Single-stage approaches were developed to address the complexity of two-stage detectors and optimize them for real-time applications. However, single-stage detectors may face challenges when it comes to detect dense and small objects. The pioneering You Only Look Once (YOLO)~\cite{yolo_v1} is the first single-stage detector for object detection. YOLO splits images into a $7\times 7$ grid, and for each grid cell, it predicts bounding boxes and class probabilities, resulting in a fixed number of predictions. This process is illustrated in Fig.~\ref{fig:yolo}. The approach of the YOLO model differs inherently from the iterative proposals and classifications of prior methods. Subsequently, various versions of YOLO~\cite{yolo_v2, yolo_v3, yolo_v4, yolo_v6, yolo_v7} have been proposed to improve the detection performance further.

In order to improve the performance of single-stage detectors for small objects, the Single Shot MultiBox Detector (SSD)~\cite{ssd} introduced techniques such as multi-resolution and multi-reference. This involves dividing the input image into an $S \times S$ grid, with different $S$ values for different scales. For each grid cell, a set of anchor boxes with different aspect ratios and scales is considered. The network is then trained to predict bounding box offsets and confidences for each anchor box and class. SSD builds upon YOLO by using anchor boxes adjusted to different object shapes. Subsequently, several single-stage object detectors were introduced. In \cite{retina_net}, Lin \textit{et al.} proposed RetinaNet, which addresses the class imbalance issue between background and foreground classes by introducing focal loss. In \cite{cornernet}, Law \textit{et al.} introduced CornerNet, which first identifies critical points and then uses additional embedding information to decouple and re-group these points, effectively forming bounding boxes. ExtremeNet \cite{extremenet}, on the other hand, addresses the difficulties of detecting corner points and proposes to detect extreme points, which often lie on an object and have consistent local appearance features that make them easier to detect. While techniques like CornerNet~\cite{cornernet} and ExtremeNet~\cite{extremenet} have introduced valuable approaches, they often involve costly post-processing steps, such as group-based keypoint assignment. In contrast, Zhou \textit{et al.} introduced CenterNet~\cite{centernet}, which streamlines the detection pipeline and eliminates the need for post-processing techniques like NMS, resulting in an efficient end-to-end detection network. These advancements collectively improve the accuracy and efficiency of single-stage detectors, particularly when addressing the challenges posed by small and densely packed objects.

\begin{figure}[t!]
    \centering
    \includegraphics[width=0.9\linewidth]{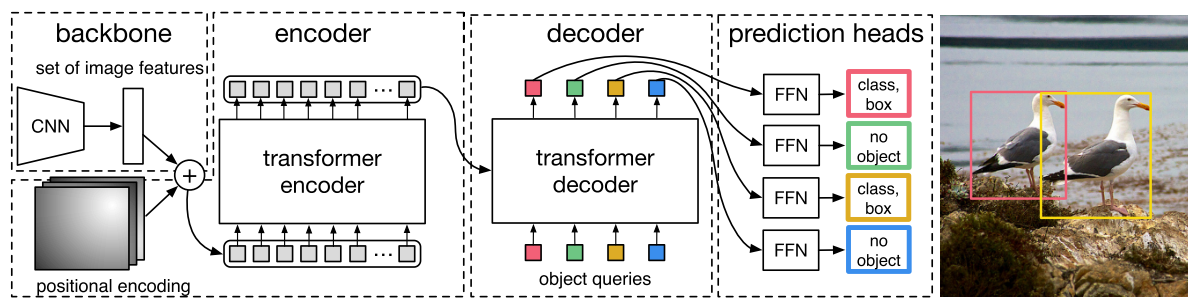}
    \caption{\textbf{Transformer-based Object Detector:} Architecture of DETR object detector. Image courtesy of Carion \textit{et al.} \cite{detr}.}
    \label{fig:DETR}
\end{figure}

\afterpage{
{\scriptsize
\begin{longtable}{|p{2.0cm}|p{7cm}|p{4cm}|}
\caption{A detailed summary of modern Object detection models.}\label{tab:OD-summary}\\
\hline
\textbf{Model} & \textbf{Keypoints} & \textbf{Loss Details} \\
\hline
\endfirsthead

\hline
\textbf{Model} & \textbf{Keypoints} & \textbf{Loss Details} \\
\hline
\endhead

\hline
\multicolumn{3}{|r|}{{Continued on next page}} \\ \hline
\endfoot

\hline \hline
\endlastfoot
\multicolumn{3}{|l|}{\textbf{CNN based Single Stage Object Detectors}} \\  
\hline
\cellcolor{lightmintbg} \rule{-2pt}{7pt}YOLO V1 & 
\cellcolor{lightmintbg}\begin{itemize}[leftmargin=0.2cm,noitemsep,topsep=2pt,before = \tablistcommand,after  = \tablistcommand]
    \item Treats detection as a regression problem from image pixels to box coordinates and class probabilities.
\end{itemize} & 
\multirow{10}{=}{%
\begin{itemize}[leftmargin=0.2cm,noitemsep,topsep=2pt,before = \tablistcommand,after  = \tablistcommand]
    \item $L_{\text{box}}$: Bounding box loss in center, height, and width format.
    \item $L_{\text{confidence}}$: IOU between predicted box and any ground truth box.
    \item $L_{\text{class probabilities}}$: Conditional classification loss.
\end{itemize}}  \\
\cline{1-2}
\rule{-2pt}{7pt}YOLO V2 & 
\begin{itemize}[leftmargin=0.2cm,noitemsep,topsep=2pt,before = \tablistcommand,after  = \tablistcommand]
    \item Uses anchor boxes with predefined shapes to match prototypical object shapes.
\end{itemize} &  \\
\cline{1-2}
\cellcolor{lightmintbg}\rule{-2pt}{7pt}YOLO V3 & 
\cellcolor{lightmintbg}\begin{itemize}[leftmargin=0.2cm,noitemsep,topsep=2pt,before = \tablistcommand,after  = \tablistcommand]
    \item Uses binary cross-entropy for training independent logistic classifiers, treating it as a multilabel classification problem.
\end{itemize} &  \\
\cline{1-2}
\rule{-2pt}{7pt}YOLO V4 \cite{yolo_v4} & 
\begin{itemize}[leftmargin=0.2cm,noitemsep,topsep=2pt,before = \tablistcommand,after  = \tablistcommand]
    \item It uses an Enhanced Architecture with Bag-of-Specials (BoS) Integration and integrates bag-of-freebies (BoF) for an Advanced Training Approach. \end{itemize} &  
\\ \cline{1-2}
\cellcolor{lightmintbg}\rule{-2pt}{7pt}YOLO V5 \cite{yolov5} & \cellcolor{lightmintbg}
\begin{itemize}[leftmargin=0.2cm,noitemsep,topsep=2pt,before = \tablistcommand,after  = \tablistcommand]
    \item Uses several augmentations such as Mosaic, random affine, MixUp, HSV augmentation, as well as other augmentations.  
    \end{itemize} &  \\ 
    \cline{1-2}
\rule{-2pt}{7pt}YOLO V6 \cite{yolo_v6} & \begin{itemize}[leftmargin=0.2cm,noitemsep,topsep=2pt,before = \tablistcommand,after  = \tablistcommand]
    \item Uses a classification VariFocal loss and an SIoU/GIoU regression loss.
    \end{itemize} 
    &  \\ \cline{1-2}
\cellcolor{lightmintbg}\rule{-2pt}{7pt}YOLO V7 \cite{yolo_v7} & \cellcolor{lightmintbg}
\begin{itemize}[leftmargin=0.2cm,noitemsep,topsep=2pt,before = \tablistcommand,after  = \tablistcommand]
    \item Uses (E-ELAN), that allows deep model to converge more efficiently by controlling the shortest longest gradient path.
\end{itemize} &  
\\ \cline{1-2}
\rule{-2pt}{7pt}YOLO V8 \cite{yolov8} & \begin{itemize}[leftmargin=0.2cm,noitemsep,topsep=2pt,before = \tablistcommand,after  = \tablistcommand]
    \item Uses an anchor-free model with a decoupled head to process objectness, classification, and regression tasks independently.
\end{itemize} &  \\ 
\cline{1-2}
\cellcolor{lightmintbg}\rule{-2pt}{7pt}YOLO V9 \cite{yolo_v9} & \cellcolor{lightmintbg}
\begin{itemize}[leftmargin=0.2cm,noitemsep,topsep=2pt,before = \tablistcommand,after  = \tablistcommand]
    \item Uses programmable gradient information (PGI) which provides complete input information for the target task to calculate objective function, so that reliable gradient information can be obtained to update network weights.
\end{itemize} &  \\ \cline{1-2}
\rule{-2pt}{7pt}YOLO V10 \cite{yolo_v10} & 
\begin{itemize}[leftmargin=0.2cm,noitemsep,topsep=2pt,before = \tablistcommand,after  = \tablistcommand]
    \item Uses consistent dual assignments for NMS-free training of YOLOs, which brings the competitive performance and low inference latency simultaneously
\end{itemize} &  \\ \cline{1-2}
\hline
\rowcolor{lightmintbg} \rule{-2pt}{7pt}SSD \cite{ssd}   &      
\begin{itemize}[leftmargin=0.2cm,noitemsep,topsep=2pt,before = \tablistcommand,after  = \tablistcommand]
    \item Produces a fixed-size collection of bounding boxes and scores for the presence of object class instances in those boxes.
    \item Matches default boxes to any ground truth with jaccard overlap higher than a threshold (0.5).
\end{itemize}                                                         &   
\begin{itemize}[leftmargin=0.2cm,noitemsep,topsep=2pt,before = \tablistcommand,after  = \tablistcommand]
    \item $L_{\text{confidence}}$: classification loss 
    \item $L_{\text{localization}}$: bounding box loss between the predicted box and ground truth box parameters
\end{itemize}                                                                \\ 
\hline
\rule{-2pt}{7pt}RetinaNet \cite{retina_net}   &   
\begin{itemize}[leftmargin=0.2cm,noitemsep,topsep=2pt,before = \tablistcommand,after  = \tablistcommand]
    \item Discovers that the extreme foreground-background class imbalance is the primary cause of low accuracy.
    \item To solve this, it introduces focal loss, reshaping the standard cross entropy loss.
\end{itemize}  
&  
\begin{itemize}[leftmargin=0.2cm,noitemsep,topsep=2pt,before = \tablistcommand,after  = \tablistcommand]
    \item $L_{\text{focal}}$: Focal loss for classification
\end{itemize}      
      \\       
      \hline    
\rowcolor{lightmintbg}\rule{-2pt}{7pt}CornerNet \cite{cornernet}  &
\begin{itemize}[leftmargin=0.2cm,noitemsep,topsep=2pt,before = \tablistcommand,after  = \tablistcommand]
    \item Detects an object as a pair of keypoints the top-left corner and bottom-right corner of the bounding box. 
    \item Predicts heatmaps and embedding vectors of the detector corners.
\end{itemize}
 &   
\begin{itemize}[leftmargin=0.2cm,noitemsep,topsep=2pt,before = \tablistcommand,after  = \tablistcommand]
    \item $L_{\text{det}}$: focal loss variant for classification.
    \item $L_{\text{off}}$: offset prediction loss of keypoints in heatmap from ground-truth images.
    \item $L_{\text{pull}}$: pull loss to group corner embeddings.
    \item $L_{\text{push}}$: push loss to separate corner embeddings.
\end{itemize}       
      \\       
      \hline    
\rule{-2pt}{7pt}ExtremeNet \cite{extremenet}   &  
\begin{itemize}[leftmargin=0.2cm,noitemsep,topsep=2pt,before = \tablistcommand,after  = \tablistcommand]
    \item Most objects are not axis-aligned boxes, and fitting them inside a box includes many distracting background pixels. 
    \item ExtremeNet detects five keypoints per class (four extreme points, and one center).
\end{itemize}
           &   
\begin{itemize}[leftmargin=0.2cm,noitemsep,topsep=2pt,before = \tablistcommand,after  = \tablistcommand]
    \item Same as CornerNet  
    \end{itemize}
      \\       
      \hline
\rowcolor{lightmintbg} \rule{-2pt}{7pt}DMNet \cite{dmnet}   &
\begin{itemize}[leftmargin=0.2cm,noitemsep,topsep=2pt,before = \tablistcommand,after  = \tablistcommand]
    \item  It contains two submodules; \textbf{DRT} achieves objectness prediction, anchor shape prediction, and decoupled feature adaption
    while \textbf{IDML} improves the generalization ability of the detector in terms of the few-shot condition.
\end{itemize}
           &   
\begin{itemize}[leftmargin=0.2cm,noitemsep,topsep=2pt,before = \tablistcommand,after  = \tablistcommand]
    \item $L_{\text{obj}}$: objectness prediction loss
    \item $L_{\text{shape}}$: anchor shape prediction loss
    \item $L_{\text{cls}}$: classification loss
    \item $L_{\text{loc}}$: localization loss
    \item $L_{\text{emb}}$: triplet loss
    \end{itemize}
      \\       
      \hline
\multicolumn{3}{|l|}{\textbf{CNN based Two-Stage Object Detectors}} \\  
\hline
\rule{-2pt}{7pt}RCNN  \cite{rcnn}  &   
\noindent Contains three modules: 
\begin{itemize}[leftmargin=0.2cm,noitemsep,topsep=7pt,before = \tablistcommand,after  = \tablistcommand]
    \item Generates category-independent region proposals.
    \item A large CNN that extracts a feature vector from each region.
    \item A set of class specific linear SVMs.
\end{itemize}
             &  
\begin{itemize}[leftmargin=0.2cm,noitemsep,topsep=2pt,before = \tablistcommand,after  = \tablistcommand]
    \item $L_{\text{loc}}$: Localization loss using L2 loss
    \item $L_{\text{classification}}$: Category specific linear SVM loss
\end{itemize}          \\ 
\hline
\rowcolor{lightmintbg}\rule{-2pt}{7pt}Fast RCNN  \cite{fast_rcnn}   &     
\begin{itemize}[leftmargin=0.2cm,noitemsep,topsep=2pt,before = \tablistcommand,after  = \tablistcommand]
\item Instead of feeding the region proposals to the CNN, it feeds the input image to the CNN to generate a convolutional  feature map. 
\item Uses the softmax classifier learnt during fine-tuning instead of training one-vs-rest linear SVMs post-hoc.
\end{itemize}  
& 
\begin{itemize}[leftmargin=0.2cm,noitemsep,topsep=2pt,before = \tablistcommand,after  = \tablistcommand]
    \item $L_{\text{loc}}$: Localization loss using smooth L1 loss
    \item $L_{\text{classification}}$: classification loss, which is a log loss for the true class
\end{itemize}     
      \\       
      \hline    
\rule{-2pt}{7pt}Faster RCNN \cite{faster_rcnn}   &  \begin{itemize}[leftmargin=0.2cm,noitemsep,topsep=2pt,before = \tablistcommand,after  = \tablistcommand]
\item Instead of using selective search algorithm on the feature map to identify the region proposals, a separate network (called Region Proposal Network (RPN)) is used to predict the region proposals.
\end{itemize}    
&   
\begin{itemize}[leftmargin=0.2cm,noitemsep,topsep=2pt,before = \tablistcommand,after  = \tablistcommand]
    \item $L_{\text{RPN Classification}}$: Classification loss over two classes (object and non-object) for the RPN. 
    \item $L_{\text{RPN regression}}$: Smooth L1 loss for regression.
    \item $L_{\text{loc}}$, $L_{\text{classification}}$: Fast RCNN losses 
\end{itemize}       
      \\       
      \hline    
\rowcolor{lightmintbg}\rule{-2pt}{7pt}SPPNet \cite{spp_net}        & 
\begin{itemize}[leftmargin=0.2cm,noitemsep,topsep=2pt,before = \tablistcommand,after  = \tablistcommand]
\item Generates a fixed-length representation regardless of image size/scale.  
\item It extracts the feature maps from the entire image only once, then it applies the spatial pyramid pooling on each candidate window of the feature maps.
\end{itemize}                                                                       &     
\begin{itemize}[leftmargin=0.2cm,noitemsep,topsep=2pt,before = \tablistcommand,after  = \tablistcommand]
    \item Same as RCNN  
    \end{itemize}
      \\       
      \hline 
\multicolumn{3}{|l|}{\textbf{Transformer based Object Detectors}} \\  
\hline
\rule{-2pt}{7pt}DETR  \cite{detr}  &  
\begin{itemize}[leftmargin=0.2cm,noitemsep,topsep=2pt,before = \tablistcommand,after  = \tablistcommand]
    \item Predicts all objects at once, and is trained end-to-end with a set loss function. 
    \item Performs bipartite matching between predicted and GT objects.
\end{itemize}
           &  
\begin{itemize}[leftmargin=0.2cm,noitemsep,topsep=2pt,before = \tablistcommand,after  = \tablistcommand]
    \item $L_{\text{match}}$: Bipartite matching loss.
    \item $L_{\text{Hungarian}}$: Hungarian loss: linear combination of a negative log-likelihood for class prediction and a box loss.
\end{itemize}      
\\ \hline
\rowcolor{lightmintbg}\rule{-2pt}{7pt}Deformable DETR  \cite{deformable_detr}    &  
\begin{itemize}[leftmargin=0.2cm,noitemsep,topsep=2pt,before = \tablistcommand,after  = \tablistcommand]
    \item Proposes a deformable attention module, which attends to a small set of sampling locations as a pre-filter for prominent key elements out of all the feature map pixels.
\end{itemize}
                   &     
\begin{itemize}[leftmargin=0.2cm,noitemsep,topsep=2pt,before = \tablistcommand,after  = \tablistcommand]
    \item Same as DETR  
    \end{itemize}
      \\       
      \hline    
\rule{-2pt}{7pt}UP-DETR \cite{up_detr}   &  
Contains pre-training and fine-tuning procedures:
\begin{itemize}[leftmargin=0.2cm,noitemsep,topsep=7pt,before = \tablistcommand,after  = \tablistcommand]
    \item The transformer is unsupervisedly pre-trained on a large-scale dataset,
    \item The model is fine-tuned with labeled data similar DETR.
\end{itemize}
  &     
\begin{itemize}[leftmargin=0.2cm,noitemsep,topsep=5pt,before = \tablistcommand,after  = \tablistcommand]
    \item $L_{\text{match}}$, $L_{\text{Hungarian}}$: DETR Losses
    \item $L_{\text{reconstruction}}$: Reconstruction loss to preserve classification during localization pre-training.
\end{itemize} 
      \\       
      \hline    
\rowcolor{lightmintbg}\rule{-2pt}{7pt}DAB-DETR   \cite{dab_detr}    & \begin{itemize}[leftmargin=0.2cm,noitemsep,topsep=2pt,before = \tablistcommand,after  = \tablistcommand]
    \item Directly uses box coordinates as queries in Transformer decoders and dynamically updates them layer-by-layer.
\end{itemize}
              &     
\begin{itemize}[leftmargin=0.2cm,noitemsep,topsep=2pt,before = \tablistcommand,after  = \tablistcommand]
    \item Same as DETR  
    \end{itemize}
      \\       
      \hline
\rule{-2pt}{7pt}DN-DETR \cite{dn_detr}     &   
\begin{itemize}[leftmargin=0.2cm,noitemsep,topsep=2pt,before = \tablistcommand,after  = \tablistcommand]
    \item Addition to Hungarian loss, it adds denoising loss as easier auxiliary task.
    \item The denoising task accelerates training and bypasses the problem faced in bipartite matching and stabilizes it.
\end{itemize}                                                             &     
\begin{itemize}[leftmargin=0.2cm,noitemsep,topsep=2pt,before = \tablistcommand,after  = \tablistcommand]
    \item $L_{\text{match}}$, $L_{\text{Hungarian}}$: DETR Losses
    \item $L_{\text{reconstruction}}$: Denoising loss (L1) to reconstruct the original ground-truth bounding box.
\end{itemize} 
      \\       
      \hline
\rowcolor{lightmintbg}\rule{-2pt}{7pt}DINO-DETR \cite{dino}   &  
\begin{itemize}[leftmargin=0.2cm,noitemsep,topsep=2pt,before = \tablistcommand,after  = \tablistcommand]
    \item Improves performance and efficiency by using a contrastive way for denoising training, a mixed query selection method for anchor initialization, and a look forward twice scheme for box prediction.
\end{itemize}
              & 
\begin{itemize}[leftmargin=0.2cm,noitemsep,topsep=2pt,before = \tablistcommand,after  = \tablistcommand]
    \item Same as DN-DETR  
    \end{itemize}
      \\       
      \hline
\end{longtable} 
}
}
\subsection{Transformer-based object detectors} 
\label{subsec:OD-transformer}
Recently, transformer-based architectures have led to significant improvements in solving language and vision problems. Carion \textit{et al.} proposed a model called DETR \cite{detr}, which treats object detection as a set prediction problem and proposed an end-to-end detection network with transformers. The architecture of the DETR object detector is shown in Fig.~\ref{fig:DETR}. Here, the image is fed to the backbone, and positional encodings are added to the features before being fed into the transformer encoder. The decoder takes object query embeddings as input and cross-attends to the encoded representation while performing self-attention on the transformed query embeddings. It then outputs a ﬁxed number of object detections, which are ﬁnally thresholded, without needing NMS.

On top of DETR, various models are been proposed. In \cite{deformable_detr}, Zhu \textit{et al.} proposed Deformable DETR to address the long convergence issues in DETR, where attention modules only attend to a small set of key sampling points around a reference. Dai \textit{et al.}~\cite{up_detr} proposed unsupervised pre-training DETR, where they took inspiration from natural language pre-trained transformers and used crop patches from the given image as queries to the decoder. 
Recently, Liu \textit{et al.} introduced Swin Transformer~\cite{swin_trans_v1}, a hierarchical Transformer with shifted windows that enhance efficiency by limiting self-attention computation to non-overlapping local windows. Swin Transformer V2~\cite{swin_trans_v2} builds upon this architecture, further scaling model capacity and window resolution. DAB-DETR~\cite{dab_detr} formulates DETR queries as dynamic anchor boxes (DAB), bridging the gap between anchor-based and DETR-like detectors. DN-DETR~\cite{dn_detr} addresses bipartite matching instability by introducing denoising (DN) techniques. Building on these ideas, DINO (DETR with Improved deNoising anchOr box)~\cite{dino} proposes contrastive denoising training and mixed query selection to enhance object detection performance.

\section{Survey of Few Shot Object Detection Methods}
\begin{figure}[t]
    \centering
    \includegraphics[width=0.99\linewidth]{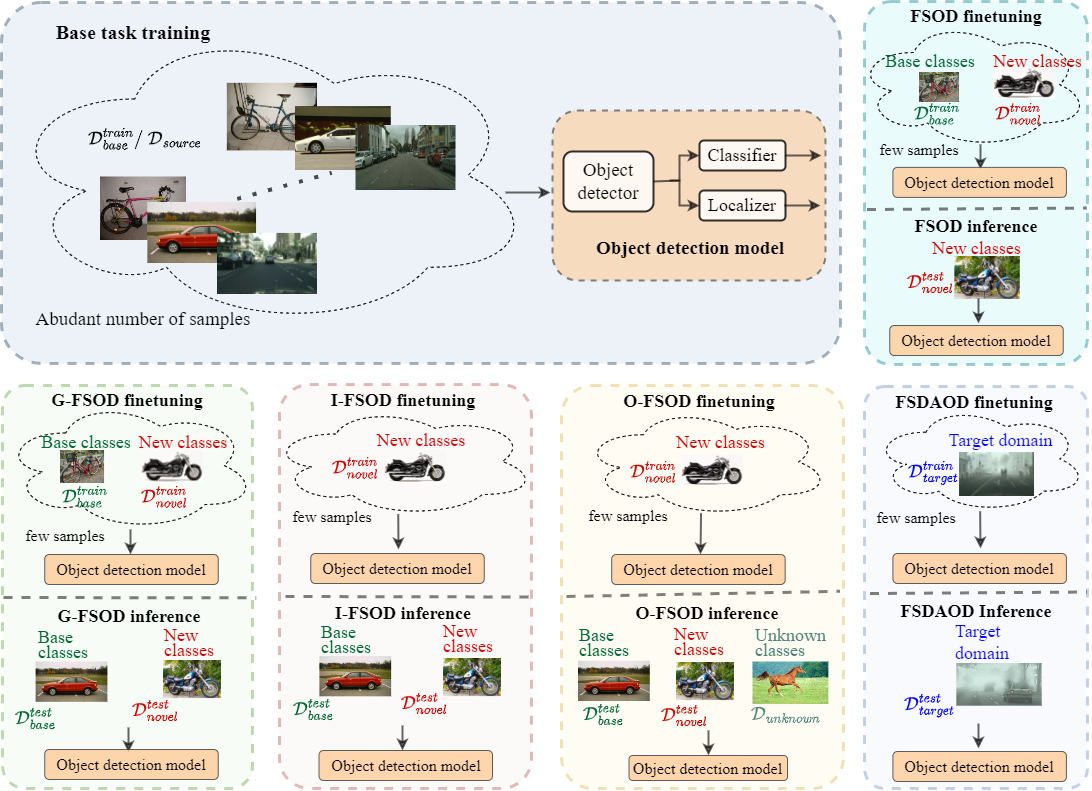}
    \caption{Comparative analysis of five key FSOD tasks: standard FSOD, G-FSOD, I-FSOD, O-FSOD and FSDAOD. The base task training remains consistent across all tasks, leveraging an extensive dataset of samples from base classes. (i) Standard FSOD: During finetuning, a limited number of samples from both old and new classes are accessible. However, during inference, the model is tested solely on new classes. (ii) G-FSOD: Finetuning sample availability is the same as FSOD. However, during inference, the model is evaluated on all base and new classes. (iii) I-FSOD: This task lacks access to samples from base classes during finetuning, relying solely on a few samples from new classes. During inference, the model is expected to perform well on both base and new classes. (iv) O-FSOD: The base training and finetuning scenario is similar to I-FSOD. Evaluation is performed on unknown/unseen classes along with seen or already trained classes. (v) FSDAOD: In this task, the base training is performed on the source dataset $\mathcal{D}_{source}$, the fine-tuning is performed on the target dataset $\mathcal{D}_{target}^{train}$. The evaluation is done on the adapted/target dataset $\mathcal{D}_{target}^{test}$.}
    \label{main-figure}
    \vspace{-1em}
\end{figure}

Figure \ref{main-figure} intuitively illustrates the network flow for various FSOD settings, including standard FSOD, G-FSOD, I-FSOD, O-FSOD, and FSDAOD. These settings are also detailed in Algorithm \ref{alg:FSOD}, which outlines the base training, fine-tuning, and inference steps for each FSOD task. This section begins with a problem definition, including notations and the differences between FSOD settings. Following this, we will discuss the taxonomy of methods related to FSOD tasks in detail. 

\subsection{Problem Definition and Difference Between FSOD Settings} 
\label{sec:prob_def}
Let $(x, y) \in \mathcal{D}$, where $\mathcal{D}$ represents the dataset consisting of images $x$ paired with their corresponding labels $y$. The labels in $y$ contain information about the category class label and bounding box coordinates. In FSOD research, the dataset $\mathcal{D}$ is typically divided into two subsets: the base dataset $\mathcal{D}_{base}$ and the novel dataset $\mathcal{D}_{novel}$, where $\mathcal{D}_{base} \cap \mathcal{D}_{novel} = \varnothing$. $\mathcal{D}_{base}$ contains a substantial amount of data and includes classes represented by labels $y_{base} \in \mathcal{C}_{B}$ (base classes). On the other hand, $\mathcal{D}_{novel}$ contains only a few instances from each category, representing classes denoted by labels $y_{novel} \in \mathcal{C}_{N}$ (novel classes). The standard notation for the few-shot object detection problem is \textit{'K-shot, M-way,'} where 'K' signifies the number of labeled instances per category, and 'M' represents the total number of distinct classes. For instance, in a \textit{'10-shot, 20-way'} setting, the model learns to recognize 20 novel categories, each with 10 instances. 

\begin{algorithm}[t]
\small
\caption{\small FSOD (Few-Shot Object Detection): Generalized View} \label{alg:FSOD}
\begin{algorithmic}[1]
\Require $\textcolor{magenta}{\mathbf{\mathcal{W}_{init}}}$: Initial Model for FSOD
\Require $\textcolor{magenta}{\mathbf{\mathcal{D}_{base}}} = \mathcal{D}_{base}^{train} \cup \mathcal{D}_{base}^{test}$: Base training data with train and test splits
\Require $\textcolor{magenta}{\mathbf{\mathcal{D}_{novel}}} = \mathcal{D}_{novel}^{train} \cup \mathcal{D}_{novel}^{test}$: Novel fine-tuning data with train and test splits
\Require $\textcolor{magenta}{\mathbf{\mathcal{D}_{unknown}}}$: Unknown classes based dataset
\Require $\textcolor{magenta}{\mathbf{\mathcal{D}_{source}}}$: Source/Base training dataset
\Require $\textcolor{magenta}{\mathbf{\mathcal{D}_{target}}} = \mathcal{D}_{target}^{train} \cup \mathcal{D}_{target}^{test}$: Target/Fine-tuning dataset
\Require $\textcolor{magenta}{\mathbf{\mathcal{N}_{base}}, \mathbf{\mathcal{N}_{novel}}}$: Number of epochs for base and novel task training respectively

\State \textcolor{brown}{\# Base Training Task}
\If{Task == ``FSDAOD''}
    \State $\mathcal{W}_{source} \leftarrow \text{TrainTask}(\mathcal{W}_{init}, \mathcal{D}_{source}, \mathcal{N}_{base})$
\Else
    \State $\mathcal{W}_{base} \leftarrow \text{TrainTask}(\mathcal{W}_{init}, \mathcal{D}_{base}^{train}, \mathcal{N}_{base})$
\EndIf

\State \textcolor{brown}{\# Fine-tuning Task}
\If{Task == ``Standard FSOD'' \textbf{or} Task == "G-FSOD"}
    \State $\mathcal{W}_{novel} \leftarrow \text{TrainTask}(\mathcal{W}_{base}, \mathcal{D}_{base}^{train} \cup \mathcal{D}_{novel}^{train}, \mathcal{N}_{novel})$
\ElsIf{Task == ``I-FSOD'' \textbf{or} Task == ``O-FSOD''}
    \State $\mathcal{W}_{novel} \leftarrow \text{TrainTask}(\mathcal{W}_{base}, \mathcal{D}_{novel}^{train}, \mathcal{N}_{novel})$
\ElsIf{Task == ``FSDAOD''}
    \State $\mathcal{W}_{target} \leftarrow \text{TrainTask}(\mathcal{W}_{source}, \mathcal{D}_{target}^{train}, \mathcal{N}_{novel})$
\EndIf

\State \textcolor{brown}{\# Inference Task} 
\If{Task == ``Standard FSOD''}
    \State Evaluation $\leftarrow$ \text{Inference}($\mathcal{W}_{novel}$, $\mathcal{D}_{novel}^{test}$)
\ElsIf{Task == ``G-FSOD'' \textbf{or} Task == "I-FSOD"}
    \State Evaluation $\leftarrow$ \text{Inference}($\mathcal{W}_{novel}$, $\mathcal{D}_{base}^{test} \cup \mathcal{D}_{novel}^{test}$)
\ElsIf{Task == ``O-FSOD''}
    \State Evaluation $\leftarrow$ \text{Inference}($\mathcal{W}_{novel}$, $\mathcal{D}_{base}^{test} \cup \mathcal{D}_{novel}^{test} \cup \mathcal{D}_{unknown}$)
\ElsIf{Task == "FSDAOD"}
    \State Evaluation $\leftarrow$ \text{Inference}($\mathcal{W}_{novel}$, $\mathcal{D}_{target}^{test}$)
\EndIf

\Function{TrainTask}{$\mathcal{W}$, $\mathcal{D}_{train}$, $\mathcal{N}$} \Comment{Function for base training / fine-tuning task}
    \For{epoch from 1 to $\mathcal{N}$}
        \State $\mathcal{L} \leftarrow \text{LossFunction}(\mathcal{D}_{train})$
        \State $\mathcal{W} \leftarrow \mathcal{W} - \text{learning\_rate} \times \nabla_W \mathcal{L}$
    \EndFor
    \State \Return $\mathcal{W}$
\EndFunction

\Function{Inference}{$\mathcal{W}$, $\mathcal{D}_{test}$} \Comment{Function for inference task}
    \State Predictions $\leftarrow$ []
    \For{image in $\mathcal{D}_{test}$}
        \State prediction $\leftarrow \mathcal{W}(\text{image})$
        \State Predictions.append(prediction)
    \EndFor
    \State \Return Predictions
\EndFunction

\vspace{-0.2em}
\end{algorithmic}
\end{algorithm}
Below, we outline various settings for differentiating FSOD tasks.
\begin{itemize}[leftmargin=0.2cm,noitemsep,topsep=15pt,
                    before = \tablistcommand,
                    after  = \tablistcommand]
\item \textbf{Standard FSOD:} The training process of standard FSOD methods involves two stages. Initially, a base model $M_{base}$ is trained on the dataset  $\mathcal{D}_{base}$, which contains the target classes $\mathcal{C}_{B}$. The prior knowledge acquired in $M_{base}$ is subsequently transferred to $\mathcal{D}_{finetune} = \mathcal{D}_{base}^{train} \bigcup \mathcal{D}_{novel}^{train}$, a dataset comprising both the base classes $\mathcal{C}_{B}$ and few-shot classes $\mathcal{C}_{N}$. Finally, the evaluation is carried out on $\mathcal{D}_{novel}^{test}$, which includes only the novel classes $\mathcal{C}_{N}$. For example, detecting rare wildlife species often involves limited data, which can lead to overfitting and poor performance. FSOD enhances detection performance on rare objects by pre-training the model on abundant data from other species.
\item \textbf{G-FSOD:} The base training and fine-tuning phases are similar to FSOD. However, the evaluation is conducted on $\mathcal{D}_{test} = \mathcal{D}_{base}^{test} \bigcup \mathcal{D}_{novel}^{test}$, which contains both the base classes $\mathcal{C}_{B}$ and the novel classes $\mathcal{C}_{N}$. For instance, an autonomous vehicle in a dynamic urban environment must recognize rare objects without forgetting previously learned information. A small amount of data from base classes can be combined with new class data to ensure high performance across all categories.
\item \textbf{I-FSOD:} The base training process is identical to FSOD. However, during the fine-tuning stage, only the novel classes dataset $\mathcal{D}_{novel}^{train}$, containing $\mathcal{C}_{N}$, is used, without access to the base dataset $\mathcal{D}_{base}$. The evaluation process is the same as in G-FSOD and is conducted on $\mathcal{D}_{test}$, which includes both the sets of base classes $\mathcal{C}_{B}$ and the novel classes $\mathcal{C}_{N}$. The primary goal of I-FSOD is to learn the novel classes $\mathcal{C}_{N}$ while avoiding catastrophic forgetting of the base classes $\mathcal{C}_{B}$. For instance, consider an AI system designed to recognize new faces with limited data. Due to privacy regulations, it is not allowed to store images and information of previously recognized faces. In this scenario, I-FSOD ensures that the face recognition system maintains high accuracy in identifying new and previously encountered individuals while adhering to privacy constraints.
\item \textbf{O-FSOD:} The base training is performed on $\mathcal{D}_{base}^{train}$ which contains abundant base classes $\mathcal{C}_{B}$ while the fine-tuning is performed on $\mathcal{D}_{novel}^{train}$ with scarce novel classes $\mathcal{C}_{N}$ which together forms the known classes $\mathcal{C}_{K} = \mathcal{C}_{B} \bigcup \mathcal{C}_{N}$. The final model is tested on $\mathcal{D}_{test} = \mathcal{D}_{base}^{test} \bigcup \mathcal{D}_{novel}^{test} \bigcup \mathcal{D}_{unknown}$ which contains the test classes $\mathcal{C}_{test}$, where $\mathcal{C}_{test} = \mathcal{C}_{K} \bigcup \mathcal{C}_{U}$, $\mathcal{C}_{U}$ is the unknown class, and $\mathcal{C}_{K} \cap \mathcal{C}_{U} = \emptyset$. The goal is to employ the unbalanced data to train a detector, which can be used to identify the base classes, the novel classes, and the unknown class. For example, an autonomous vehicle may encounter an unusual type of construction equipment or a new kind of road obstacle, like debris or temporary signs. With O-FSOD, the vehicle's detection system can quickly learn to recognize these new objects using just a few labeled examples, allowing it to navigate safely around them.
\item \textbf{FSDAOD:} The model is initially trained on a source domain dataset $\mathcal{D}_{source}$, which is sufficiently large and contains classes $\mathcal{C}_{source}$.  It is then adapted to a target domain dataset $\mathcal{D}_{target}$, which is data-scarce and contains classes $\mathcal{C}_{target}$. This approach addresses transfer scenarios involving domain discrepancies between the source and target distributions. The label space is identical for both domains, i.e., $\mathcal{C}_{source} = \mathcal{C}_{target}$, but the data distributions differ. During the base training stage, the source dataset serves as the base dataset. In the fine-tuning/adaptation stage, the target training dataset $\mathcal{D}_{target}^{train}$ is used, and the evaluation is performed on the target testing dataset $\mathcal{D}_{target}^{test}$. For instance, training an object detector in autonomous driving typically requires abundant data across various classes. However, collecting sufficient real-world data for all necessary classes is challenging. To overcome this, the detector is initially trained on simulated data, which can be generated in large quantities despite having domain discrepancies with real-world data. After this, the model is adapted to real-world data using only a few samples for each class.
\end{itemize}

\begin{figure*}[t!]
    \centering
    \includegraphics[width=\linewidth]{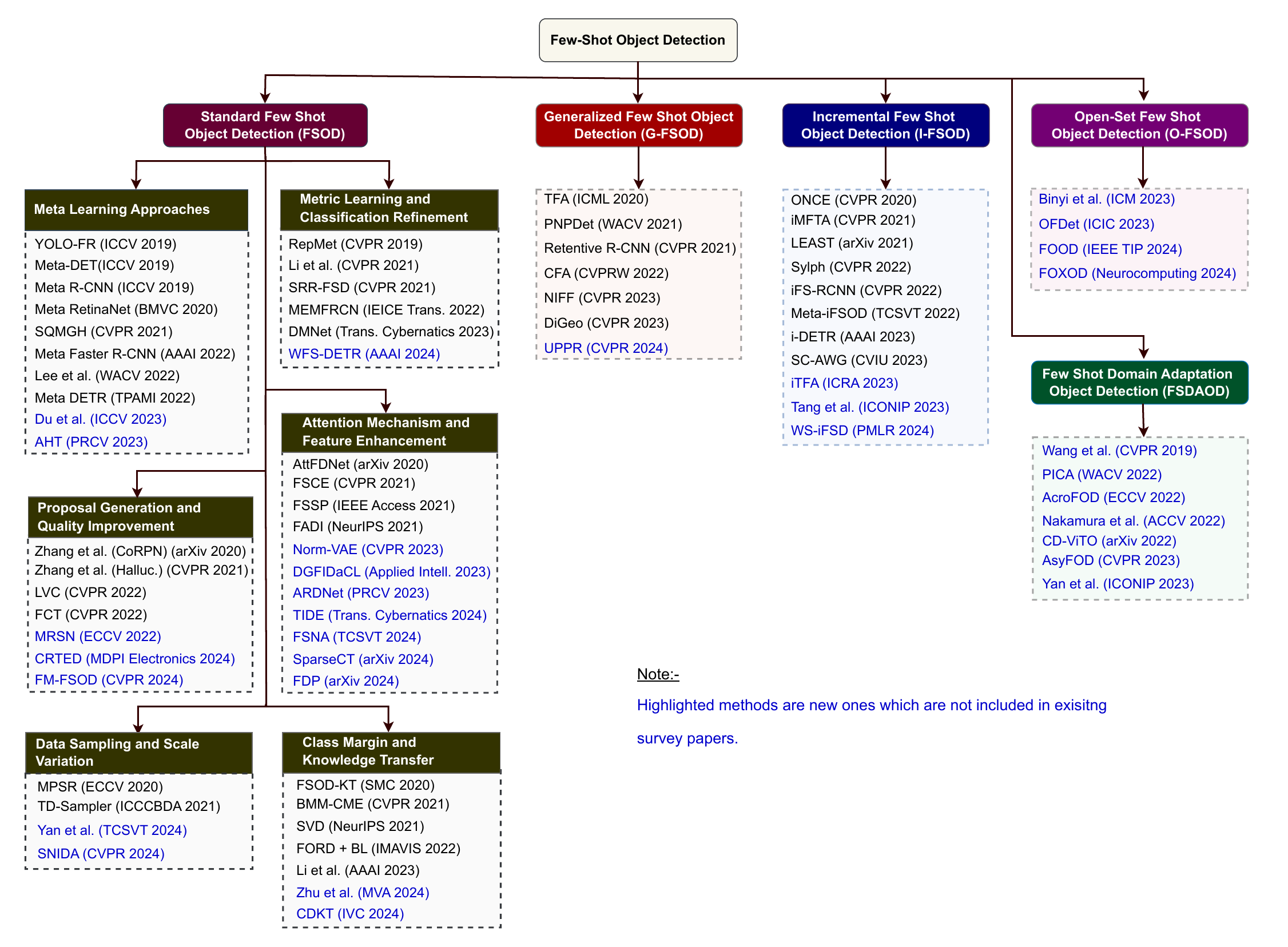}
    \vspace{-2em}
    \caption{A taxonomy of few-shot approaches for object detection task along with publication timeline. Here \textcolor{blue}{\textbf{highlighted methods}} indicate methods which are not included in other existing survey article.}
    \label{fig:taxonomy}
    \vspace{-1em}
\end{figure*}

In Fig.~\ref{fig:taxonomy}, we summarize the taxonomy of the above-mentioned FSOD approaches along with their publication timeline. Subsequent subsections will discuss these methods in detail.

\subsection{Standard Few Shot Object Detection (FSOD)}
\label{subsec:fsod_survey}
We classified the standard FSOD approaches into subcategories based on the proposed techniques: 1) Meta-learning based approaches, 2) Metric learning and classification refinement-based approaches, 3) Data sampling and scale variation-based approaches, 4) Attention mechanism and feature enhancement-based approaches, 5) Class margin and knowledge transfer based approaches, 6) Proposal generation and quality improvement based approaches. In the following subsections, we discuss these different approaches in detail.

\subsubsection{Meta-Learning Approaches:} 
Meta-learning is a widely used approach in few-shot learning that enables models to acquire the ability to learn. By exposing models to various training scenarios with limited data, meta-learning allows quick adaptation and generalization to new tasks. The support set utilized in meta-training includes examples from base tasks, which helps the model grasp general patterns. On the other hand, the query set used during evaluation consists of examples from new tasks to evaluate the model's generalization and prediction capabilities. Mainly meta-learning approaches adopt either feature reweighting~\cite{YOLO-FR, Meta-Det} or its variants to aggregate query and support features~\cite{apsp, SQMGH} to tackle the FSOD problem.

In this category, YOLO-FR~\cite{YOLO-FR} addressed the FSOD problem by using a single-stage YOLO-v2 object detection model. This approach utilizes meta-training to extract generalizable meta-features from fully labeled base classes. A reweighting module is used to assign importance to meta-features for novel object detection. By taking support images as input, embedding them into class-specific representations, and using these embeddings to reweigh the meta-features, YOLO-FR produce more crucial features for detecting new target objects. 

Wang \textit{et al.}~\cite{Meta-Det} introduced a unified meta-learning approach known as Meta-Det for few-shot classification and localization tasks. This separates the learning process of category-agnostic and category-specific parameters in CNN-based detectors, specifically Faster-RCNN. Meta-Det is initially trained with a large dataset to acquire category-agnostic parameters and then fine-tuned with samples from few-shot tasks to learn category-specific parameters. Despite the challenges of effectively learning from limited examples, the authors utilize a meta-model trained through a meta-training procedure to estimate category-agnostic transformations and parametrized weights for classification based on these transformations. In \cite{Meta-RCNN}, Yan \textit{et al.} extended the meta-learning capabilities to both object detection and segmentation tasks using Faster/Mask R-CNN (referred to as Meta-RCNN). The meta-predictor head of Meta-RCNN predicts bounding boxes and segmentation masks based on RoI features generated by Meta-RCNN using the support set.

The above-mentioned meta-learning strategies utilized a single prototype for each category derived from support samples. Recent advancements aim to enhance the utilization of information from each support sample. Lee \textit{et al.}~\cite{apsp} introduced the concept of Attending to Per-Sample-Prototype (APSP), which treats each support sample as an individual prototype. By employing an attention mechanism, APSP enhances model feature representations by capturing shared information among these individual prototypes. This versatile module can be seamlessly integrated into existing meta-learning frameworks. In contrast, Support-Query Mutual Guidance (SQMGH)~\cite{SQMGH} utilizes a support-query mutual guidance approach to obtain more relevant support proposals. This method generates the final aggregated support feature using a query guidance strategy and achieves mutual guidance between support and query features using contrastive loss and focal loss.

Han \textit{et al.}~\cite{meta_faster_rcnn} observed that proposals for few-shot classes tend to be less accurate compared to those for many-shot classes, resulting in problems like missing boxes due to misclassification or imprecise spatial locations from noisy RPN proposals. To overcome this limitation, Han \textit{et al.} introduced the prototype matching network known as Meta Faster R-CNN. This method replaces the traditional linear object classifier in RPN with a ``Meta-Classifier", showcasing enhanced accuracy of produced bounding boxes for few-shot classes. Li \textit{et al.}~\cite{meta_retina_net} introduced the Meta RetinaNet method, incorporating the single-stage RetinaNet architecture. The authors argue that the focal loss function in RetinaNet helps alleviate bias towards base classes, ultimately improving generalization on new classes by enhancing proposals. Additionally, they propose a balanced loss to work alongside the focal loss, boosting performance in the FSOD scenario. Nonetheless, the approach does not address how RetinaNet handles noisy region proposals. 

Within the region-based detection framework, the accuracy of the final predictions relies heavily on the proposed regions. However, generating high-quality region proposals is challenging in the limited supervision of few-shot settings. To address this issue, Zhang \textit{et al.}~\cite{meta_detr} introduced an approach utilizing transformer architectures. Their solution, Meta-DETR, enhances the DETR architecture by incorporating inter-class correlation training to manage multiple support classes concurrently. Integrating the transformer architecture into FSOD facilitates the direct generation of high-quality proposals from queries inputted into the DETR decoder. 

In \cite{aht}, Lai \textit{et al.} introduced a novel approach called AHT to address the limitations of previous approaches. 
ATH incorporates multi-level fine-grained two-branch interactions and dynamically generates class-specific primary network weights using a feed-forward meta-learning approach. The proposed method consists of two modules: the Dynamic Aggregation Module (DAM) adaptively generates inter-image prominent features into aggregated weights, and the Conditional Adaptation Hypernetworks (CAH) module uses these aggregated weight vectors as conditions for the hypernetwork to generate class-specific parameters dynamically.

In their study, Du \textit{et al.} \cite{alpha-adaptive} addressed the drawback of meta-learning-based methods utilizing a single K-average-pooled prototype in both RPN and detection head for query detection. This straightforward approach impacts FSOD performance in two key ways: 1) the poor quality of the prototype and 2) the equivocal guidance due to the contradictions between the RPN and the detection head. Du \textit{et al.} prioritize salient representations and de-emphasize trivial variations by accessing both angle distance and magnitude dispersion ($\sigma$) across K-support samples to generate high-quality proposals. It robustly deals with intra-class variations, and a simple K-average pooling is enough to generate a high-quality prototype for meta-testing.

\subsubsection{Metric Learning and Classification Refinement:}
This section explores two key techniques in standard FSOD: metric learning and classification refinement. 
Metric learning focuses on learning a similarity function that can accurately measure the similarity or dissimilarity between different objects. 
By learning an effective metric, the model can better distinguish between similar and dissimilar objects, even with limited data.
Classification refinement, on the other hand, improves a pre-trained classifier's decision-making, often tackling specific data challenges with ease.

One notable work is Representative-based metric learning (RepMet)~\cite{RepMet}, which uses a unique strategy for FSOD by employing metric learning based on class representatives. Here, each class is represented by a multi-modal mixture model, with representative vectors as the centers, capturing intra-class variations and creating a customized embedding space for similarity-based classification. Instead of using a traditional classifier head, a subnet calculates class posteriors for each region of interest (ROI) by comparing its embedding vector to the class representatives. This architecture allows joint training of the embedding space and mixture distributions, enabling few-shot capabilities. Once trained, the distance metric learner classifier can easily accommodate new categories with minimal supervision, transforming into a powerful few-shot detector.

RepMet has the potential for few-shot detection but is inefficient due to its two-stage approach with ROI pooling in the distance metric learning (DML) module. Lu \textit{et al.} \cite{dmnet} proposed DMNet that improves efficiency with a single-stage FSOD design. DMNet comprises two key components: Decoupled Representation Transformation (DRT) and Image-Level DML (IDML). DRT focuses on three areas: 1) extracting foreground representations to filter backgrounds, 2) predicting adaptive anchor shapes to improve over manually crafted ones, and 3) adjusting features for classification and localization tasks based on their receptive field needs. This separation optimizes feature learning for each task. On the other hand, IDML works on the entire feature map, enabling parallel multi-object inference and boosting efficiency and generalization. This integration fits smoothly into single-stage detection pipelines, marking DMNet as a significant advancement in FSOD.

Classifier refinement is crucial in zero-shot recognition and detection, using semantic words for tasks without prior examples. Zhu \textit{et al.}~\cite{srr_fsd} introduced a method called Semantic Relation Reasoning for Shot-Stable Few-Shot Object Detection (SRR-FSD), which uses semantic embeddings from a large text dataset to represent categories. The detector maps object images into this embedding space and employs a dynamic relation graph driven by image data to enhance initial embeddings. This dynamic process creates a more reliable and consistent few-shot detector, especially with limited training data for new objects.

Beyond the domain gap between visual and semantic spaces, FSOD methods face two major challenges: \textit{destructive samples} and \textit{limited inter-class separability}~\cite{cdgp}. Destructive samples result from incomplete annotations in base set images, where novel class objects are unlabeled, misleading the model. Additionally, the scarcity of data for novel classes often leads to poor distinguishability between categories, causing confusion. These issues partly result from Faster-RCNN-based detectors, which use a shared feature representation for classification and localization. This is suboptimal, as localization needs translation-covariant features, while classification requires translation-invariant features, affecting classifier performance with limited training data. 

Li \textit{et al.}~\cite{cdgp} addressed these challenges through two strategies: (i) Confidence-Guided Dataset Pruning (CGDP) to refine the training dataset by eliminating distractions, and (ii) the Few-Shot Correction Network (FSCN), an independent classifier that preserves translation-invariant features for classification by refining region proposals, separating the tasks of classification and localization. 

MemeFRCN~\cite{memfrcn} addressed the destructive samples and inter-class separability issues with three components:  a memory-based classifier (MemCla), a fully connected neural network classifier (FCC), and an adaptive fusion block (AdFus). MemCla stores embedding vectors to preserve previously encountered classes and enhance inter-class separability. FCC predicts categories based on current features. AdFus adjusts the fusion method between MemCla and FCC based on available samples, ensuring adaptability. However, empirical evidence of this approach preventing catastrophic forgetting is lacking, which is crucial for improving inter-class separability.
 
Zhang \textit{et al.} \cite{wfs_detr} addressed the limitations of previous FSOD methods that require strong annotations like category labels and bounding boxes. They propose WFS-DETR for weakly supervised settings. Initially, it develops object localization and integrity judgment on large-scale pretraining data. Then, it incorporates object integrity into multiple-instance learning, using both semantic and visual data to address local optima issues. Finally, simple fine-tuning transfers knowledge from base classes to novel classes, enabling the accurate detection of novel objects.

\subsubsection{Proposal Generation and Quality Improvement}
This section explores various methods to improve the quality of proposals in object detection networks to tackle challenges in FSOD. Proposal generation is a popular step in an object detection task, and therefore optimizing this process is crucial for enhancing detection accuracy and efficiency in few-shot object detection scenarios. According to Zhang \textit{et al.} ~\cite{CoRPN}, the absence of just one high-IOU training box during RPN training can significantly impact the classifier's ability to capture object appearance variations. To overcome this issue, they introduced CoRPN (Cooperating RPNs), which involves training multiple redundant RPNs. These RPNs work independently but collaborate to ensure that if one misses a high-IOU box, another will likely detect it. This approach enhances the quality of proposals and ultimately aids in classifier training with limited data. 

In a subsequent study, Zhang \textit{et al.}~\cite{Halluc} introduced the Hallucinator Network, which generates additional training examples in the Region of Interest (RoI) feature space. These examples are then integrated into the object detection model to select high-IOU boxes. The authors emphasized the importance of effectively addressing the lack of variation in training data for extremely few-shot detection performance. Training the hallucinator and the detector's classifier using an expectation-maximization (EM)-like approach is crucial in addressing this issue.

Kaul \textit{et al.}~\cite{lvc} take a different approach by utilizing unlabeled data and a pseudo-labeling technique to enhance proposal quality. They generate high-quality pseudo-annotations for novel categories by leveraging unlabeled images in a few-shot adaptation. This involves building a classifier for novel categories using features from a self-supervised network to verify candidate detections and training a specialized box regressor to refine the bounding boxes of verified candidates. Through this two-step verification and refinement process, they can achieve high-precision pseudo-annotations, effectively balancing the training data and boosting the performance of FSOD.

On the other hand, Han \textit{et al.}~\cite{fct} draw inspiration from the transformer architecture to propose a cross-transformer RoI feature extractor called FCT. This method integrates the transformer architecture into the backbone network to improve proposal quality. By incorporating self-attention mechanisms, FCT can capture long-range dependencies and contextual information within the features. This can potentially enhance proposal generation and object detection in few-shot scenarios.

Ma \textit{et al.} in their work MRSN \cite{mrsn}, highlight the drawbacks of using meta-learning and transfer learning-based approaches, where images from the base set containing unlabeled novel-class objects can easily lead to performance degradation and poor plasticity since those novel objects are served as the background. In contrast, MRSN uses a semi-supervised framework to identify unlabeled novel class instances. It includes a mining model to discover these instances and an absorbed model to learn from them. It designs the Proposal Contrastive Consistency (PCC) module in the absorbed model to exploit class characteristics and avoid bias from noise labels. It utilizes PCC at the proposal level to compare the global and local information of the instance simultaneously.

In \cite{crted}, Chen \textit{et al.} identified issues with previous methods, such as increased latency from extensive fine-tuning and subpar performance when adapting to new classes. To address these shortcomings, they introduced a new FSOD model called CRTED. This model utilizes a correlation-aware region proposal network (Correlation-RPN) structure to enhance detectors' object localization and generalization capabilities. The CRTED model focuses on learning object-specific features related to inter-class correlation and intra-class compactness while minimizing object-agnostic background features, even with limited annotated samples. This approach fosters the learning of correlated features across different categories, which in turn facilitates the transfer of knowledge from base to novel categories for object detection.

Han \textit{et al.} \cite{fm_fsod} studied FSOD using various foundational models for visual feature extraction and few-shot proposal classification. They proposed a method called FM-FSOD, which is evaluated on multiple pre-trained vision models \cite{clip,sam,dinov2}. Their findings showed that DINOv2, pre-trained with both image-level and patch-level self-supervised objectives and equipped with a Transformer-based detection framework, achieved the best performance. For proposal generation, FM-FSOD utilizes the in-context learning capabilities of pre-trained Large Language Models (LLMs) for contextualized few-shot proposal classification in FSOD. The FM-FSOD can automatically exploit various contextual information between proposals and classes through the LLMs, including proposal-proposal, proposal-class, and class-class relations. The extracted context information significantly enhances few-shot proposal classification from the same query image.

\subsubsection{Attention Mechanisms and Feature Enhancement:}
This section explores methods that utilize attention mechanisms and feature enhancement for FSOD tasks. Attention mechanisms are often used to direct models towards relevant spatial areas, prioritizing them during parameter updates. However, training an attention model that can generalize well can be difficult with limited training data, as it heavily depends on top-down supervision. To address this, Chen \textit{et al.}~\cite{AttFDNet} introduced the Attentive Few-Shot Detection Network (AttFDNet) with two key innovations: (i) Bottom-up Attention: Instead of just top-down attention~\cite{clss_power_cdfc_ref_4}, AttFDNet uses bottom-up attention to leverage visual saliency for identifying interesting objects, even from unseen categories, aiding accurate classification with limited data, and (ii) Enhanced Intra-Class Agreement: AttFDNet uses object and background concentration loss to improve learning from few samples. These losses help objects of the same class to cluster together and push background regions apart, addressing issues with ``hard negative anchors" and enhancing intra-class consistency.

To enhance general feature space representations, Cao \textit{et al.}~\cite{fadi} proposed a model called FADI. This approach involves a two-step strategy. Firstly, in the association step, FADI constructs a compact feature space for novel classes by explicitly imitating a specific base class that is semantically similar. This allows the novel class to benefit from the associated base class's well-trained feature space instead of implicitly relying on multiple base classes. Secondly, in the discrimination step, FADI ensures separability between the novel and associated base classes by utilizing separate classification branches. To further enhance inter-class separability, a set-specialized margin loss is employed.

Sun \textit{et al.}~\cite{fsce} proposed FSCE (Few-Shot Object Detection via Contrastive Proposals Encoding) to enhance feature representations in FSOD by using contrastive loss. FSCE introduces a contrastive branch to the primary RoI head when transferring the base detector to few-shot novel data. This branch measures the similarity between object proposal encodings. The supervised contrastive objective, called Contrastive Proposal Encoding (CPE) loss, reduces the variance of embeddings from the same category and separates different categories. This multi-task integration improves the model's ability to discern object proposal similarities and differences, enhancing feature representations for FSOD.

Xu \textit{et al.} \cite{fssp} introduced a sample processing-based FSOD method, known as FSSP, which includes Self-Attention Module (SAM) and Positive-Sample Augmentation (PSA) module. The SAM aims to enhance the extraction of representative features from challenging samples. Meanwhile, the PSA module increases the number of positive samples and diversifies their scale distribution, thereby inhibiting the growth of negative samples. The self-attention mechanism in SAM mimics human vision by identifying valuable features and encouraging heterogeneous objects within the same category to align with one another.

In \cite{norm_vae}, Xu \textit{et al.} proposed Norm-VAE to address the lack of crop-related diversity in training data for novel classes. Norm-VAE transforms the latent space so that different latent code norms represent different crop-related variations, allowing the generation of features with varying difficulty levels. The generative model controls the difficulty of generated samples by ensuring that the latent code's magnitude correlates with the feature's difficulty level.

Huang \textit{et al.} \cite{dgfidacl} addressed two major problems in meta-learning-based approaches: how to interact with information efficiently and how to learn a reasonable decision boundary in the embedding space. To address these challenges, they integrate dense spatial and global context attention to capture the correlation across the support and query images using a dense global feature interaction module (DGFI). The DGFI module then leverages the correlation to yield the interactive features. They additionally incorporate object-level contrastive learning to endow learned features with good intra-class similarity and inter-class distinction using a dual-contrastive learning (DaCL) module. The more discriminative features form a well-separated decision boundary in the embedding space and alleviate the typical misclassification problem.

Recently, Duan \textit{et al.} \cite{ardnet} introduced a new FSOD algorithm called Adaptive Relation Distillation-based Detection Network (ARDNet). This algorithm improves the fusion of query and support features by utilizing adaptive relation distillation. To extract enough information from the support features, it uses the Adaptive Relational Distillation Module (ARDM) using a hybrid attention mechanism. The ARDM enhances previous algorithms by discarding hand-designed and inefficient query information utilization strategies. It makes full use of the support set and is not limited to the global information or local details of the support set.

Li \textit{et al.} \cite{tide} addressed the challenges of parametric readjustments in generalizing to novel objects, especially in industrial applications. These challenges include limited time for fine-tuning and the unavailability of model parameters due to privilege protection, making fine-tuning impractical or impossible. To overcome these issues, they propose TIDE, a novel approach that avoids the need for model fine-tuning during the configuration process. TIDE uses a dynamic classifier that adapts based on the support instance, enabling the model to learn and generalize to novel objects without additional fine-tuning. This approach ensures that the model can be quickly and efficiently configured for new tasks, making it highly suitable for industrial applications.

Zhu \textit{et al.} \cite{fsna} introduced a new FSOD approach called FSNA, which utilizes Neighborhood Information Adaption (NIA) and an attention mechanism to create informative features that improve data utilization. The NIA module dynamically uses object-encompassing features to recognize objects in the FSOD task. Notably, the NIA module is a parameter-free method that adapts information according to the sizes and positions of the objects. Furthermore, FSNA employs an attention mechanism to capture the affinity between different pixels within a sample and assimilate the potential relationships among diverse samples, ultimately obtaining more global and refined feature representations for few-shot data.

Mei \textit{et al.} \cite{sparse_ct} introduced a sparse context transformer (SparseCT) that efficiently uses object knowledge from the source domain while automatically learning a sparse context from a limited number of training images in the target domain. This approach integrates various relevant clues to enhance the discrimination capability of the learned detectors and minimize class confusion. To efficiently capture task-related context from the limited training data, they designed an attention-focus layer that amplifies the representation of contextual fields and reduces category confusion.

Wang \textit{et al.} \cite{fpd} discovered that in meta-learning approaches, the features of the support and query branches are fused on top of the framework to make the final prediction. However, most layers remain separate and do not exchange information, which hinders the model from learning the correlations among detailed features, especially in data-scarce scenarios. To address this issue, Wang \textit{et al.} proposed a feature enhancement module called Fine-Grained Prototypes Distillation (FPD) for FSOD. In FDP, Wang \textit{et al.} introduced Fine-Grained Feature Aggregation (FFA) mechanism to aggregate the mid-level features. This module distills support features into fine-grained prototypes before integrating them into query feature maps, thereby helping the model to grasp key information. They also proposed a Balanced Class-Agnostic Sampling (B-CAS) strategy and a Non-Linear Fusion (NLF) module to fuse high-level features effectively.

\subsubsection{Data Sampling and Scale Variation:}
This section discusses the data sampling and scale variations-based approaches in the FSOD task. Data sampling enhances the model's ability to generalize and perform accurately on new, unseen classes. Proper data sampling ensures diverse and representative examples.

Scale variations pose a significant challenge in FSOD, as objects can appear at different sizes within an image. Wu \textit{et al.}~\cite{mpsr} highlighted that object detection, involving both classification and localization, is inherently more complex than image classification. They introduced the Multi-scale Positive Sample Refinement (MPSR) approach for FSOD, explicitly targeting the sparse scale distribution challenge. MPSR builds upon the Faster R-CNN model, incorporating a Feature Pyramid Network (FPN) in the backbone network to enhance tolerance to scale variations. An auxiliary refinement branch is employed to generate multi-scale positive samples in the form of object pyramids, refining predictions. This additional branch shares weights with the original Faster R-CNN, classifying the extracted object pyramids in the RPN and the detector head during training. To maintain scale-consistent predictions without introducing improper negatives, anchor matching rules are abandoned, and FPN stage and spatial locations are adaptively assigned to object pyramids as positives. MPSR achieves performance gains without adding extra weights during inference, making it efficient and deployable on various detectors.

In FSOD, during the fine-tuning process, the typical meta-learning pipeline pre-selects a limited amount of data (K-shot) for both base and novel classes, which can lead to significant class imbalance, potentially skewing the learning process towards the dominant class. To address this, Wu \textit{et al.}~\cite{td_sampler} proposed a new model named Training Difficulty Sampler (TD-SAmpler). The TD-Sampler draws inspiration from curriculum and self-paced learning, gradually increasing training difficulty by presenting more straightforward base class data. However, unlike traditional fixed-complexity curriculum learning,  the TD-Sampler dynamically adjusts data complexity during training. This ensures that the model focuses on informative base class examples and mitigates class imbalance, leading to better performance in FSOD tasks.

Yan \textit{et al.} \cite{unp} highlighted the issue of incomplete annotation while building the instance-level training benchmark. Missing annotations are regarded as background, resulting in erroneous training gradients back-propagated through the detector, thereby compromising the detection performance. Yan \textit{et al.} introduced UNP \cite{unp} consisting Confusing Proposals Separation (CPS) and Affinity-Driven Gradient Relaxation (ADGR). The CPS isolates confusing negative proposals by evaluating the IoU between proposals and annotations. After identifying confusing proposals, the ADGR reweights the gradients based on the affinity between these proposals and class prototypes, dynamically allocating different optimization coefficients to improve detection performance.

Wang \textit{et al.} \cite{snida}  proposed a data augmentation technique known as SNIDA, which separates the foreground and background in order to enhance their diversity. Unlike earlier augmentation methods, SNIDA utilizes semantic-guided non-linear transformation spaces to augment training data, addressing the limitations of previous techniques that often lacked diversity and resulted in overfitting. SNIDA demonstrates improved adaptability to diverse category shapes and displays enhanced semantic awareness.

\subsubsection{Class Margin and Knowledge Transfer:}
This section explores approaches that enhance class boundaries and utilize knowledge transfer mechanisms for better adaptation to new data. Knowledge transfer allows the model to leverage previously learned information from known classes to recognize and detect new, unseen classes. This significantly improves the model's performance and efficiency by reducing the need for large amounts of labeled data for new classes and enhancing adaptability in dynamic environments.

Chen \textit{et al.} \cite{lstd} is the first to propose a work in the area of few-shot object detection. In their work, they proposed LSTD \cite{lstd}, a knowledge transfer-based FSOD approach. They use a transfer knowledge (TK) module that transfers the source object-label knowledge for each target-domain proposal to generalize low-shot learning in the target domain.

Li \textit{et al.} \cite{BMM-CME} propose a Class Margin Equilibrium (CME) approach for the FSOD task, aiming to optimize feature space partitioning and novel class reconstruction. CME tackles the challenge of balancing category-agnostic and category-specific components within a CNN-based detection model.

Kim \textit{et al.} \cite{FSOD-KT} introduced a new FSOD approach that leverages knowledge transfer from base classes to detect objects with few training examples for novel classes. Their method utilizes a prototype matching network and RoI feature alignment to facilitate effective few-shot adaptation, improving detection performance with limited training examples.

Wu \textit{et al.} \cite{svd} explored the use of Singular Value Decomposition (SVD) to enhance both the generalization and discrimination abilities of few-shot object detectors. They proposed the SVD-Dictionary enhancement method, which refines feature spaces based on sorted singular values, leading to improved detection accuracy.

Vu \textit{et al.} \cite{ford_bl} proposes a new FSOD framework called FORD+BL, which utilizes a Baby learning mechanism with multiple receptive fields. This mechanism effectively reuses previous knowledge in a novel domain by imitating a baby's learning process through visual cues, enhancing the model's adaptability and performance.

Li \textit{et al.} \cite{disentangle_and_remerge} identified that limited training data hinders the model's ability to explore semantic information fully. To address this, they proposed a knowledge distillation-based FSOD approach to leverage semantic information from large-scale pre-trained models. They developed a Structural Causal Model (SCM) to understand the learning process of knowledge distillation in FSOD from a causal perspective. Guided by this SCM, they introduced the ``Disentangle and Remerge" method, implementing knowledge distillation with backdoor adjustment to enhance the acquisition of semantic information.

Zhu \textit{et al.} \cite{data_augmentation_and_distribution_calibration} addressed two significant challenges: i) The limited availability of extreme samples exacerbates bias in proposal distribution, hindering the adaptation of ROI heads to new categories, ii) The scarcity of samples in novel categories makes the RPN a primary source of classification errors, leading to a decline in detection performance for these categories. To address these challenges, Zhu \textit{et al.} proposed a knowledge transfer method based on distributed calibration and data augmentation. Their model aligns the skewed distributions of novel categories with the foundational category distributions. It uses a drift compensation strategy to mitigate the adverse effects of classifying new categories during the fine-tuning phase. 
Moreover, the domain-aware data augmentation technique addresses data scarcity concerns by leveraging inter-image foreground-background blends, thereby enhancing the diversity and coherence of augmented datasets.

Wang \textit{et al.} \cite{ckdt} addressed the drawbacks of previous methods that rely on shared parameters for implicit transfer knowledge without explicit induction. This often leads to novel-class representations that are easily confused with similar base classes and are poorly suited to diverse patterns of variation in the truth distribution. To tackle this, Wang \textit{et al.} proposed an inter-class Correlation and intra-class Diversity based Knowledge Transfer (CDKT) method for FSOD. They designed a graph that dynamically captures the relationship between base and novel class representations. Then, they introduced distillation techniques to overcome the lack of correlation knowledge in few-shot labels. Additionally, they proposed a diversity knowledge transfer module based on the data hallucination, which adaptively disentangles class-independent variation patterns from base-class features and generates additional trainable hallucinated instances for novel classes.

\begin{table*}
\caption{Comparison between different FSOD approaches. We highlight the merits and demerits of each FSOD approach} \label{tab:s-fsod_sub}
\vspace{-1em}
  \resizebox{\textwidth}{!}
  { \footnotesize
 \begin{tabular} 
 {p{3cm}p{2cm}p{5.5cm}p{5cm}}
\hline \hline
\textbf{Methods} & \textbf{References} & \textbf{Merits} & \textbf{Demerits} \\ \hline \hline
Meta Learning & \cite{YOLO-FR, Meta-Det, Meta-RCNN, meta_retina_net, meta_detr, meta_faster_rcnn, apsp, SQMGH, aht, alpha-adaptive} & 
\begin{itemize}[leftmargin=0.2cm,noitemsep,topsep=2pt,
                    before = \tablistcommand,
                    after  = \tablistcommand]
    \item Good generalization to detect novel object classes.
    \item Easier adaptation to novel classes.
    \end{itemize} & 
    \begin{itemize}[leftmargin=0.2cm,noitemsep,topsep=2pt,
                    before = \tablistcommand,
                    after  = \tablistcommand]
    \item Can lead to significant class imbalance.
    \item Can be computationally intensive and complex
    \end{itemize} 
    \\
\rowcolor{lightmintbg} \rule{-2pt}{10pt} Metric Learning and Classification Refinement & \cite{RepMet, dmnet, srr_fsd, cdgp, memfrcn, wfs_detr} & 
\begin{itemize}[leftmargin=0.2cm,noitemsep,topsep=2pt,
                    before = \tablistcommand,
                    after  = \tablistcommand]
    \item Makes the embedding space semantically meaningful, such that objects from the same category are close. 
    \item Classification refinement enhances decision boundaries, leading to more distinct and accurate classifications.
    \end{itemize} & 
    \begin{itemize}[leftmargin=0.2cm,noitemsep,topsep=2pt,
                    before = \tablistcommand,
                    after  = \tablistcommand]
    \item Limited to classification head.
    \end{itemize} 
    \\
Proposal Generation and Quality Improvement & \cite{CoRPN, Halluc, fadi, lvc, fct, mrsn, crted, fm_fsod} & 
\begin{itemize}[leftmargin=0.2cm,noitemsep,topsep=2pt,
                    before = \tablistcommand,
                    after  = \tablistcommand]
    \item Provides high localization accuracy 
    \item Improved proposal quality can help in detecting objects even in challenging conditions, such as occlusion and cluttered scenes
    \end{itemize} & 
\begin{itemize}[leftmargin=0.2cm,noitemsep,topsep=2pt,
                    before = \tablistcommand,
                    after  = \tablistcommand]
    \item Region-Proposal-Network is prone to classes with few examples, as if it misses a even one high IOU training box, it will drastically affect the classifier. 
    \item Highly sensitive to training data
    \end{itemize} 
    \\
\rowcolor{lightmintbg} \rule{-2pt}{10pt} Attention Mechanisms and Feature Enhancement & \cite{AttFDNet, fsce, dgfidacl, norm_vae, ardnet, tide, fssp, fsna, sparse_ct, fpd} & 
\begin{itemize}[leftmargin=0.2cm,noitemsep,topsep=2pt,
                    before = \tablistcommand,
                    after  = \tablistcommand]
    \item Provides prior knowledge about salient regions. 
    \item Extracts more informative and discriminative features from the limited training data
    \end{itemize} 
    & 
\begin{itemize}[leftmargin=0.2cm,noitemsep,topsep=2pt,
                    before = \tablistcommand,
                    after  = \tablistcommand]
    \item High computational cost
\end{itemize}
\\
Data Sampling and Scale Variation & \cite{mpsr, td_sampler, unp, snida} & \begin{itemize}[leftmargin=0.2cm,noitemsep,topsep=2pt,
                    before = \tablistcommand,
                    after  = \tablistcommand]
    \item Ensure tolerance to scale variation 
    \item Sampling data from a wide range of scales and variations prevents the model from overfitting to specific instances or features.
    \end{itemize} & 
    \begin{itemize}[leftmargin=0.2cm,noitemsep,topsep=2pt,
                    before = \tablistcommand,
                    after  = \tablistcommand]
    \item Limited to Feature Pyramidal Networks (FPN) 
    \end{itemize} \\
\rowcolor{lightmintbg} \rule{-2pt}{10pt} Class Margin and Knowledge Transfer & \cite{BMM-CME, ford_bl, FSOD-KT, svd, disentangle_and_remerge, data_augmentation_and_distribution_calibration, ckdt} & \begin{itemize}[leftmargin=0.2cm,noitemsep,topsep=2pt,
                    before = \tablistcommand,
                    after  = \tablistcommand]
    \item Addresses the tradeoff between better representation and better classification of novel classes, and achieves an equilibrium.
    \item Better leveraging of knowledge from the base classes to detect objects of novel classed with few examples.
\end{itemize} & 
\begin{itemize}[leftmargin=0.2cm,noitemsep,topsep=2pt,
                    before = \tablistcommand,
                    after  = \tablistcommand]
    \item Limited to improvement to object classification and cannot be extended to localization 
    \item Highly sensitive to outliers
    \end{itemize}
    \\
\hline
\end{tabular}
}\vspace{-1em}
\end{table*}

\subsubsection{Conclusion:}
Standard FSOD enables models to generalize using a small amount of training data, which is particularly valuable when it is difficult or expensive to acquire large annotated datasets. These methods are categorized based on their techniques, each with its strengths and weaknesses, as outlined in Table~\ref{tab:s-fsod_sub}. Meta-learning-based approaches offer high generalization in detecting novel classes but can lead to significant class imbalance. Metric learning-based approaches emphasize the importance of semantically meaningful class boundaries. Proposal generation-based approaches achieve high localization accuracy but are vulnerable with few examples, where missing a single high IoU training box can impact performance. Attention-mechanism-based approaches benefit from attention formulations in detecting salient objects. Scale variation-based methods emphasize tolerance to scale variations. Class-margin-based approaches balance novel class classification and representation in the same feature space. Knowledge transfer-based methods leverage knowledge from base classes to improve the localization of novel classes.

\subsection{Generalized Few Shot Object Detection (G-FSOD)}
\label{subsec:gfsod_survey}
G-FSOD training follows a similar process to standard FSOD, starting with a large dataset from specific classes and then fine-tuning on a new set with limited samples. However, G-FSOD also focuses on preserving the object detection model's performance on previously learned classes while incorporating new ones. This is essential for practical applications where learning new classes should not lead to forgetting old ones. This section reviews various approaches to G-FSOD.

Wang \textit{et al.} \cite{tfa} introduced TFA, a two-stage fine-tuning approach, as one of the pioneering methods for G-FSOD. TFA fine-tunes the detector using a balanced base and novel class sample set while keeping the backbone and RPN frozen. Previous research has highlighted meta-learning as a promising approach for FSOD, but fine-tuning techniques have received less attention. Wang \textit{et al.} found that fine-tuning only the last layer of existing detectors for rare classes is crucial for FSOD. The TFA approach involves training the entire detector on base classes and then fine-tuning the last layers on a balanced set of base and novel classes, which outperforms meta-learning methods. 

PNPDet or Plug-and-Play Detector \cite{pnpdet} is a novel approach for efficient few-shot detection without forgetting. It introduces a simple yet effective architecture with separate sub-networks for recognizing base and novel categories, preventing performance degradation on known categories while learning new concepts. Distance metric learning is further incorporated into sub-networks, consistently enhancing detection performance for base and novel categories.

In \cite{retentive_rcnn}, Fan \textit{et al.} addressed two neglected properties in prior works: 1) the pre-trained base class detector does not predict many false positives on novel class instances despite their saliency, and 2) the RPN is biased towards its seen classes instead of being ideally class-agnostic, thus freezing it without exposure to new classes can be sub-optimal. To tackle this, Fan \textit{et al.} introduced a Retentive R-CNN method, which combines base and novel class detectors using a Bias-Balanced RPN and a Re-detector. The Bias-Balanced RPN adapts better to novel classes while remaining effective on base classes, and the Re-detector uses a consistency loss to regularize adaptation during fine-tuning, allowing incremental detection without forgetting.

Guirguis \textit{et al.} \cite{cfa} introduced CFA to alleviate catastrophic forgetting. It adopts a continual learning method, namely Average Gradient Episodic Memory (A-GEM), to the task of G-FSOD. CFA introduces a new gradient update rule to address the risk of catastrophic forgetting in A-GEM when the angle between the loss gradient vectors of previous tasks and the proposed gradient update for the current tasks is obtuse. Therefore, it tries to minimize the angle rather than projecting the novel gradient orthogonally in case of violation.

NIFF \cite{niff} proposed the first data-free knowledge distillation approach that leverages the statistics of the RoI features from the base model to forge instance-level features without accessing the base images. It shows that the statistics of instance-level RoI head features adequately represent the distribution of base classes. It also shows that a standalone lightweight generator can be trained in a distillation fashion to match the gathered statistics and synthesize class-wise base features.

Ma \textit{et al.} \cite{digeo} introduced a model called DiGeo that learns geometry-aware features of inter-class separation and intra-class compactness. An offline simplex equiangular tight frame (ETF) classifier guides the separation of feature clusters, with weights serving as maximally and equally separated class centers. To tighten clusters for each class, DiGeo incorporates adaptive class-specific margins into the classification loss, encouraging features to stay close to class centers.

Guirguis \textit{et al.} \cite{uppr} also proposed the Uncertainty-based Progressive Proposal Refinement (UPPR) approach, which leverages uncertainty estimation to enhance object proposals, improving overall detection performance and reducing forgetting. It focuses explicitly on modeling predictive uncertainties, allowing for the refinement of object proposals, which enhances detection performance while mitigating the issue of forgetting by explicitly incorporating uncertainty modeling.

\subsubsection{Conclusion:}
G-FSOD is an improvement over standard FSOD in terms of retaining knowledge. The primary challenge in G-FSOD is knowledge retention. 
The TFA approach \cite{tfa} was one of the first to address G-FSOD, fine-tuning the detector with a balanced set of base and novel class samples while freezing the backbone and RPN. This method reduces forgetting but significantly drops performance in novel classes. Retentive R-CNN \cite{retentive_rcnn} uses a knowledge distillation approach to address catastrophic forgetting of base classes. However, these methods assume the availability of base data when learning new classes. NIFF \cite{niff} is the first data-free knowledge distillation method for G-FSOD, reducing the memory footprint by not storing base data. The most recent work, DiGeo \cite{digeo}, enhances knowledge retention and generalization to novel classes by learning discriminative features through inter-class separation and intra-class compactness.

\subsection{Incremental Few Shot Object Detection (I-FSOD)}
\label{subsec:ifsod_survey}
Both standard FSOD and G-FSOD tasks require base class data during fine-tuning on novel classes, which can be impractical due to the unavailability of base class data. I-FSOD eliminates this constraint by not requiring base class data during fine-tuning, making it more practical for real-world applications with its superior knowledge retention. I-FSOD can scale to accommodate more object classes over time, integrating new classes without compromising performance on existing ones. Therefore, I-FSOD aims to learn novel classes while avoiding catastrophic forgetting of base classes. This subsection outlines various approaches for the I-FSOD task.

The I-FSOD task was first introduced by Perez-Rua \textit{et al.}~\cite{once} with the OpeNended CentrenEt (ONCE) model. ONCE allows the incremental registration of new classes with just a few examples by adapting the CentreNet detector to few-shot learning and using meta-learning to create class-specific codes for new class registration. This model demonstrated the ability to incrementally register novel classes with minimal examples in a feed-forward manner without revisiting the training data for base classes. 

In \cite{sylph}, Yin \textit{et al.} proposed an accurate and flexible framework called SYLPH which decouples the object classification from localization and leverages base detectors pre-trained for class-agnostic localization. Yin \textit{et al.} highlighted the effectiveness of finetune-free I-FOSD, especially when a large number of base categories with abundant data are available for meta-training.

Ganea \textit{et al.}~\cite{imfta} introduced iMFTA as an incremental approach to address few-shot object detection and instance segmentation. It utilizes discriminative embeddings for object instances, which are then consolidated into class representatives and effectively addresses the issue of memory overhead. The iMFTA model matches class embeddings at the RoI level using cosine similarity, allowing to process new classes without additional training or access to prior training data.

LEAST~\cite{least}, introduced by Li \textit{et al.}, adapts models to novel objects using a few annotated samples while retaining previously learned information. It achieves this with minimal forgetting, reduced training resources, and enhanced transfer capability. It incorporates a transfer strategy to minimize unnecessary weight adjustments, employs knowledge distillation techniques, and utilizes a clustering-based exemplar selection process. 

iFSRCNN \cite{ifsrcnn} extends the Mask-RCNN framework for I-FSOD by introducing a new object class classifier based on the probit function and an uncertainty-guided bounding-box predictor. It leverages Bayesian learning to address the scarcity of training examples for new classes and estimates the uncertainty of predictions for bounding box refinement.

Cheng \textit{et al.} \cite{meta_ifsod} proposed a CenterNet framework-based model called Meta-iFSOD which employs meta-learning to adapt the model to new classes while mitigating forgetting. A meta-learner is trained with base-class samples, providing a good weight initialization for the object locator. During finetuning with novel-class samples, filters correlated to base classes are preserved.

Dong \textit{et al.} \cite{idetr} introduced a new model called Incremental-DETR (iDETR) for I-FSOD task using finetuning and self-supervised learning on the DETR object detector. It uses finetuning and self-supervised learning on the DETR object detector to detect novel classes without forgetting the base classes. The class-specific components of DETR are finetuned with self-supervision, and knowledge distillation is applied to encourage the network to detect novel classes.

Choi \textit{et al.} \cite{ifsod_sft} proposed a simple finetuning strategy-based approach called Incremental Two-stage Finetuning Approach (iTFA). The iTFA model consists of three steps: training the base model with abundant base classes, separating the RoI feature extractor and classifier into base and novel class branches, and finetuning the novel branch with only a few novel class examples.

Zhang \textit{et al.} \cite{sc_awg} introduced a model called SC AWG, which tackles I-FSOD by reorganizing base weights using the responses of novel region features to transfer learned information from base weights to relevant novel weights. It introduces two strategies, scale-aware and centerness-aware, to obtain representative region features for generating novel weights. The scale-aware strategy adapts to objects of different sizes, while the centerness-aware strategy focuses on objects' central regions, enhancing the effectiveness of novel class detection.

Tang \textit{et al.} \cite{non-registrable_weights} proposes using novel-registrable weights for RoI classification, which memorizes class-specific weights to alleviate forgetting old knowledge and registers new weights for novel classes. They also introduce region-level contrastive learning during the base training stage by augmenting proposal boxes. This approach enhances the generalizability of feature representations and the plasticity of the detector, allowing it to better adapt to new classes.

WS-iFSD \cite{ws_ifsd} is introduced as a weakly supervised method that significantly enhances the meta-training of the hyper-network by augmenting it with many weakly localized objects from a much larger set of object categories. Employing an off-the-shelf object localization model increases the number of classes and images used in training. This approach allows the hyper-network to generate significantly improved class codes for novel categories, demonstrating that weak supervision can effectively enhance meta-learning for few-shot detection.

\subsubsection{Conclusion:}
In contrast to standard FSOD and G-FSOD, the I-FSOD does not require old class data during fine-tuning, making it more practical for real-world scenarios. The I-FSOD offers better knowledge retention and scalability, seamlessly incorporating new classes without sacrificing performance on existing ones. The task was first introduced by ONCE \cite{once}, which enabled detectors to incrementally recognize novel categories while still detecting base category objects. Then various methods have been proposed \cite{sylph, imfta, least, ifsrcnn, meta_ifsod, idetr, ifsod_sft, sc_awg, non-registrable_weights} to enhance the performance of I-FSOD further. The most recent work WS-iFSD \cite{ws_ifsd} addresses poor generalization to novel categories and inferior performance on base categories. It also addresses the issue of inferior detection performance for base categories by freezing the backbone and detection head during meta-training.

\subsection{Open-Set Few Shot Object Detection (O-FSOD)}
\label{subsec:ofsod_survey}
O-FSOD addresses scenarios where objects of unknown classes exist. It extends the capabilities of traditional FSOD by enabling models to detect and classify objects of unknown categories not seen during training. This ability is crucial for real-world applications where new object categories may emerge over time. It has numerous practical applications in areas such as autonomous driving, surveillance, and robotics, where detecting novel objects in uncontrolled environments is essential. The primary objective of the O-FSOD task is to train the detector similar to the I-FSOD setting, but during inference, it detects the known classes (i.e., base classes and novel classes) as well as the unknown classes. In this subsection, we provide an in-depth survey of the O-FSOD approaches.

Binyi \textit{et al.} \cite{food} proposed a model called FOOD that incorporates classifier placeholders to handle novel classes alongside an unknown class and integrates a Class Weight Sparsification Module (CWSC) to mitigate overfitting. During the training phase, the CWSC reduces the mutual dependency between individual classes and their neighboring counterparts. This enhances the model's ability to generalize, especially in the context of open-set detection within few-shot scenarios. Additionally, the Unknown Decoupling Learner (UDL) is introduced, which aids the model in defining a concise decision boundary for unknown instances without relying on pseudo-unknown samples and addresses the challenge of accurately representing unknown data distributions.

In another work, Binyi \textit{et al.} \cite{hsic} proposed a new few-shot open-set object detector using a HSIC-based moving weight averaging technique. They also employed a novel approach for mining new unknown samples based on evidential uncertainty estimation to augment the training dataset. Furthermore, by considering localization quality, they introduced an IoU-aware unknown training objective to improve the decision boundary between known and pseudo-unknown data.

Liu \textit{et al.} \cite{foxod} introduced a model called FOXOD, which typically looks into O-FSOD for X-ray hazard detection, where extracting weak features from X-ray images is more challenging than natural scenes. The FOXOD model comprises four novel elements: the Overlapping Object Separator (OOS), the Unknown Interest Advisor (UIA), Knowledge Augmentation (KA), and the Discriminant Classifier (DC). To address challenges arising from high object overlap, OOS employs a hierarchical attention mechanism for sampling local knowledge, thereby disentangling features of overlapping individual targets. The UIA module helps Region Proposal Networks (RPNs) identify potential unknown proposals without relying on explicit supervision. While the KA module mitigates the catastrophic forgetting of known classes by the model, the DC module plays a crucial role in preventing overfitting and discerning unknown positive objects amidst many background proposals.

Chen \textit{et al.} \cite{ofdet} introduced a new Open World Few-Shot Object Detection (OFDet) model to detect unknown objects with only a few examples accurately. It includes three modules: the Class-agnostic Localization Module (CALM) for generating class-agnostic proposals and localizing potential unknown proposals, the Base Classification Module (BCM) for classifying objects from class features, and the Novel Detection Module (NDM) for detecting novel objects.

\subsubsection{Conclusion:}
O-FSOD extends traditional FSOD by enabling models to detect and classify novel objects not seen during training. This is crucial for real-world applications where new object categories can emerge over time, and detecting novel objects in uncontrolled environments is essential. This task was initially proposed by Binyi \textit{et. at.} \cite{g_food, food} with a method leveraging decoupling optimization and sparsification for few-shot unknown rejection, though it lacks generalization and poses unsatisfactory performance for real-world applications. Subsequent work by Binyi\textit{et al.} \cite{hsic} addressed the overfitting issue of the previous work and provided a solution in the weight space using a moving weight averaging method. Recently, OFDet \cite{ofdet} has been built upon a class-agnostic object detector under the two-stage fine-tuning paradigm, introducing an unknown proposals selection algorithm to select more accurate unknown objects. 

\subsection{Few Shot Domain Adaptive Object Detection (FSDAOD)}
\label{subsec:fsdaod_survey}
FSDAOD combines few-shot learning and domain adaptation to detect objects in environments with domain shifts and limited labeled data. FSDAOD models are designed to be robust across different domains, which is crucial for real-world applications where training and testing data differ. This approach is vital in areas like autonomous driving, surveillance, and robotics, where adapting to new environments and detecting objects with limited labeled data is essential. In FSDAOD, the model is trained on a source dataset and then fine-tuned on a target dataset with the same classes but a different distribution. The goal is to adapt the detector's capability to a new domain. Here, we will discuss FSDAOD approaches in detail. 

Fu \textit{et al.} \cite{cd-vito} introduced CD-ViTO, a model that integrates key elements into the DE-ViT network \cite{de_vit}, focusing on fine-tuning with limited target images. Recognizing the benefits of fine-tuning when there is a domain gap between source and target datasets, CD-ViTO fine-tunes the detection and classification heads using a small set of labeled support images. The model also employs learnable prototypes to enhance generalization and an attention module to allocate weights to instances, improving overall performance dynamically.

Gao \textit{et al.} \cite{asyfod} proposed an asymmetric adaptation paradigm called AsyFOD to tackle the issue of extreme data imbalance between source and target instances. AsyFOD begins by dividing the source instance set into two parts: target-similar and target-dissimilar instance sets. The target-similar source instances are identified using a unified discrepancy estimation function, which helps to augment the limited target instances and reduce data imbalance. The remaining source instances are classified as target-dissimilar. To further address domain-based data imbalances, AsyFOD employs asynchronous alignment between the target-dissimilar source instances and the augmented target instances, ensuring more effective adaptation to the target domain.

In another work, Gao \textit{et al.} \cite{acrofod} proposed AcroFOD, a model that employs an adaptive distribution optimization strategy to eliminate unsuitable, low-quality images. This strategy enables the detector to align more effectively with the feature distribution of the target domain, enhancing both speed and accuracy. The adaptive optimization strategy emphasizes the importance of augmentation quality and prevents the model from over-adaptation, akin to overfitting due to insufficient data. Additionally, to enrich the diversity of merged images from both source and target domains, AcroFOD formulates generalized frameworks for multi-level domain-aware augmentation. This approach ensures that the model can handle a variety of domain shifts and maintain high performance.

Wang \textit{et al.} \cite{fs_adaptive_frcnn} addresses the challenges of insufficient target domain data and over-adaptation, which can lead to instability and degraded detection performance in the target domain. To mitigate these issues, they introduce a pairing mechanism that aligns source and target features, helping to compensate for the lack of target domain samples. They also propose a bi-level adaptation module for the source-trained detector. The first level is a split pooling-based image-level adaptation module, which uniformly extracts and aligns paired local patch features across different locations, scales, and aspect ratios. The second level is an instance-level adaptation module that semantically aligns paired object features, ensuring accurate detection while avoiding inter-class confusion. This dual approach helps maintain robust detection performance despite the domain shift.

In \cite{pica}, Zhong \textit{et al.} introduced PICA, a model that extends point-wise alignment from classification to object detection. PICA leverages moving average centroids to address label noise in background Regions of Interest (ROIs). It computes a moving average centroid for each category and excludes background ROIs, retaining only the centroid for background instances, effectively reducing label noise. Additionally, PICA uses point-wise alignment across instances and centroids to tackle the challenge of limited labeled target instances.

Yan \textit{et al.} \cite{adaptive_meta_rcnn} introduced an innovative approach in their work by adding an image domain classifier to the Meta CNN framework, placed after the backbone's last layer, to reduce domain discrepancy. Additionally, to avoid confusion between class features caused by the alignment of image feature distributions, they incorporated a feature filter module called CAFFM (Class-Aware Feature Filter Module), which specifically filters out features irrelevant to particular classes, ensuring that only the most pertinent features are used for classification. This dual approach of domain classification and targeted feature filtering enhances the model's ability to adapt to new domains while maintaining accurate class-specific feature representation.

Nakamura \textit{et al.} \cite{cutmix} propose a data synthesis technique to address significant domain discrepancies. They leverage data augmentation to mix features from domains with significant gaps by pasting a part of one domain’s image onto another. Specifically, for object detection tasks, where detection targets are typically smaller than the background, they cut out the detection objects and paste them onto images from another domain. This targeted approach to domain adaptation helps bridge the gap between different data distributions effectively.

\subsubsection{Conclusion:}
FSDAOD merges few-shot learning and domain adaptation to tackle object detection in environments with domain shifts and limited labeled data. FSDAOD models are robust to domain shifts, enabling them to generalize well across different domains. This capability is crucial for applications like autonomous driving, surveillance, and robotics, where training and testing data differ. In this field, Wang \textit{et al.} \cite{fs_adaptive_frcnn} adopted a pairing mechanism that aligns source and target samples at multiple levels and employs a bi-level module to adapt the source-trained detector to the target domain. PICA \cite{pica}, on the other hand, employed point-wise alignment in the context of FSOD over instances and centroids to address the scarcity of labeled target instances. 
Gao \textit{et al.} \cite{acrofod} addressed insufficient target domain data by selecting augmented data similar to target samples, demonstrating the effectiveness of domain-mix augmentation to overcome domain shifts. AsyFOD \cite{asyfod} mitigates data imbalance by using target-similar source instances to augment limited target instances and aligning target-dissimilar source instances to avoid over-adaptation. Nakamura \textit{et al.} \cite{cutmix} used a data synthesis method to solve the large domain gap problem, where a part of the target image is pasted onto the source image, and the position of the pasted region is aligned by utilizing the information of the object bounding box.

\section{Datasets, Evaluation Protocols}\label{sub-sec:dataset}
\label{sec:datasets-evaluation}
This section provides the details of the dataset being used in different FSOD settings, i.e., standard FSOD, G-FSOD, I-FSOD, O-FSOD and FSDAOD tasks along with the details of corresponding evaluation protocols.

\subsection{Dataset details} 
\label{subsec: dataset}
The standard FSOD, G-FSOD, I-FSOD and O-FSOD tasks commonly use two main benchmark datasets, i.e., Pascal VOC \cite{pascal_voc} and MS COCO \cite{coco}, for performance evaluation. Additionally, some studies have chosen to utilize FSOD \cite{Attention-RPN}, ImageNet-LOC \cite{imagenet-loc}, and LVIS \cite{lvis} datasets for assessing their approach. Several O-FSOD methods have used SIXray \cite{sixray} and PIDray \cite{pidray} datasets for evaluation purposes. In the evaluation of FSDAOD methods, researchers have employed Cityscapes \cite{cityscapes}, Foggy Cityscapes \cite{foggy_cityscapes}, KITTY \cite{kitti}, and Sim10K \cite{sim10k} datasets. Details of these datasets are discribed below:
\begin{table}[t!]
\centering
\caption{Details of the common benchmarks used in FSOD, G-FSOD, I-FSOD, O-FSOD and FSDAOD methods.}
\vspace{-1em}
\begin{adjustbox}{width=\textwidth}
\begin{tabular}{lllccccc}
\hline \hline
\rule{-2pt}{10pt}                               &                             &       & \multicolumn{3}{c}{PASCAL} & \multirow{2}{*}{COCO} & \multirow{2}{*}{LVIS} \\ \cline{4-6}
                               &                             &       & Split1  & Split2  & Split3 &                       &                       \\
                               \hline \hline
  \multicolumn{2}{c}{\multirow{2}{*}{$\#$classes}}       & \rule{-2pt}{10pt}  Base task  &    \multicolumn{1}{c}{15}    &    \multicolumn{1}{c}{ 15}    &   \multicolumn{1}{c}{   15 } &        \multicolumn{1}{c}{ 60  }            &     \multicolumn{1}{c}{  776   }             \\
\multicolumn{2}{l}{}                                         & Novel task & \multicolumn{1}{c}{ 5 }      &  \multicolumn{1}{c}{ 5}      &   \multicolumn{1}{c}{5  }   &                \multicolumn{1}{c}{ 20   }   &           \multicolumn{1}{c}{  454}          \\
\hline
   \multicolumn{2}{c}{\multirow{2}{*}{Base training}} &  \multicolumn{1}{l}{\rule{-2pt}{10pt} $\#$Images}          &     \multicolumn{1}{c}{14,631}   &     \multicolumn{1}{c}{14,779}     &     \multicolumn{1}{c}{14,318}    &                   \multicolumn{1}{c}{98,459}    &             \multicolumn{1}{c}{68,568 }          \\
                    \multicolumn{2}{l}{}            & \multicolumn{1}{l}{$\#$bboxs}           &   \multicolumn{1}{c}{41,084}      &     \multicolumn{1}{c}{40,397}     &   \multicolumn{1}{c}{40,511}     &    \multicolumn{1}{c}{367,702 }                   &        \multicolumn{1}{c}{688,029}               \\ \hline
\multirow{4}{*}{FSOD}          & \multirow{2}{*}{Training ($\#$shots / $\#$bboxs)}   & Base classes  &    \multicolumn{1}{c}{1-10 / 15-150}      &   \multicolumn{1}{c}{1-10 / 15-150}       &  \multicolumn{1}{c}{1-10 / 15-150}        &       \multicolumn{1}{c}{1-30 / 60-1800}                 &          \multicolumn{1}{c}{8.57 /  7,760}              \\
                               &                             & Novel classes &  \multicolumn{1}{c}{1-10 / 5 - 50}        &    \multicolumn{1}{c}{1-10 / 5 - 50}      &   \multicolumn{1}{c}{1-10 / 5 - 50}      &         \multicolumn{1}{c}{1 -30/ 20 - 600}               &    \multicolumn{1}{c}{8.57 / 2,786}                    \\
                               & \multirow{2}{*}{Evaluation ($\#$bboxs)} & Base classes  &   \multicolumn{1}{c}{-}       &    \multicolumn{1}{c}{-}       &    \multicolumn{1}{c}{-}     &       \multicolumn{1}{c}{-}                 &       \multicolumn{1}{c}{-}                 \\
                               &                             & Novel classes &         \multicolumn{1}{c}{1,924} &       \multicolumn{1}{c}{2,088}    &    \multicolumn{1}{c}{1,839}     &   \multicolumn{1}{c}{20,193}                     &    \multicolumn{1}{c}{429}   \\
                                \hline
\multirow{4}{*}{G-FSOD}          & \multirow{2}{*}{Training ($\#$shots / $\#$bboxs)}   & Base classes  &    \multicolumn{1}{c}{1-10 / 15-150}      &   \multicolumn{1}{c}{1-10 / 15-150}       &  \multicolumn{1}{c}{1-10 / 15-150}        &       \multicolumn{1}{c}{1-30 / 60-1800}                 &          \multicolumn{1}{c}{8.57 /  7,760}              \\
                               &                             & Novel classes &  \multicolumn{1}{c}{1-10 / 5 - 50}        &    \multicolumn{1}{c}{1-10 / 5 - 50}      &   \multicolumn{1}{c}{1-10 / 5 - 50}      &         \multicolumn{1}{c}{1 -30/ 20 - 600}               &    \multicolumn{1}{c}{8.57 / 2,786}                    \\
                               & \multirow{2}{*}{Evaluation ($\#$bboxs)} & Base classes  &   \multicolumn{1}{c}{13,052}       &    \multicolumn{1}{c}{12,888}       &    \multicolumn{1}{c}{13,137}     &       \multicolumn{1}{c}{15,318}                 &       \multicolumn{1}{c}{50,334}                 \\
                               &                             & Novel classes &         \multicolumn{1}{c}{1,924} &       \multicolumn{1}{c}{2,088}    &    \multicolumn{1}{c}{1,839}     &   \multicolumn{1}{c}{20,193}                     &    \multicolumn{1}{c}{429}   \\
                                \hline
\multirow{4}{*}{I-FSOD}          & \multirow{2}{*}{Training ($\#$shots / $\#$bboxs)}   & Base classes  &    \multicolumn{1}{c}{-}      &   \multicolumn{1}{c}{-}       &  \multicolumn{1}{c}{-}        &       \multicolumn{1}{c}{-}                 &          \multicolumn{1}{c}{-}              \\
                               &                             & Novel classes &  \multicolumn{1}{c}{1-10 / 5 - 50}        &    \multicolumn{1}{c}{1-10 / 5 - 50}      &   \multicolumn{1}{c}{1-10 / 5 - 50}      &         \multicolumn{1}{c}{1 -30/ 20 - 600}               &    \multicolumn{1}{c}{8.57 / 2,786}                    \\
                               & \multirow{2}{*}{Evaluation ($\#$bboxs)} & Base classes  &   \multicolumn{1}{c}{13,052}       &    \multicolumn{1}{c}{12,888}       &    \multicolumn{1}{c}{13,137}     &       \multicolumn{1}{c}{15,318}                 &       \multicolumn{1}{c}{50,334}                 \\
                               &                             & Novel classes &         \multicolumn{1}{c}{1,924} &       \multicolumn{1}{c}{2,088}    &    \multicolumn{1}{c}{1,839}     &   \multicolumn{1}{c}{20,193}                     &    \multicolumn{1}{c}{429}   \\ \hline
\end{tabular}
\end{adjustbox}

\resizebox{0.6\textwidth}{!}{
\begin{tabular}{lllcccc}
\hline \hline
\rule{-2pt}{10pt}      &         &                               & PASCAL & VOC-COCO& LVIS \\
\hline \hline
  \multirow{5}{*}{O-FSOD}       &  {\multirow{3}{*}{Training}}  &  Base Known Classes & \multicolumn{1}{c}{ 10  } &        \multicolumn{1}{c}{ 20 }            &     \multicolumn{1}{c}{  315 }             \\
&  & Novel Known Classes &  \multicolumn{1}{c}{5}   &                \multicolumn{1}{c}{ 20 }   &           \multicolumn{1}{c}{ 454 }          \\
&  & Shots &  \multicolumn{1}{c}{1,3,5,10}   &                \multicolumn{1}{c}{1,3,5,10}   &           \multicolumn{1}{c}{<10 (rare classes)}          \\
\cline{2-6}
  & \multirow{3}{*}{Evaluation} &
 Base Known Classes & \multicolumn{1}{c}{ 10  } &        \multicolumn{1}{c}{ 20 }            &     \multicolumn{1}{c}{  315 }             \\
 & & Novel Known Classes &  \multicolumn{1}{c}{5}   &                \multicolumn{1}{c}{ 20 }   &           \multicolumn{1}{c}{ 454 }          \\
 & & Unknown Classes &  \multicolumn{1}{c}{5}   &                \multicolumn{1}{c}{ 40 }   &           \multicolumn{1}{c}{ 461}          \\
\hline
\end{tabular}}
\centering
\begin{adjustbox}{width=\textwidth}
\begin{tabular}{lllcccc}
\hline \hline
\rule{-2pt}{10pt}      &         &                               & Cityscapes $\rightarrow$ Foggy Cityscapes & KITTI $\rightarrow$ Cityscapes & SIM10k $\rightarrow$ Cityscapes \\
\hline \hline
  \multirow{4}{*}{FSDAOD}       &  {\multirow{2}{*}{Training}}  &  Source Images/Classes & \multicolumn{1}{c}{ 2975/8 } &        \multicolumn{1}{c}{7481/1(car)}            &     \multicolumn{1}{c}{ 10000/1(car) }             \\
&  & Target Images/Classes &  \multicolumn{1}{c}{8/8}   &                \multicolumn{1}{c}{ 8/1(car) }   &           \multicolumn{1}{c}{8/1(car) }          \\ \cline{2-6}
 & \multirow{1}{*}{Evaluation} &
 Target Images/Classes &  \multicolumn{1}{c}{3475/8}   &                \multicolumn{1}{c}{/1(car) }   &           \multicolumn{1}{c}{ 3475/1(car)}          \\
\hline
\end{tabular}
\end{adjustbox}
\end{table}
\begin{itemize}[leftmargin=0.5cm,noitemsep,topsep=15pt,
                    before = \tablistcommand,
                    after  = \tablistcommand] 
    \item \textbf{PASCAL Visual Object Classes (VOC) \cite{pascal_voc}} is a widely recognized benchmark for traditional object detection. In the context of FSOD, the object categories are divided into 15 base classes and 5 novel classes, with three different base/novel splits: Split1, Split2, and Split3. During the base training phase, the model utilizes the training and validation (trainval) images from VOC2007 and VOC2012. A fixed subset of the VOC2007 and VOC2012 trainval sets is used to support the few-shot fine-tuning stage. Various few-shot settings, such as 1-shot, 2-shot, 3-shot, 5-shot, and 10-shot, correspond to different numbers of base and novel bounding boxes. The evaluation is performed on approximately 5,000 images from the VOC2007 test set in terms of mAP50, mAP75 and mAP50-95.
    \item \textbf{Microsoft Common Objects in COntext (MS COCO) \cite{coco}} is another prominent benchmark for object detection, which has been adapted for FSOD. Object categories are divided into 20 novel classes in this setting, shared with PASCAL VOC, and 60 base categories. A subset of 5,000 images from the COCO2014 validation dataset, referred to as val5k, is commonly used to evaluate FSOD models. Base training and few-shot finetuning utilize the remaining COCO2014 training and validation images (trainvalno5k). Evaluation metrics are presented separately for base and novel categories, including COCO-style mAP, mAP50, mAP75 calculated for small, medium and large objects.
    \item \textbf{LVIS \cite{lvis}}, a newer dataset in the field of object detection, features a vast array of 1,230 classes categorized as frequent, common (with 10 or more annotations), and rare (with fewer than 10 annotations). For FSOD, TFA\cite{tfa} proposed the use of LVIS v0.5, which was divided into 776 base classes (comprising frequent and common categories) and 454 novel classes. This benchmark offers significantly more categories than COCO, with 10 times the number of categories and 50 times more than PASCAL VOC dataset. The evaluation metrics include COCO-style mAP, mAP50, and mAP75, reported separately for frequent, common, and rare objects. Additionally, aggregated metrics are computed for all three categories combined.
    
    \item \textbf{ImageNet-LOC \cite{imagenet-loc}} contains 314 classes, mostly of animals and birds. For FSOD, this dataset has been used by RepMet\cite{RepMet}, which is divided into 100 seen classes and 214 unseen classes. The first 100 categories from this dataset (mostly animals and birds) are used as base classes. The remaining 214 classes are used as testing classes. The mAP is used as an evaluation metric, reported for 1, 5, and 10 shots.

    \item \textbf{FSOD \cite{Attention-RPN}} is built from existing large-scale object detection datasets i.e., ImageNet\cite{imagenet} and OpenImage \cite{v4_dataset}. It consolidates a label system from these datasets by merging the leaf labels in their original label trees, grouping those with the same semantics (e.g., ice bear and polar bear) into one category, and removing semantics that do not belong to any leaf categories. The total number of classes is 1000, with 531 from ImageNet\cite{imagenet} and 469 from the OpenImage dataset \cite{v4_dataset}. There are 800 classes, 52,350 images, and 147,489 boxes for training, while the testing set includes 200 classes, 14,152 images, and 35,102 boxes.
    
    \item \textbf{SIXRAY \cite{sixray}} has been utilized by Liu \textit{et al.} \cite{foxod} for the O-FSOD task. The dataset comprises 8,929 images collected from subway stations and includes six categories: gun, knife, wrench, pliers, scissors, and hammer. These instances vary in proportions, viewpoints, and overlapping scenarios, contributing significant noise interference to typical visual systems. Liu \textit{et al.} randomly divides the dataset into three splits, each containing three base classes, two novel classes, and one unknown class. Specifically, Split 1 includes guns and knives, Split 2 includes pliers and wrenches, and Split 3 includes scissors and guns as their novel classes.
    
    \item \textbf{PIDRAY \cite{pidray}} has also been used by Liu \textit{et al.} \cite{foxod} for the O-FSOD. This dataset encompasses a variety of scenarios for detecting prohibited items in real-world contexts, focusing on intentionally concealed objects. It includes 12 classes of prohibited items across 47,677 X-ray images: gun, bullet, knife, wrench, pliers, power bank, baton, lighter, sprayer, hammer, scissors, and handcuffs. The images are randomly divided into 6 base classes, 3 novel classes, and 3 unknown classes.
    
    \item \textbf{CITYSCAPES \cite{cityscapes}}, introduced by Cordtz \textit{et al.}, is a widely used dataset for the FSDAOD task. This dataset comprises a vast, diverse collection of stereo video sequences captured from streets in 50 cities. It includes 5,000 images with high-quality pixel-level annotations and an additional 20,000 images with coarse annotations, facilitating methods that utilize large volumes of weakly labeled data. Specifically, it contains 3,475 real urban images, with 2,975 images used for training and 500 images for validation in FSDAOD tasks.
    
    \item \textbf{FOGGY CITYSCAPES \cite{foggy_cityscapes}} is also commonly used dataset for the FSDAOD task. Sakaridis \textit{et al.} generated the Foggy CityScapes dataset by applying fog synthesis to the Cityscapes \cite{cityscapes} dataset, resulting in 20,550 images. The highest fog intensity of the eight classes is used in the FSDAOD task. Typically, FSDAOD models are adapted from those trained on the Cityscapes dataset to perform on the Foggy Cityscapes dataset.
    
    \item \textbf{KITTI \cite{kitti}}, introduced by Geiger \textit{et al.}, is also being used by few FSDAOD approaches \cite{asyfod, acrofod}. The dataset includes 389 pairs of stereo and optical flow images, stereo visual odometry sequences spanning 39.2 km, and over 200,000 3D object annotations from cluttered environments (with up to 15 cars and 30 pedestrians visible per image). It comprises 7,481 images of the car class, which are primarily used as the source class.
    
    \item \textbf{SIM10k \cite{sim10k}}, introduced by Johnson-Roberson \textit{et al.}, is a simulated dataset created from the video game ``Grand Theft Auto V". This dataset comprises 10,000 synthetic images featuring urban driving scenes. Each image is annotated with bounding boxes, resulting in a total of 58,701 car bounding boxes. By using high-quality synthetic data, SIM10k provides a valuable resource for training and evaluating object detection models in a controlled and diverse environment.
\end{itemize}

\afterpage{
{\scriptsize
\begin{longtable}[!t]{|p{2.1cm}|p{5.9cm}|p{4.8cm}|}
\caption{Summary of Standard FSOD, G-FSOD, I-FSOD, O-OFSOD and FSDAOD methods in terms of benchmark settings for training \& evaluation. The source code listing provides the link to the source code repository and the framework used.}\label{tab:b}
\vspace{-1em}
\\
\hline \hline
\textbf{Model} & \textbf{Benchmark Settings} & \textbf{Source Code Listing} \\
\hline \hline
\endfirsthead

\hline
\textbf{Model} & \textbf{Benchmark Settings} & \textbf{Source Code Listing} \\
\hline \hline
\endhead

\hline
\multicolumn{3}{r}{\footnotesize\textit{(Continued on next page)}} \\
\hline
\endfoot

\hline
\endlastfoot

\multicolumn{3}{|l|}{\textbf{\textcolor{brown}{Few Shot Object Detection Methods}}} \\ \hline
RepMet  \cite{RepMet} & \textbf{ImageNet-LOC}: 100 seen classes and 214 unseen classes. Number of shots are 1, 5, 10. & MXNet: github.com/jshtok/RepMet \\ \hline
\rowcolor{lightmintbg} Meta R-CNN  \cite{Meta-RCNN} & \textbf{VOC:} 3 novel/base splits with 1, 2, 3, 4, and 5 shots. \textbf{COCO:} 20 VOC-shared novel classes and 60 base classes with 10 and 30 shots. & PyTorch: github.com/yanxp/MetaR-CNN \\ \hline
YOLO-FR \cite{YOLO-FR}  & Same as Meta-RCNN & PyTorch: github.com/bingykang/Fewshot\_Detection \\ \hline
\rowcolor{lightmintbg} MetaDet \cite{Meta-Det}  & Same as Meta-RCNN & Not available \\ \hline
AttFDNet \cite{AttFDNet}  & Same as Meta-RCNN & github.com/chenxy99/AttFDNet \\ \hline
\rowcolor{lightmintbg} Zhang \textit{et al.} \cite{CoRPN}  & Same as Meta-RCNN & Not available \\ \hline
MPSR \cite{mpsr} & Same as Meta-RCNN & PyTorch: github.com/jiaxi-wu/MPSR \\ \hline
\rowcolor{lightmintbg} Meta-RetinaNet \cite{meta_retina_net} & Same as Meta-RCNN & Not available \\ \hline
SQMGH \cite{SQMGH} & Same as Meta-RCNN & Not available \\ \hline
\rowcolor{lightmintbg} Li \textit{et al.} \cite{cdgp} & Same as Meta-RCNN & Not available \\ \hline
Zhang et.al. \cite{Halluc} & Same as Meta-RCNN & github.com/pppplin/HallucFsDet \\ \hline
\rowcolor{lightmintbg} FCT \cite{fct} & \textbf{VOC}: Same as Meta-RCNN. \textbf{COCO}: Split same as Meta-RCNN. Number of shots are 1, 2, 3, 5, 10, 30. & PyTorch: github.com/GuangxingHan/FCT \\ \hline
SRR-FSD \cite{srr_fsd} & Same as Meta-RCNN & Not available \\ \hline
\rowcolor{lightmintbg} FADI \cite{fadi} & Same as Meta-RCNN & PyTorch: github.com/yhcao6/FADI \\ \hline
TD-Sampler \cite{td_sampler} & Same as Meta-RCNN & Not available \\ \hline
\rowcolor{lightmintbg} Meta-Faster-RCNN  \cite{meta_faster_rcnn} & Same as Meta-RCNN & PyTorch: github.com/GuangxingHan/Meta-Faster-R-CNN \\ \hline
FSCE \cite{fsce}  & Same as Meta-RCNN & PyTorch: github.com/megvii-research/FSCE \\ \hline
\rowcolor{lightmintbg} LVC  \cite{lvc} & Same as Meta-RCNN & PyTorch: github.com/prannaykaul/lvc \\ \hline
Meta-DETR  \cite{meta_detr} & \textbf{VOC}: Same as Meta RCNN \textbf{COCO}: Split same as Meta-RCNN. Number of shots are set to 1, 3, 5, 10, and 30. & PyTorch: github.com/ZhangGongjie/Meta-DETR \\ \hline
\rowcolor{lightmintbg} Lee \textit{et al.} \cite{apsp} & Same as Meta-DETR & Not available \\ \hline
MemFRCN \cite{memfrcn} & Same as Meta-RCNN & Not available \\ \hline
\rowcolor{lightmintbg} DMNet \cite{dmnet} & Same as Meta-RCNN & PyTorch: github.com/yrqs/DMNet \\ \hline
Demirel \textit{et al.} \cite{reinforce} & Same as Meta-RCNN & Not available \\ \hline
\rowcolor{lightmintbg} Norm-VAE \cite{norm_vae} & Same as Meta-RCNN & Not available \\ \hline
Du \textit{et al.} \cite{alpha-adaptive} & Same as Meta-RCNN & Not available \\ \hline
\rowcolor{lightmintbg} FSSP \cite{fssp} & Same as Meta-RCNN & Not available \\ \hline
Li \textit{et al.} \cite{disentangle_and_remerge} & Same as Meta-RCNN & PyTorch: github.com/ZYN-1101/DandR \\ \hline
\rowcolor{lightmintbg} Lee \textit{et al.} \cite{fsod_multi_domain} & Pretraining datasets include ImageNet\cite{imagenet}, COCO\cite{coco}, FSODD\cite{Attention-RPN}, LVIS\cite{lvis}, and Unified. The benchmark contains 10 datasets of 10 different domains: VisDrone\cite{visdrone}, DeepFruits\cite{deepfruits}, iWildCam\cite{iwildcam}, Clipart, iMaterialist\cite{imaterialist}, Oktoberfest\cite{oktoberfest}, LogoDet-3K\cite{logodet-3k}, CrowdHuman\cite{crowdhuman}, SIXray\cite{sixray}, KITTI\cite{kitti}. Number of shots are 1, 3, 5, 10. & PyTorch: github.com/amazon-science/few-shot-object-detection-benchmark \\ \hline
Fan \textit{et al.} \cite{model_calibration} & Same as Meta-RCNN & PyTorch: github.com/fanq15/FewX \\ \hline
\rowcolor{lightmintbg} MRSN \cite{mrsn} & Same as Meta-RCNN & https://github.com/MMatx/MRSN \\ \hline
AHT \cite{aht} & Same as Meta-RCNN & Not available \\ \hline
\rowcolor{lightmintbg} ARDNet \cite{ardnet} & \textbf{VOC}: Same as Meta-RCNN & Not available \\ \hline
Zhu \textit{et al.} \cite{data_augmentation_and_distribution_calibration} & \textbf{VOC}: Same as Meta-RCNN. \textbf{COCO}: 60 basic classes, 20 novel classes. Number of shots are 1, 2, 3. & Not available \\ \hline
\rowcolor{lightmintbg} FSOD V2 \cite{fsod_v2} & \textbf{ImageNet-LOC}: Same as RepMet. \textbf{COCO}: Same as Meta-RCNN. & Not available \\ \hline
Madan \textit{et al.} \cite{fsod_vision_lang_models} & \textbf{LVIS}: `Frequent' and `Common' classes are used for base training. `Rare' classes are used for novel finetuning. & Not available \\ \hline
\rowcolor{lightmintbg} Tide \cite{tide} & Trained on COCO 60 base classes, which are disjointed from VOC and the rest 20 classes are the novel categories. Evaluated both on COCO and on VOC images with 1, 2, 3, and 5 shots. & PyTorch: github.com/deku-0621/TIDE \\ \hline
ST-FSOD \cite{st_fsod} & \textbf{NWPU-VHR10 v2}: There are 7 base classes and 3 novel classes with shots of 3, 5, 10, and 20. \textbf{DIOR}: In the first setting, there are 15 base classes and 5 novel classes. In the second setting, there are 4 splits, each with 15 base classes and 5 novel classes. \textbf{iSAID}: The dataset has 15 categories with three splits: 1) 5 novel and 10 base classes, 2) 3 novel and 12 base classes, and 3) 5 novel and 10 base classes. The number of shots are 10, 50, and 100. & PyTorch: github.com/zhu-xlab/ST-FSOD \\ \hline
\rowcolor{lightmintbg} Yan \textit{et al.} \cite{unp} & Same as Meta-RCNN & PyTorch: github.com/Ybowei/UNP \\ \hline
DGFIDaCL \cite{dgfidacl} & \textbf{VOC}: Same as Meta-RCNN. \textbf{COCO}: Same as Meta-RCNN. Number of shots is fixed to 10. \textbf{Object365}: 15 selected disjoint classes from VOC and COCO as base classes. 5 other disjoint classes are novel classes. Number of shots: 1, 2, 3, 5, 10. & PyTorch: github.com/HuangLian126/DGFIDaC \\ \hline
\rowcolor{lightmintbg} Sun \textit{et al.} \cite{vpl} & \textbf{Hangzhou traffic dataset}: 3 novel classes, the rest 5 classes make up the base categories. Number of shots are 1, 2, 3, 5, 10. \textbf{GM traffic dataset}: 3 novel classes and the remaining 5 classes are the base classes. Number of shots are 1, 2, 3, 5, 10. \textbf{MSCOCO}: 9 categories in traffic scenarios. 3 novel classes. Remaining classes are used as base classes. Number of shots are 1, 2, 3, 5, 10. & Not available \\ \hline
WFS-DETR\cite{wfs_detr} & \textbf{ImageNetLoc-FS}: Total 331 classes. 101 base classes, 214 novel classes, 16 classes for validation. \textbf{CUB-200}: Total 200 classes. 100 base classes, 50 novel classes, 50 classes for validation. \textbf{VOC}: Total 20 classes. 15 base classes, 5 novel classes. & Not available \\ \hline
\rowcolor{lightmintbg} CKDT \cite{ckdt} & Same as Meta-RCNN & Not available \\ \hline
FPD \cite{fpd} & Same as Meta-RCNN & PyTorch: github.com/wangchen1801/FPD \\ \hline
\rowcolor{lightmintbg} SparseCT \cite{sparse_ct} & Same as Meta-RCNN & Not available \\ \hline
FSNA \cite{fsna} & Same as Meta-RCNN & Pytorch: github.com/FSNA2022/FSNA/ \\ \hline
\rowcolor{lightmintbg} CRTED \cite{crted} & Same as Meta-RCNN & Not available \\ \hline
SNIDA \cite{snida} & Same as Meta-RCNN & Not available \\ \hline
\rowcolor{lightmintbg} FM-FSOD \cite{fm_fsod} & Same as Meta-RCNN & Not available \\ \hline
\multicolumn{3}{l}{\textbf{\textcolor{brown}{Generalized Few Shot Object Detection Methods}}} \\ \hline \hline
TFA \cite{tfa} & Same as Meta-RCNN. \textbf{LVIS}: Training classes include the \textit{frequent} classes (>100 images), \textit{common} classes (10-100 images). Testing classes include the \textit{rare} classes (<10 images). & PyTorch(Official): github.com/ucbdrive/few-shot-object-detection \\ \hline
\rowcolor{lightmintbg} PNPDet \cite{pnpdet} & Same as Meta-RCNN & Not available \\ \hline 
Retentive RCNN \cite{retentive_rcnn} & Same as Meta-RCNN & PyTorch(Official): github.com/Megvii-BaseDetection/GFSD \\ \hline
CFA \cite{cfa} & Same as Meta-RCNN & Not available \\ \hline
\rowcolor{lightmintbg} NIFF \cite{niff} & Same as Meta-RCNN & Not available \\ \hline
DiGeo \cite{digeo} & Same as TFA & Not available \\ \hline 
UPPR \cite{uppr} & Same as Meta-RCNN & Not available \\ \hline \hline

\multicolumn{3}{l}{\textbf{\textcolor{brown}{Incremental Few Shot Object Detection Methods}}} \\ \hline \hline
ONCE \cite{once} & \textbf{COCO}: Split same as Meta-RCNN. \textbf{Same Dataset Evaluation}: Evaluated on 20 novel classes of COCO. Number of shots are 1, 5, 10. \textbf{Cross Dataset Evaluation}: Evaluated on VOC 2007 test set. Number of shots are 5, 10. & Not available \\ \hline
\rowcolor{lightmintbg} LEAST \cite{least} & Same as ONCE & Not available \\ \hline
iMFTA \cite{imfta} & Same as ONCE & PyTorch(Official): github.com/danganea/iMTFA \\ \hline
\rowcolor{lightmintbg} Meta-iFSOD \cite{meta_ifsod} & Same as ONCE & PyTorch(Official): github.com/Tongji-MIC-Lab/ML-iFSOD \\ \hline
Sylph \cite{sylph} & \textbf{COCO}: Split same as ONCE. Number of shots are 1, 5, 10, 20, 30. \textbf{LVIS}: Same as TFA & PyTorch(Official): github.com/facebookresearch/sylph-few-shot-detection \\ \hline
\rowcolor{lightmintbg} iFS-RCNN \cite{ifsrcnn} & \textbf{COCO}: Split same as ONCE. Number of shots are 1, 2, 3, 5, 10, 30. \textbf{LVIS}: Same as TFA. & PyTorch(Official): github.com/ducminhkhoi/iFS-RCNN \\ \hline
i-DETR \cite{idetr} & Same as ONCE & Official: github.com/dongnana777/Incremental-DETR (Code not available) \\ \hline
\rowcolor{lightmintbg} SC AWG \cite{sc_awg} & Same as ONCE & Not available \\ \hline
iTFA \cite{ifsod_sft} & Same as TFA & Official: github.com/TMIU/iTFA (Code not available) \\ \hline
\rowcolor{lightmintbg} Tang \textit{et al.} \cite{non-registrable_weights} & Same as ONCE & Not available \\ \hline
WS-iFSD \cite{ws_ifsd} & Same as ONCE & Not available \\ \hline \hline

\multicolumn{3}{l}{\textbf{\textcolor{brown}{Open-Set Few Shot Object Detection Methods}}} \\ \hline \hline
FOOD \cite{g_food, food} & The benchmarks include single and cross-dataset evaluations. For the single dataset benchmarks, \textbf{VOC10-5-5} is divided into 10 base known, 5 novel known, and 5 unknown classes, while \textbf{LVIS315-454-461} comprises 315 frequent base known, 454 rare novel known, and 461 common unknown classes. For the cross-dataset benchmark, \textbf{VOC-COCO} features 20 VOC base known and 20 non-VOC novel known classes as closed-set data, with the remaining 40 COCO classes used as unknown classes. The number of shots evaluated are 1, 2, 3, 5, 10, and 30. & Not available \\ \hline
\rowcolor{lightmintbg} Binyi \textit{et al.} \cite{hsic} & Same as FOOD for VOC10-5-5 and VOC-COCO & PyTorch: github.com/binyisu/food \\ \hline
FOXOD \cite{foxod} & \textbf{SIXray}: Three splits. Each split has 2 base classes, 2 novel classes, and 1 unknown class. Number of shots are 10, 30. \textbf{PIDray}: Total 12 classes out of which 6 are base classes, 3 are novel classes, and 3 unknown classes. Number of shots are 10, 30. \textbf{VOC}: 10 base classes, 5 novel classes, 5 unknown classes. Number of shots are 1, 2, 3, 5, 10, 30. & Not available \\ \hline
\rowcolor{lightmintbg} OFDet \cite{ofdet} & Same as Meta-RCNN & Not available \\ \hline \hline

\multicolumn{3}{l}{\textbf{\textcolor{brown}{Few Shot Domain Adaptive Object Detection}}} \\ \hline \hline
CD-ViTO \cite{cd-vito} & Source Dataset: COCO\cite{coco}. Target Datasets: ArTaxOr, Clipart1k, DIOR, DeepFish, NEU-DET, UODD. & Not available \\ \hline
\rowcolor{lightmintbg} AsyFOD \cite{asyfod} & The \textbf{Source} $\rightarrow$ \textbf{Target} datasets are: CityScapes $\rightarrow$ Foggy CityScapes; SIM10K $\rightarrow$ CityScapes; ViPeD $\rightarrow$ COCO; KITTI $\rightarrow$ CityScapes & PyTorch(Official): github.com/Hlings/AsyFOD \\ \hline
AcroFOD \cite{acrofod} & Same as AsyFOD & PyTorch(Official): github.com/Hlings/AcroFOD \\ \hline
\rowcolor{lightmintbg} Wang \textit{et al.} \cite{fs_adaptive_frcnn} & The \textbf{Source} $\rightarrow$ \textbf{Target} datasets are: SIM10k $\rightarrow$ Udacity; SIM10k $\rightarrow$ CityScapes; CityScapes $\rightarrow$ Udacity; Udacity $\rightarrow$ CityScapes; CityScapes $\rightarrow$ Foggy CityScapes. & Not available \\ \hline
PICA \cite{pica} & The \textbf{Source} $\rightarrow$ \textbf{Target} datasets are: CityScapes $\rightarrow$ Foggy CityScapes; SIM10k $\rightarrow$ CityScapes; Udacity $\rightarrow$ CityScapes. & Not available \\ \hline
\rowcolor{lightmintbg} Nakamura \textit{et al.} \cite{cutmix} & The \textbf{Source} $\rightarrow$ \textbf{Target} datasets are: BDD100K $\rightarrow$ FLIR; Caltech $\rightarrow$ KAIST; SIM10K $\rightarrow$ Cityscapes. & Not available \\ \hline
Yan \textit{et al.} \cite{adaptive_meta_rcnn} & The \textbf{Source} $\rightarrow$ \textbf{Target} datasets are: VOC $\rightarrow$ Cityscapes; VOC $\rightarrow$ Clipart; KITTI $\rightarrow$ Cityscapes. & Not available \\ \hline
\end{longtable}
}
}
\subsection{Evaluation Protocol} \label{subsec: evaluation}
\textbf{Metrics}: An effective evaluation of predicted bounding boxes is essential in object detection (OD) models. One key metric for this purpose is the Intersection over Union (IoU), which measures the overlap between predicted bounding boxes and the ground truth bounding boxes. Mathematically, it is computed as
\begin{equation}
    \text{IoU} = \frac{{\text{{area of overlap}}}}{{\text{{area of union}}}} = \frac{{b_{\text{true}} \cap b_{\text{pred}}}}{{b_{\text{true}} \cup b_{\text{pred}}}}
\end{equation}
Given a confidence threshold $t_{c}$, an IoU threshold $t_{IoU}$, and a set of ground-truth annotations G, true positives (\textbf{TP}) are detections that meet three criteria: (i) the confidence score is greater than $t_{c}$, (ii) the predicted class matches the class of ground truth, and (iii) the predicted bounding box has an IoU greater than $t_{IOU}$ with the ground truth. Detections that fail to meet either of the last two criteria are classified as false positives (\textbf{FP}). Conversely, if a valid detection does not meet the confidence score threshold, it is considered a false negative (\textbf{FN}). Detections with low confidence scores are flagged as true negatives (\textbf{TN}). 

Precision is the ratio of TP detections to the total number of positive detections (TP + FP), i.e., precision = TP / (TP + FP). Meanwhile, recall is the ratio of TP detections to the total number of ground truth positive instances (TP + FN), i.e., recall = TP / (TP + FN). Different pairs of Precision and Recall values can be obtained by varying the confidence score threshold, which is then used to plot a Precision-Recall curve, which visualizes how the two metrics are related. Similarly, varying $t_{IOU}$ provides different pairs of IoU and recall values, which can be used to draw an IoU-Recall curve that helps evaluate the effectiveness of detection proposals. The Precision-Recall curve can subsequently be used to compute the Average Precision (AP) by averaging the precision across all recall levels. This can be formulated as 
\begin{equation}
    AP = \frac{1}{N} \sum_{i=1}^{N} P_i \cdot \Delta R_i
\end{equation}
where $N$ represents the total number of recall or precision predictions, $\text{P}_{i}$ and $\Delta\text{R}_{i}$ are precision and change in Recall (R) at i-th prediction. When averaged over all classes, this yields the mean Average Precision (mAP). Similarly, the Average Recall (AR) is the Recall averaged over all IoU $\in$ [0.5, 1.0] and the mean Average Recall (mAR) is the corresponding averaged version.
\\
\\
\textbf{Computing AP}: The AP is computed differently in the VOC and COCO datasets. Here, we describe how it is computed for each dataset.
\begin{itemize}
    \item \textbf{VOC Dataset}: This dataset includes 20 object categories. The steps to compute AP in VOC are as follows:
    \begin{enumerate}
        \item For each category, calculate the precision-recall curve by varying the confidence threshold of the model’s predictions.
        \item Calculate each category’s average precision (AP) using an interpolated 11-point sampling of the precision-recall curve.
        \item Compute the final average precision (AP) by taking the mean of the APs across all 20 categories.
    \end{enumerate}
    \item \textbf{COCO Dataset}: This dataset includes 80 object categories and employs a more complex method for calculating AP. Instead of the 11-point interpolation, it uses a 101-point interpolation, computing precision for 101 recall thresholds from 0 to 1 in increments of 0.01. The AP is obtained by averaging over multiple IoU values instead of just one, except for a common AP metric called AP50, which is the AP for a single IoU threshold of 0.5. The AP calculation in COCO follows these steps:
    \begin{enumerate}
        \item For each category, generate the precision-recall curve by varying the confidence threshold of the model’s predictions.
        \item Compute each category’s average precision (AP) using 101 recall thresholds.
        \item Calculate AP at different IoU thresholds, typically ranging from 0.5 to 0.95 with a step size of 0.05. Higher IoU thresholds require more accurate predictions to be classified as true positives.
        \item For each IoU threshold, compute the mean of the APs across all 80 categories.
        \item Finally, compute the overall AP by averaging the AP values obtained at each IoU threshold.
    \end{enumerate}
\end{itemize}
The differences in AP calculation methods make it difficult to directly compare the performance of object detection models across the VOC and COCO datasets. The COCO AP has become the current standard due to its more detailed evaluation of model performance at various IoU thresholds.

\section{Result Discussion} \label{sec: Discussion}
\subsection{Discussion on results of Standard FSOD methods} \label{subsec:discussion-FSOD}
Standard FSOD methods have evaluated their performance on the PASCAL-VOC and COCO benchmarks. The evaluation metric for PASCAL-VOC is mAP, while for COCO, it is AP at various IOU thresholds (AP50, AP75, AP50-95). 
For the COCO dataset, as shown in Table \ref{tab:fsod_coco}, FM-FSOD \cite{fm_fsod} obtains the best results across all metrics (AP50, AP75, AP50-95), for 10 and 30 shots settings, gaining 13.2-58.4\% and 12-44.4\% respectively over previous best-performing LVC model \cite{lvc}. 

For the PASCAL-VOC dataset, as detailed in Table \ref{tab:fsod_pascal}, CRTED \cite{crted} demonstrates the best performance across all splits and shot settings. It achieves a gain of 3.3-8.5\%, 30.8-60.4\%, 13.8-19.8\% in split 1, split 2 and split 3 respectively, compared to SNIDA \cite{snida} with the MFD backbone.

\begin{table}[t]
\caption{Result Comparison between standard FSOD methods on COCO dataset with shots \textbf{K}=1,10,30 on the novel classes. The metrics used for comparison are \textbf{AP50}, \textbf{AP75} and \textbf{AP50-95}}
\label{tab:fsod_coco}
\vspace{-1em}
\centering
\begin{adjustbox}{max width=0.99\linewidth}
\begin{tabular}{lrl|ccccccccc}
\hline \hline
\multirow{3}{*}{\textbf{Method}} & \multirow{3}{*}{\textbf{Publication}} & \multirow{3}{*}{\textbf{Model}} & \multicolumn{9}{c}{\rule{-2pt}{10pt}\textbf{Novel classes performance}}   \\ \cline{4-12} 
                        &                              &                        & \multicolumn{3}{c|}{\rule{-2pt}{10pt}\textbf{K = 1}}   & \multicolumn{3}{c|}{\textbf{K = 10}} &  \multicolumn{3}{c}{\textbf{K = 30}}     \\   \cline{4-12}
                         &                              &                        & \multicolumn{1}{c}{\rule{-2pt}{10pt}\textbf{AP50}} & \multicolumn{1}{c}{\textbf{AP75}} & \multicolumn{1}{c|}{\textbf{AP50-95}}  & \multicolumn{1}{c}{\textbf{AP50}} & \multicolumn{1}{c}{\textbf{AP75}} & \multicolumn{1}{c|}{\textbf{AP50-95}}  & \multicolumn{1}{c}{\textbf{AP50}} & \multicolumn{1}{c}{\textbf{AP75}} & \multicolumn{1}{c}{\textbf{AP50-95}}    \\\hline \hline

\rowcolor{lightmintbg} \rule{-2pt}{10pt} YOLO-FR \cite{YOLO-FR}                  & ICCV 2019                   & YOLOv2   & \multicolumn{1}{c}{-}  & \multicolumn{1}{c}{-}  & \multicolumn{1}{c|}{ - } & \multicolumn{1}{c}{12.3}  & \multicolumn{1}{c}{4.6}  &\multicolumn{1}{c|}{ 5.6}    & \multicolumn{1}{c}{19.0}  & \multicolumn{1}{c}{7.6}    & 9.1 \\
 
\rule{-2pt}{10pt} MetaDet \cite{Meta-Det}                  & ICCV 2019                   & Faster R-CNN   & \multicolumn{1}{c}{-}  & \multicolumn{1}{c}{-}  & \multicolumn{1}{c|}{ - } & \multicolumn{1}{c}{14.6}  & \multicolumn{1}{c}{6.1}  &\multicolumn{1}{c|}{ 7.1}    & \multicolumn{1}{c}{21.7}  & \multicolumn{1}{c}{8.1}    & 11.3 \\

\rowcolor{lightmintbg} \rule{-2pt}{10pt} Meta R-CNN \cite{Meta-RCNN}                  & ICCV 2019                   & Mask R-CNN   & \multicolumn{1}{c}{-}  & \multicolumn{1}{c}{-}  & \multicolumn{1}{c|}{ - } & \multicolumn{1}{c}{19.1}  & \multicolumn{1}{c}{6.6}  &\multicolumn{1}{c|}{ 8.7}    & \multicolumn{1}{c}{25.3}  & \multicolumn{1}{c}{10.8}    & 12.4 \\

\rule{-2pt}{10pt} AttFDNet \cite{AttFDNet}                  & arXiv 2020              & SDD VGG-16   & \multicolumn{1}{c}{-}  & \multicolumn{1}{c}{-}  & \multicolumn{1}{c|}{-} & \multicolumn{1}{c}{19.5}  & \multicolumn{1}{c}{13.9}  &\multicolumn{1}{c|}{12.9}    & \multicolumn{1}{c}{24.6}  & \multicolumn{1}{c}{17.3}    & 16.3 \\ 

\rowcolor{lightmintbg} \rule{-2pt}{10pt} MPSR \cite{mpsr}                  &        ECCV 2020        & Faster R-CNN R-101   & \multicolumn{1}{c}{-}  & \multicolumn{1}{c}{-}  & \multicolumn{1}{c|}{-} & \multicolumn{1}{c}{17.9}  & \multicolumn{1}{c}{9.7}  &\multicolumn{1}{c|}{9.8}    & \multicolumn{1}{c}{25.4}  & \multicolumn{1}{c}{14.1}    & 14.2 \\ 

\rule{-2pt}{10pt} Meta-RetinaNet \cite{meta_retina_net}                  &        BMVC 2020        & RetinaNet R-18 & \multicolumn{1}{c}{-}  & \multicolumn{1}{c}{-}  & \multicolumn{1}{c|}{-} & \multicolumn{1}{c}{19.9}  & \multicolumn{1}{c}{7.7}  &\multicolumn{1}{c|}{9.7}    & \multicolumn{1}{c}{26.7}  & \multicolumn{1}{c}{11.2}    & 13.1 \\ 

\rowcolor{lightmintbg} \rule{-2pt}{10pt} SQMGH \cite{SQMGH}                  & CVPR 2021               & Faster R-CNN R-101   & \multicolumn{1}{c}{-}  & \multicolumn{1}{c}{-}  & \multicolumn{1}{c|}{-} & \multicolumn{1}{c}{29.5}  & \multicolumn{1}{c}{11.7}  &\multicolumn{1}{c|}{13.9}    & \multicolumn{1}{c}{-}  & \multicolumn{1}{c}{-}    & - \\  

\rule{-2pt}{10pt} BMM-CME \cite{BMM-CME}                  & CVPR 2021               & YOLOv2   & \multicolumn{1}{c}{-}  & \multicolumn{1}{c}{-}  & \multicolumn{1}{c|}{-} & \multicolumn{1}{c}{24.6}  & \multicolumn{1}{c}{16.4}  &\multicolumn{1}{c|}{15.1}    & \multicolumn{1}{c}{28.0}  & \multicolumn{1}{c}{17.8}    & 16.9 \\ 

\rowcolor{lightmintbg} \rule{-2pt}{10pt} Halluc. + CoRPN \cite{Halluc}                  &        CVPR 2021           & Faster R-CNN R-101    & \multicolumn{1}{c}{7.5}  & \multicolumn{1}{c}{4.9}  & \multicolumn{1}{c|}{4.4} & \multicolumn{1}{c}{-}  & \multicolumn{1}{c}{-}  &\multicolumn{1}{c|}{-}    & \multicolumn{1}{c}{-}  & \multicolumn{1}{c}{-}    & - \\ 

\rule{-2pt}{10pt} CGDP+FSCN \cite{cdgp}                  &        CVPR 2021           & Faster R-CNN R-50    & \multicolumn{1}{c}{-}  & \multicolumn{1}{c}{-}  & \multicolumn{1}{c|}{-} & \multicolumn{1}{c}{20.3}  & \multicolumn{1}{c}{-}  &\multicolumn{1}{c|}{11.3}    & \multicolumn{1}{c}{29.4}  & \multicolumn{1}{c}{-}    & 15.1 \\ 

\rowcolor{lightmintbg} \rule{-2pt}{10pt} SRR-FSD \cite{srr_fsd}                  &        CVPR 2021 & Faster R-CNN R-101 & \multicolumn{1}{c}{-}  & \multicolumn{1}{c}{-}  & \multicolumn{1}{c|}{-} & \multicolumn{1}{c}{23.0}  & \multicolumn{1}{c}{9.8}  &\multicolumn{1}{c|}{11.3}    & \multicolumn{1}{c}{29.2}  & \multicolumn{1}{c}{13.5}    & 14.7 \\ 

\rule{-2pt}{10pt} FSCE \cite{fsce}                  &        CVPR 2021 & Faster R-CNN R-101 & \multicolumn{1}{c}{-}  & \multicolumn{1}{c}{-}  & \multicolumn{1}{c|}{-} & \multicolumn{1}{c}{-}  & \multicolumn{1}{c}{10.5}  &\multicolumn{1}{c|}{11.9}    & \multicolumn{1}{c}{-}  & \multicolumn{1}{c}{16.2}    & 16.4 \\ 

\rowcolor{lightmintbg} \rule{-2pt}{10pt} SVD \cite{svd}                  &        NeurIPS 2021        & Faster R-CNN R-101   & \multicolumn{1}{c}{-}  & \multicolumn{1}{c}{-}  & \multicolumn{1}{c|}{-} & \multicolumn{1}{c}{-}  & \multicolumn{1}{c}{10.4}  &\multicolumn{1}{c|}{12.0}    & \multicolumn{1}{c}{-}  & \multicolumn{1}{c}{15.3}    & 16.0 \\ 

\rule{-2pt}{10pt} FADI \cite{fadi}                  &        NeurIPS 2021 & Faster R-CNN R-101 & \multicolumn{1}{c}{-}  & \multicolumn{1}{c}{-}  & \multicolumn{1}{c|}{-} & \multicolumn{1}{c}{-}  & \multicolumn{1}{c}{11.9}  &\multicolumn{1}{c|}{12.2}    & \multicolumn{1}{c}{-}  & \multicolumn{1}{c}{15.8}    & 16.1 \\ 

\rowcolor{lightmintbg} \rule{-2pt}{10pt} Lee \textit{et al.} (APSP - avg) \cite{apsp}                  &        WACV 2022 & Faster R-CNN R-50 & \multicolumn{1}{c}{-}  & \multicolumn{1}{c}{-}  & \multicolumn{1}{c|}{-} & \multicolumn{1}{c}{30.6}  & \multicolumn{1}{c}{9.1}  &\multicolumn{1}{c|}{13.4}    & \multicolumn{1}{c}{35.2}  & \multicolumn{1}{c}{14.7}    & 17.1 \\ 

\rule{-2pt}{10pt} Meta-Faster-RCNN \cite{meta_faster_rcnn}                  &        AAAI 2022        & Faster R-CNN R-101 & \multicolumn{1}{c}{10.7}  & \multicolumn{1}{c}{4.3}  & \multicolumn{1}{c|}{5.1} & \multicolumn{1}{c}{25.7}  & \multicolumn{1}{c}{10.8}  &\multicolumn{1}{c|}{12.7}    & \multicolumn{1}{c}{31.9}  & \multicolumn{1}{c}{14.7}    & 15.9 \\ 

\rowcolor{lightmintbg} \rule{-2pt}{10pt} TD-sampler \cite{td_sampler}                  &        ICCCBDA 2022        & Faster R-CNN R-101   & \multicolumn{1}{c}{-}  & \multicolumn{1}{c}{-}  & \multicolumn{1}{c|}{-} & \multicolumn{1}{c}{-}  & \multicolumn{1}{c}{14.8}  &\multicolumn{1}{c|}{13.6}    & \multicolumn{1}{c}{-}  & \multicolumn{1}{c}{19.2}    & 19.9 \\ 

\rule{-2pt}{10pt} Meta-DETR \cite{meta_detr}                  &        TPAMI 2022        & Def. DETR R-101 & \multicolumn{1}{c}{12.5}  & \multicolumn{1}{c}{7.7}  & \multicolumn{1}{c|}{7.7} & \multicolumn{1}{c}{30.5}  & \multicolumn{1}{c}{19.7}  &\multicolumn{1}{c|}{19.0}    & \multicolumn{1}{c}{35.0}  & \multicolumn{1}{c}{22.8}    & 22.2 \\ 

\rowcolor{lightmintbg} \rule{-2pt}{10pt} FCT \cite{fct}                  &        CVPR 2022        & Faster R-CNN PVTv2-B2-Li & \multicolumn{1}{c}{-}  & \multicolumn{1}{c}{-}  & \multicolumn{1}{c|}{5.6} & \multicolumn{1}{c}{-}  & \multicolumn{1}{c}{-}  &\multicolumn{1}{c|}{17.1}    & \multicolumn{1}{c}{-}  & \multicolumn{1}{c}{-}    & 21.4 \\ 

\rule{-2pt}{10pt} \multirow{1}{*}{LVC \cite{lvc}}                  &        \multirow{1}{*} {CVPR 2022}        & Faster R-CNN Swin-S & \multicolumn{1}{c}{-}  & \multicolumn{1}{c}{-}  & \multicolumn{1}{c|}{-} & \multicolumn{1}{c}{34.1}  & \multicolumn{1}{c}{19.0}  &\multicolumn{1}{c|}{19.0}    & \multicolumn{1}{c}{45.8}  & \multicolumn{1}{c}{27.5}    & 26.8 \\ 

\rowcolor{lightmintbg} \rule{-2pt}{10pt} FORD+BL \cite{ford_bl}                  &        IMAVIS 2022        & Faster R-CNN R-101 & \multicolumn{1}{c}{7.1}  & \multicolumn{1}{c}{3.5}  & \multicolumn{1}{c|}{3.6} & \multicolumn{1}{c}{22.5}  & \multicolumn{1}{c}{10.2}  &\multicolumn{1}{c|}{11.2}    & \multicolumn{1}{c}{28.9}  & \multicolumn{1}{c}{13.9}    & 14.8 \\ 

\rule{-2pt}{10pt} MEMFRCN  \cite{memfrcn}                  &        IEICE Trans. 2022& Faster R-CNN R-101 & \multicolumn{1}{c}{-}  & \multicolumn{1}{c}{-}  & \multicolumn{1}{c|}{5.2} & \multicolumn{1}{c}{-}  & \multicolumn{1}{c}{-}  &\multicolumn{1}{c|}{14.0}    & \multicolumn{1}{c}{-}  & \multicolumn{1}{c}{-}    & 17.5 \\ 

\rowcolor{lightmintbg} \rule{-2pt}{10pt} DMNet \cite{dmnet}                  &        TCyb. 2023        & DMNet R-101 & \multicolumn{1}{c}{-}  & \multicolumn{1}{c}{-}  & \multicolumn{1}{c|}{-} & \multicolumn{1}{c}{17.4}  & \multicolumn{1}{c}{10.0}  &\multicolumn{1}{c|}{10.4}    & \multicolumn{1}{c}{29.7}  & \multicolumn{1}{c}{17.7}    & 17.1 \\ 

 \rule{-2pt}{10pt} DeFRCN \cite{defrcn}                  &        ICCV 2021        & Faster R-CNN R-101 & \multicolumn{1}{c}{-}  & \multicolumn{1}{c}{-}  & \multicolumn{1}{c|}{9.3} & \multicolumn{1}{c}{-}  & \multicolumn{1}{c}{-}  &\multicolumn{1}{c|}{18.5}    & \multicolumn{1}{c}{-}  & \multicolumn{1}{c}{-}    & 22.6 \\ 

\rowcolor{lightmintbg} \rule{-2pt}{10pt} Norm-VAE   \cite{norm_vae}          &        CVPR 2023        & Faster R-CNN R-101 & \multicolumn{1}{c}{-}  & \multicolumn{1}{c}{8.8}  & \multicolumn{1}{c|}{9.5} & \multicolumn{1}{c}{-}  & \multicolumn{1}{c}{17.8}  &\multicolumn{1}{c|}{18.7}    & \multicolumn{1}{c}{-}  & \multicolumn{1}{c}{22.4}    & 22.5 \\ 

 \rule{-2pt}{10pt} Du \textit{et al.} \cite{alpha-adaptive} &        ICCV 2023        & Faster R-CNN R-101 & \multicolumn{1}{c}{-}  & \multicolumn{1}{c}{-}  & \multicolumn{1}{c|}{-} & \multicolumn{1}{c}{-}  & \multicolumn{1}{c}{20.8}  &\multicolumn{1}{c|}{-}    & \multicolumn{1}{c}{-}  & \multicolumn{1}{c}{23.6}  & - \\ 

\rowcolor{lightmintbg} \rule{-2pt}{10pt} FSSP \cite{fssp} &       IEEE Access 2021  & YOLOv3-SPP & \multicolumn{1}{c}{-}  & \multicolumn{1}{c}{-}  & \multicolumn{1}{c|}{-} & \multicolumn{1}{c}{20.4}  & \multicolumn{1}{c}{9.6}  &\multicolumn{1}{c|}{-}    & \multicolumn{1}{c}{25.0}  & \multicolumn{1}{c}{13.9}    &  - \\

 \rule{-2pt}{10pt} Li \textit{et al.} \cite{disentangle_and_remerge} &      AAAI 2023        & Faster R-CNN & \multicolumn{1}{c}{-}  & \multicolumn{1}{c}{-}  & \multicolumn{1}{c|}{6.1} & \multicolumn{1}{c}{-}  & \multicolumn{1}{c}{-}  &\multicolumn{1}{c|}{16.4}    & \multicolumn{1}{c}{-}  & \multicolumn{1}{c}{-}    &  20.0\\ 

\rowcolor{lightmintbg} \rule{-2pt}{10pt} MRSN  \cite{mrsn} &      ECCV 2022     & Faster R-CNN R-101 & \multicolumn{1}{c}{-}  & \multicolumn{1}{c}{-}  & \multicolumn{1}{c|}{-} & \multicolumn{1}{c}{-}  & \multicolumn{1}{c}{14.8}  &\multicolumn{1}{c|}{-}    & \multicolumn{1}{c}{-}  & \multicolumn{1}{c}{17.9}    & - \\ 

 \rule{-2pt}{10pt} AHT \cite{aht} & PRCV 2023  & Pyramid Vision Transformer & \multicolumn{1}{c}{-}  & \multicolumn{1}{c}{-}  & \multicolumn{1}{c|}{-} & \multicolumn{1}{c}{24.7}  & \multicolumn{1}{c}{13.5}  &\multicolumn{1}{c|}{-}    & \multicolumn{1}{c}{29.3}  & \multicolumn{1}{c}{17.9}    & - \\

\rowcolor{lightmintbg} \rule{-2pt}{10pt} FPD \cite{fpd} & arXiv 2024 & Faster R-CNN R-101 & \multicolumn{1}{c}{-}  & \multicolumn{1}{c}{-}  & \multicolumn{1}{c|}{-} & \multicolumn{1}{c}{-}  & \multicolumn{1}{c}{-}  &\multicolumn{1}{c|}{15.9}    & \multicolumn{1}{c}{-}  & \multicolumn{1}{c}{-}    & 19.3\\

\rule{-2pt}{10pt} SparseCT \cite{sparse_ct} & arXiv 2024 & SSD & \multicolumn{1}{c}{-}  & \multicolumn{1}{c}{-}  & \multicolumn{1}{c|}{-} & \multicolumn{1}{c}{14.3}  & \multicolumn{1}{c}{7.7}  &\multicolumn{1}{c|}{7.9}    & \multicolumn{1}{c}{20.2}  & \multicolumn{1}{c}{11.3}    & 11.2\\

\rowcolor{lightmintbg} \rule{-2pt}{10pt} CKDT \cite{ckdt} & IMAVIS 2024 & Faster R-CNN R-101 & \multicolumn{1}{c}{-}  & \multicolumn{1}{c}{-}  & \multicolumn{1}{c|}{-} & \multicolumn{1}{c}{33.0}  & \multicolumn{1}{c}{17.9}  &\multicolumn{1}{c|}{18.5}    & \multicolumn{1}{c}{39.2}  & \multicolumn{1}{c}{22.0}    & 22.3 \\

\rule{-2pt}{10pt} Zhu \textit{et al.} \cite{data_augmentation_and_distribution_calibration} & MVA 2024 & Faster R-CNN R-101& \multicolumn{1}{c}{11.2}  & \multicolumn{1}{c}{7.2}  & \multicolumn{1}{c|}{-} & \multicolumn{1}{c}{-}  & \multicolumn{1}{c}{-}  &\multicolumn{1}{c|}{-}    & \multicolumn{1}{c}{-}  & \multicolumn{1}{c}{-}    &  - \\

\rowcolor{lightmintbg} \rule{-2pt}{10pt} Yan \textit{et al.} \cite{unp} & TCSVT 2024 & Faster R-CNN R-101 & \multicolumn{1}{c}{-}  & \multicolumn{1}{c}{-}  & \multicolumn{1}{c|}{-} & \multicolumn{1}{c}{23.1}  & \multicolumn{1}{c}{11.5}  &\multicolumn{1}{c|}{-}    & \multicolumn{1}{c}{28.5}  & \multicolumn{1}{c}{14.8}    & -\\

\rule{-2pt}{10pt} FSNA \cite{fsna} & TCSVT 2024 & Faster R-CNN R-101 & \multicolumn{1}{c}{-}  & \multicolumn{1}{c}{-}  & \multicolumn{1}{c|}{-} & \multicolumn{1}{c}{25.4}  & \multicolumn{1}{c}{10.3}  &\multicolumn{1}{c|}{11.9}    & \multicolumn{1}{c}{31.1}  & \multicolumn{1}{c}{15.1}    & 16.1\\

\rowcolor{lightmintbg} \rule{-2pt}{10pt} CRTED \cite{crted} & MDPI Elec. 2024 & Faster R-CNN R-101 & \multicolumn{1}{c}{-}  & \multicolumn{1}{c}{-}  & \multicolumn{1}{c|}{6.3} & \multicolumn{1}{c}{-}  & \multicolumn{1}{c}{-}  &\multicolumn{1}{c|}{12.8}    & \multicolumn{1}{c}{-}  & \multicolumn{1}{c}{-}    & 19.3\\

\rule{-2pt}{10pt} SNIDA-MFD \cite{snida} & CVPR 2024 & Faster R-CNN & \multicolumn{1}{c}{-}  & \multicolumn{1}{c}{-}  & \multicolumn{1}{c|}{12.0} & \multicolumn{1}{c}{-}  & \multicolumn{1}{c}{-}  &\multicolumn{1}{c|}{20.7}    & \multicolumn{1}{c}{-}  & \multicolumn{1}{c}{-}    & 23.8\\

\rowcolor{lightmintbg} \rule{-2pt}{10pt} FM-FSOD \cite{fm_fsod} & CVPR 2024 & DINOv2 & \multicolumn{1}{c}{7.8}  & \multicolumn{1}{c}{6.2}  & \multicolumn{1}{c|}{5.7} & \multicolumn{1}{c}{38.6}  & \multicolumn{1}{c}{30.1}  &\multicolumn{1}{c|}{27.7}    & \multicolumn{1}{c}{51.3}  & \multicolumn{1}{c}{39.7}    & 37.0\\

\hline \hline
\end{tabular}
\end{adjustbox}
\end{table}
\begin{table}[t]
\caption{Result comparison between standard FSOD methods on the 3 splits of PASCAL VOC dataset with shots \textbf{K}=1, 2, 3, 5, 10 on the novel classes. The metric used for comparison is \textbf{AP}. }
\label{tab:fsod_pascal}
\vspace{-1em}
\centering
\begin{adjustbox}{max width=\linewidth}
\begin{tabular}{ll|ccccc|ccccc|ccccc}
\hline \hline
\multirow{2}{*}{\textbf{Method}} & \multirow{2}{*}{\textbf{Model}} & \multicolumn{5}{c|}{\rule{-2pt}{10pt}\textbf{Novel set1}}                                                                            & \multicolumn{5}{c|}{\textbf{Novel set 2}}                                                                       & \multicolumn{5}{c}{\textbf{Novel set 3}}                                                                        \\ \cline{3-17} 
                        &                                     & \multicolumn{1}{c}{\rule{-2pt}{10pt}\textbf{K = 1}} & \multicolumn{1}{c}{\textbf{2}} & \multicolumn{1}{c}{\textbf{3}} & \multicolumn{1}{c}{\textbf{5}} & \textbf{10} & \multicolumn{1}{c}{\textbf{1}} & \multicolumn{1}{c}{\textbf{2}} & \multicolumn{1}{c}{\textbf{3}} & \multicolumn{1}{c}{\textbf{5}} & \textbf{10} & \multicolumn{1}{c}{\textbf{1}} & \multicolumn{1}{c}{\textbf{2}} & \multicolumn{1}{c}{\textbf{3}} & \multicolumn{1}{c}{\textbf{5}} & \textbf{10}    \\ \hline \hline

\rowcolor{lightmintbg} \rule{-2pt}{10pt} YOLO-FR \cite{YOLO-FR}                 & YOLOv2   & \multicolumn{1}{c}{14.8}  & \multicolumn{1}{c}{15.5}  & \multicolumn{1}{c}{26.7}  & \multicolumn{1}{c}{33.9}  &  47.2  & \multicolumn{1}{c}{15.7}  & \multicolumn{1}{c}{15.3}  & \multicolumn{1}{c}{22.7}  & \multicolumn{1}{c}{30.1}  &  39.2  & \multicolumn{1}{c}{19.2}  & \multicolumn{1}{c}{21.7}  & \multicolumn{1}{c}{25.7}  & \multicolumn{1}{c}{40.6}  & 41.3       \\ 

\rule{-2pt}{10pt} \multirow{1}{*}{\rule{-2pt}{10pt}Meta-Det \cite{Meta-Det} }  
& Faster R-CNN R-101 & \multicolumn{1}{c}{18.9}  & \multicolumn{1}{c}{20.6}  & \multicolumn{1}{c}{30.2}  & \multicolumn{1}{c}{36.8}  &  49.6  & \multicolumn{1}{c}{21.8}  & \multicolumn{1}{c}{23.1}  & \multicolumn{1}{c}{27.8}  & \multicolumn{1}{c}{31.7}  & 43.0   & \multicolumn{1}{c}{20.6}  & \multicolumn{1}{c}{23.9}  & \multicolumn{1}{c}{29.1}  & \multicolumn{1}{c}{43.9}  & 44.1       \\

\rowcolor{lightmintbg} \rule{-2pt}{10pt} Meta R-CNN \cite{Meta-RCNN}             & Mask R-CNN R-101   & \multicolumn{1}{c}{19.9}  & \multicolumn{1}{c}{25.5}  & \multicolumn{1}{c}{35.0}  & \multicolumn{1}{c}{45.7}  &  51.5  & \multicolumn{1}{c}{10.4}  & \multicolumn{1}{c}{19.4}  & \multicolumn{1}{c}{29.6}  & \multicolumn{1}{c}{34.8}  &  45.4  & \multicolumn{1}{c}{14.3}  & \multicolumn{1}{c}{18.2}  & \multicolumn{1}{c}{27.5}  & \multicolumn{1}{c}{41.2}  & 48.1       \\ 

\rule{-2pt}{10pt} RepMet \cite{RepMet}    & Faster R-CNN R-101   & \multicolumn{1}{c}{26.1}  & \multicolumn{1}{c}{32.9}  & \multicolumn{1}{c}{34.4}  & \multicolumn{1}{c}{38.6}  &   41.3 & \multicolumn{1}{c}{17.2}  & \multicolumn{1}{c}{22.1}  & \multicolumn{1}{c}{23.4}  & \multicolumn{1}{c}{28.3}  & 35.8   & \multicolumn{1}{c}{27.5}  & \multicolumn{1}{c}{31.1}  & \multicolumn{1}{c}{31.5}  & \multicolumn{1}{c}{34.4}  & 37.2         \\

\rowcolor{lightmintbg} \rule{-2pt}{10pt} RNI \cite{RNI}                  & Faster R-CNN R-101   & \multicolumn{1}{c}{37.8}  & \multicolumn{1}{c}{40.3}  & \multicolumn{1}{c}{41.7}  & \multicolumn{1}{c}{47.3}  &   49.4 & \multicolumn{1}{c}{41.6}  & \multicolumn{1}{c}{43.0}  & \multicolumn{1}{c}{43.4}  & \multicolumn{1}{c}{47.4}  & 49.1   & \multicolumn{1}{c}{33.3}  & \multicolumn{1}{c}{38.0}  & \multicolumn{1}{c}{39.8}  & \multicolumn{1}{c}{41.5}  & 44.8         \\

\rule{-2pt}{10pt} AttFDNet \cite{AttFDNet}                   & SSD VGG-16  & \multicolumn{1}{c}{29.6}  & \multicolumn{1}{c}{34.9}  & \multicolumn{1}{c}{35.1}  & \multicolumn{1}{c}{-}  &   - & \multicolumn{1}{c}{16.0}  & \multicolumn{1}{c}{20.7}  & \multicolumn{1}{c}{22.1}  & \multicolumn{1}{c}{-}  & -  & \multicolumn{1}{c}{22.6}  & \multicolumn{1}{c}{29.1}  & \multicolumn{1}{c}{32.0}  & \multicolumn{1}{c}{-}  & -         \\

\rowcolor{lightmintbg} \rule{-2pt}{10pt} Zhang \textit{et al.} \cite{CoRPN}                  & Faster R-CNN R-101  & \multicolumn{1}{c}{44.4}  & \multicolumn{1}{c}{38.5}  & \multicolumn{1}{c}{46.4}  & \multicolumn{1}{c}{54.1}  &   55.7 & \multicolumn{1}{c}{25.7}  & \multicolumn{1}{c}{29.5}  & \multicolumn{1}{c}{37.3}  & \multicolumn{1}{c}{36.2}  & 41.3  & \multicolumn{1}{c}{35.8}  & \multicolumn{1}{c}{41.8}  & \multicolumn{1}{c}{44.6}  & \multicolumn{1}{c}{51.6}  & 49.6        \\

\rule{-2pt}{10pt} Kim \textit{et al.} \cite{FSOD-KT}                  &  Faster R-CNN R-101  & \multicolumn{1}{c}{27.8}  & \multicolumn{1}{c}{41.4}  & \multicolumn{1}{c}{46.2}  & \multicolumn{1}{c}{55.2}  &   56.8 & \multicolumn{1}{c}{19.8}  & \multicolumn{1}{c}{27.9}  & \multicolumn{1}{c}{38.7}  & \multicolumn{1}{c}{38.9}  & 41.5  & \multicolumn{1}{c}{29.5}  & \multicolumn{1}{c}{30.6}  & \multicolumn{1}{c}{38.6}  & \multicolumn{1}{c}{43.8}  & 45.7        \\

\rowcolor{lightmintbg} \rule{-2pt}{10pt} MPSR \cite{mpsr}                   & Faster R-CNN R-101  & \multicolumn{1}{c}{41.7}  & \multicolumn{1}{c}{-}  & \multicolumn{1}{c}{51.4}  & \multicolumn{1}{c}{55.2}  &   61.8 & \multicolumn{1}{c}{24.4}  & \multicolumn{1}{c}{-}  & \multicolumn{1}{c}{39.2}  & \multicolumn{1}{c}{39.9}  & 47.8  & \multicolumn{1}{c}{35.6}  & \multicolumn{1}{c}{-}  & \multicolumn{1}{c}{42.3}  & \multicolumn{1}{c}{48.0}  & 49.7        \\

\rule{-2pt}{10pt} Meta-RetinaNet \cite{meta_retina_net}                  & RetinaNet R-18  & \multicolumn{1}{c}{38.3}  & \multicolumn{1}{c}{51.8}  & \multicolumn{1}{c}{59.3}  & \multicolumn{1}{c}{65.3}  &   71.5 & \multicolumn{1}{c}{28.4}  & \multicolumn{1}{c}{36.8}  & \multicolumn{1}{c}{42.4}  & \multicolumn{1}{c}{45.5}  & 50.9  & \multicolumn{1}{c}{35.9}  & \multicolumn{1}{c}{48.1}  & \multicolumn{1}{c}{53.2}  & \multicolumn{1}{c}{58.0}  & 63.6        \\

\rowcolor{lightmintbg} \multirow{1}{*}{SQMGH \cite{SQMGH}}  & Faster R-CNN R-101 & \multicolumn{1}{c}{48.6}  & \multicolumn{1}{c}{51.1}  & \multicolumn{1}{c}{52.0}  & \multicolumn{1}{c}{53.7}  &  54.3  & \multicolumn{1}{c}{41.6}  & \multicolumn{1}{c}{45.4}  & \multicolumn{1}{c}{45.8}  & \multicolumn{1}{c}{46.3}  & 48.0   & \multicolumn{1}{c}{46.1}  & \multicolumn{1}{c}{51.7}  & \multicolumn{1}{c}{52.6}  & \multicolumn{1}{c}{54.1}  & 55.0       \\

\multirow{1}{*}{BMM-CME (MSPR) \cite{BMM-CME}}  & Faster R-CNN R-101 & \multicolumn{1}{c}{41.5}  & \multicolumn{1}{c}{47.5}  & \multicolumn{1}{c}{50.4}  & \multicolumn{1}{c}{58.2}  &  60.9  & \multicolumn{1}{c}{27.2}  & \multicolumn{1}{c}{30.2}  & \multicolumn{1}{c}{41.4}  & \multicolumn{1}{c}{42.5}  & 46.8   & \multicolumn{1}{c}{34.3}  & \multicolumn{1}{c}{39.6}  & \multicolumn{1}{c}{45.1}  & \multicolumn{1}{c}{48.3}  & 51.5       \\

\rowcolor{lightmintbg} \rule{-2pt}{10pt} Halluc. + CoRPN \cite{Halluc}                  & Faster R-CNN R-101  & \multicolumn{1}{c}{47.0}  & \multicolumn{1}{c}{44.9}  & \multicolumn{1}{c}{46.5}  & \multicolumn{1}{c}{54.7}  &   54.7 & \multicolumn{1}{c}{26.3}  & \multicolumn{1}{c}{31.8}  & \multicolumn{1}{c}{37.4}  & \multicolumn{1}{c}{37.4}  & 41.2  & \multicolumn{1}{c}{40.4}  & \multicolumn{1}{c}{42.1}  & \multicolumn{1}{c}{43.3}  & \multicolumn{1}{c}{51.4}  & 49.6        \\

\rule{-2pt}{10pt} CGDP+FSCN \cite{cdgp}          & Faster R-CNN R-50  & \multicolumn{1}{c}{40.7}  & \multicolumn{1}{c}{45.1}  & \multicolumn{1}{c}{46.5}  & \multicolumn{1}{c}{57.4}  &   62.4 & \multicolumn{1}{c}{27.3}  & \multicolumn{1}{c}{31.4}  & \multicolumn{1}{c}{40.8}  & \multicolumn{1}{c}{42.7}  & 46.3  & \multicolumn{1}{c}{31.2}  & \multicolumn{1}{c}{36.4}  & \multicolumn{1}{c}{43.7}  & \multicolumn{1}{c}{50.1}  & 55.6        \\
 
\rowcolor{lightmintbg} \rule{-2pt}{10pt}  SRR-FSD \cite{srr_fsd} & Faster R-CNN R-101 & \multicolumn{1}{c}{47.8}  & \multicolumn{1}{c}{50.5}  & \multicolumn{1}{c}{51.3}  & \multicolumn{1}{c}{55.2}  &   56.8 & \multicolumn{1}{c}{32.5}  & \multicolumn{1}{c}{35.3}  & \multicolumn{1}{c}{39.1}  & \multicolumn{1}{c}{40.8}  & 43.8  & \multicolumn{1}{c}{40.1}  & \multicolumn{1}{c}{41.5}  & \multicolumn{1}{c}{44.3}  & \multicolumn{1}{c}{46.9}  & 46.4        \\

\rule{-2pt}{10pt} FSCE \cite{fsce}                  & Faster R-CNN R-101 & \multicolumn{1}{c}{44.2}  & \multicolumn{1}{c}{43.8}  & \multicolumn{1}{c}{51.4}  & \multicolumn{1}{c}{61.9}  &   63.4 & \multicolumn{1}{c}{27.3}  & \multicolumn{1}{c}{29.5}  & \multicolumn{1}{c}{43.5}  & \multicolumn{1}{c}{44.2}  & 50.2  & \multicolumn{1}{c}{37.2}  & \multicolumn{1}{c}{41.9}  & \multicolumn{1}{c}{47.5}  & \multicolumn{1}{c}{54.6}  & 58.5        \\

\rowcolor{lightmintbg} \rule{-2pt}{10pt}  FADI \cite{fadi}    & Faster R-CNN R-101 & \multicolumn{1}{c}{50.3}  & \multicolumn{1}{c}{54.8}  & \multicolumn{1}{c}{54.2}  & \multicolumn{1}{c}{59.3}  &   63.2 & \multicolumn{1}{c}{30.6}  & \multicolumn{1}{c}{35.0}  & \multicolumn{1}{c}{40.3}  & \multicolumn{1}{c}{42.8}  & 48.0  & \multicolumn{1}{c}{45.7}  & \multicolumn{1}{c}{49.7}  & \multicolumn{1}{c}{49.1}  & \multicolumn{1}{c}{55.0}  & 59.6        \\
   
\rule{-2pt}{10pt} SVD \cite{svd}            & Faster R-CNN R-101  & \multicolumn{1}{c}{41.5}  & \multicolumn{1}{c}{47.4}  & \multicolumn{1}{c}{51.5}  & \multicolumn{1}{c}{57.7}  &   61.2 & \multicolumn{1}{c}{29.4}  & \multicolumn{1}{c}{29.6}  & \multicolumn{1}{c}{39.8}  & \multicolumn{1}{c}{41.2}  & 51.5  & \multicolumn{1}{c}{36.0}  & \multicolumn{1}{c}{39.4}  & \multicolumn{1}{c}{45.4}  & \multicolumn{1}{c}{50.4}  & 51.3        \\

\rowcolor{lightmintbg} \rule{-2pt}{10pt}  Lee \textit{et al.} \cite{apsp}       & Faster R-CNN R-101 & \multicolumn{1}{c}{31.1}  & \multicolumn{1}{c}{36.1}  & \multicolumn{1}{c}{39.2}  & \multicolumn{1}{c}{50.7}  &   59.4 & \multicolumn{1}{c}{22.9}  & \multicolumn{1}{c}{29.4}  & \multicolumn{1}{c}{32.1}  & \multicolumn{1}{c}{35.4}  & 42.7  & \multicolumn{1}{c}{24.3}  & \multicolumn{1}{c}{28.6}  & \multicolumn{1}{c}{35.0}  & \multicolumn{1}{c}{50.0}  & 53.6        \\

\rule{-2pt}{10pt}  Meta-Faster-RCNN \cite{meta_faster_rcnn}       & Faster R-CNN R-101 & \multicolumn{1}{c}{43.0}  & \multicolumn{1}{c}{54.6}  & \multicolumn{1}{c}{60.6}  & \multicolumn{1}{c}{66.1}  &   65.4 & \multicolumn{1}{c}{27.7}  & \multicolumn{1}{c}{35.5}  & \multicolumn{1}{c}{46.1}  & \multicolumn{1}{c}{47.8}  & 51.4  & \multicolumn{1}{c}{40.6}  & \multicolumn{1}{c}{46.4}  & \multicolumn{1}{c}{53.4}  & \multicolumn{1}{c}{59.9}  & 58.6        \\

\rowcolor{lightmintbg} \rule{-2pt}{10pt}  FCT \cite{fct}      & Faster R-CNN PVTv2-B2-Li & \multicolumn{1}{c}{49.9}  & \multicolumn{1}{c}{57.1}  & \multicolumn{1}{c}{57.9}  & \multicolumn{1}{c}{63.2}  &   67.1 & \multicolumn{1}{c}{27.6}  & \multicolumn{1}{c}{34.5}  & \multicolumn{1}{c}{43.7}  & \multicolumn{1}{c}{49.2}  & 51.2  & \multicolumn{1}{c}{39.5}  & \multicolumn{1}{c}{54.7}  & \multicolumn{1}{c}{52.3}  & \multicolumn{1}{c}{57.0}  & 58.7        \\

\rule{-2pt}{10pt}  LVC \cite{lvc}       & Faster R-CNN R-101 & \multicolumn{1}{c}{54.5}  & \multicolumn{1}{c}{53.2}  & \multicolumn{1}{c}{58.8}  & \multicolumn{1}{c}{63.2}  &   65.7 & \multicolumn{1}{c}{32.8}  & \multicolumn{1}{c}{29.2}  & \multicolumn{1}{c}{50.7}  & \multicolumn{1}{c}{44.8}  & 50.6  & \multicolumn{1}{c}{48.4}  & \multicolumn{1}{c}{52.7}  & \multicolumn{1}{c}{55.0}  & \multicolumn{1}{c}{59.6}  & 59.6        \\

\rowcolor{lightmintbg} \rule{-2pt}{10pt}  FORD+BL \cite{ford_bl}            & Faster R-CNN R-101 & \multicolumn{1}{c}{46.3}  & \multicolumn{1}{c}{54.2}  & \multicolumn{1}{c}{49.9}  & \multicolumn{1}{c}{56.3}  &   61.8 & \multicolumn{1}{c}{19.0}  & \multicolumn{1}{c}{30.8}  & \multicolumn{1}{c}{38.4}  & \multicolumn{1}{c}{39.3}  & 47.3  & \multicolumn{1}{c}{36.4}  & \multicolumn{1}{c}{46.5}  & \multicolumn{1}{c}{45.4}  & \multicolumn{1}{c}{53.2}  & 55.8        \\

\rule{-2pt}{10pt} TD-sampler \cite{td_sampler}                  &   Faster R-CNN R-101  & \multicolumn{1}{c}{37.1}  & \multicolumn{1}{c}{47.8}  & \multicolumn{1}{c}{50.5}  & \multicolumn{1}{c}{56.2}  &   63.1 & \multicolumn{1}{c}{26.3}  & \multicolumn{1}{c}{35.0}  & \multicolumn{1}{c}{42.9}  & \multicolumn{1}{c}{46.8}  & 52.0  & \multicolumn{1}{c}{25.5}  & \multicolumn{1}{c}{36.8}  & \multicolumn{1}{c}{43.5}  & \multicolumn{1}{c}{50.9}  & 58.2        \\

\rowcolor{lightmintbg} \rule{-2pt}{10pt}  MEMFRCN \cite{memfrcn}        & Faster R-CNN R-101 & \multicolumn{1}{c}{36.4}  & \multicolumn{1}{c}{37.4}  & \multicolumn{1}{c}{40.6}  & \multicolumn{1}{c}{45.5}  &   46.6 & \multicolumn{1}{c}{18.0}  & \multicolumn{1}{c}{26.8}  & \multicolumn{1}{c}{32.1}  & \multicolumn{1}{c}{36.3}  & 32.4  & \multicolumn{1}{c}{30.3}  & \multicolumn{1}{c}{32.3}  & \multicolumn{1}{c}{37.3}  & \multicolumn{1}{c}{37.8}  & 38.5        \\

\rule{-2pt}{10pt}  Meta-DETR \cite{meta_detr}        & Def. DETR R-101 & \multicolumn{1}{c}{40.6}  & \multicolumn{1}{c}{51.4}  & \multicolumn{1}{c}{58.0}  & \multicolumn{1}{c}{59.2}  &   63.6 & \multicolumn{1}{c}{37.0}  & \multicolumn{1}{c}{36.6}  & \multicolumn{1}{c}{43.7}  & \multicolumn{1}{c}{49.1}  & 54.6  & \multicolumn{1}{c}{41.6}  & \multicolumn{1}{c}{45.9}  & \multicolumn{1}{c}{52.7}  & \multicolumn{1}{c}{58.9}  & 60.6        \\

\rowcolor{lightmintbg} \rule{-2pt}{10pt}  DMNet \cite{dmnet}              & DMNet R-101 & \multicolumn{1}{c}{39.0}  & \multicolumn{1}{c}{48.9}  & \multicolumn{1}{c}{50.7}  & \multicolumn{1}{c}{58.6}  &   62.5 & \multicolumn{1}{c}{31.2}  & \multicolumn{1}{c}{32.4}  & \multicolumn{1}{c}{40.3}  & \multicolumn{1}{c}{47.6}  & 52.0  & \multicolumn{1}{c}{41.7}  & \multicolumn{1}{c}{41.8}  & \multicolumn{1}{c}{42.7}  & \multicolumn{1}{c}{50.3}  & 52.1        \\

\rule{-2pt}{10pt}  DeFRCN \cite{reinforce}   & Faster R-CNN & \multicolumn{1}{c}{58.4}  & \multicolumn{1}{c}{62.4}  & \multicolumn{1}{c}{63.2}  & \multicolumn{1}{c}{67.6}  &  67.7  & \multicolumn{1}{c}{34.0}  & \multicolumn{1}{c}{43.1}  & \multicolumn{1}{c}{51.0}  & \multicolumn{1}{c}{53.6}  & 54.0  & \multicolumn{1}{c}{55.1}  & \multicolumn{1}{c}{56.6}  & \multicolumn{1}{c}{57.3}  & \multicolumn{1}{c}{62.6}  &  63.7       \\

\rowcolor{lightmintbg} \rule{-2pt}{10pt}  Norm-VAE \cite{norm_vae}          &  Faster R-CNN R-101& \multicolumn{1}{c}{62.1}  & \multicolumn{1}{c}{64.9}  & \multicolumn{1}{c}{67.8}  & \multicolumn{1}{c}{69.2}  & 67.5   & \multicolumn{1}{c}{39.9}  & \multicolumn{1}{c}{46.8}  & \multicolumn{1}{c}{54.4}  & \multicolumn{1}{c}{54.2}  &  53.6 & \multicolumn{1}{c}{58.2}  & \multicolumn{1}{c}{60.3}  & \multicolumn{1}{c}{61.0}  & \multicolumn{1}{c}{64.0}  &   65.5      \\

\rule{-2pt}{10pt}  Du \textit{et al.} \cite{alpha-adaptive}              & Faster R-CNN R-101 & \multicolumn{1}{c}{35.9}  & \multicolumn{1}{c}{40.3}  & \multicolumn{1}{c}{49.8}  & \multicolumn{1}{c}{56.8}  &   65.1 & \multicolumn{1}{c}{25.6}  & \multicolumn{1}{c}{30.3}  & \multicolumn{1}{c}{41.7}  & \multicolumn{1}{c}{41.8}  & 50.3  & \multicolumn{1}{c}{33.9}  & \multicolumn{1}{c}{35.6}  & \multicolumn{1}{c}{43.5}  & \multicolumn{1}{c}{47.1}  & 55.9   \\

\rowcolor{lightmintbg} \rule{-2pt}{10pt}  FSSP \cite{fssp}       &  YOLOv3-SPP & \multicolumn{1}{c}{41.6}  & \multicolumn{1}{c}{-}  & \multicolumn{1}{c}{49.1}  & \multicolumn{1}{c}{54.2}  &  56.5  & \multicolumn{1}{c}{30.5}  & \multicolumn{1}{c}{-}  & \multicolumn{1}{c}{39.5}  & \multicolumn{1}{c}{41.4}  & 45.1  & \multicolumn{1}{c}{36.7}  & \multicolumn{1}{c}{-}  & \multicolumn{1}{c}{45.3}  & \multicolumn{1}{c}{49.4}  & 51.3        \\

\rule{-2pt}{10pt}  Li \textit{et al.} \cite{disentangle_and_remerge}        & Faster R-CNN & \multicolumn{1}{c}{41.0}  & \multicolumn{1}{c}{51.7}  & \multicolumn{1}{c}{55.7}  & \multicolumn{1}{c}{61.8}  &  65.4  & \multicolumn{1}{c}{30.7}  & \multicolumn{1}{c}{39.0}  & \multicolumn{1}{c}{42.5}  & \multicolumn{1}{c}{46.6}  & 51.7  & \multicolumn{1}{c}{37.9}  & \multicolumn{1}{c}{47.1}  & \multicolumn{1}{c}{51.7}  & \multicolumn{1}{c}{56.8}  & 59.5        \\

\rowcolor{lightmintbg} \rule{-2pt}{10pt}  Fan \textit{et al.}   \cite{model_calibration}       & Faster R-CNN R50 & \multicolumn{1}{c}{40.1}  & \multicolumn{1}{c}{44.2}  & \multicolumn{1}{c}{51.2}  & \multicolumn{1}{c}{62.0}  &  63.0  & \multicolumn{1}{c}{33.3}  & \multicolumn{1}{c}{33.1}  & \multicolumn{1}{c}{42.3}  & \multicolumn{1}{c}{46.3}  & 52.3  & \multicolumn{1}{c}{36.1}  & \multicolumn{1}{c}{43.1}  & \multicolumn{1}{c}{43.5}  & \multicolumn{1}{c}{52.0}  &  56.0       \\

\rule{-2pt}{10pt}  MRSN \cite{mrsn}      &   Faster R-CNN R-101   & \multicolumn{1}{c}{47.6}  & \multicolumn{1}{c}{48.6}  & \multicolumn{1}{c}{57.8}  & \multicolumn{1}{c}{61.9}  & 62.6   & \multicolumn{1}{c}{31.2}  & \multicolumn{1}{c}{38.3}  & \multicolumn{1}{c}{46.7}  & \multicolumn{1}{c}{47.1}  & 50.6  & \multicolumn{1}{c}{35.5}  & \multicolumn{1}{c}{30.9}  & \multicolumn{1}{c}{45.6}  & \multicolumn{1}{c}{54.4}  &  57.4       \\

\rowcolor{lightmintbg} \rule{-2pt}{10pt} AHT \cite{aht} & Pyramid Vision Transformer& \multicolumn{1}{c}{53.9}  & \multicolumn{1}{c}{64.9}  & \multicolumn{1}{c}{62.0}  & \multicolumn{1}{c}{68.2}  &  69.0  & \multicolumn{1}{c}{27.1}  & \multicolumn{1}{c}{33.9}  & \multicolumn{1}{c}{41.5}  & \multicolumn{1}{c}{45.7}  & 51.4  & \multicolumn{1}{c}{40.9}  & \multicolumn{1}{c}{50.4}  & \multicolumn{1}{c}{53.3}  & \multicolumn{1}{c}{63.2}  &  64.0       \\

\rule{-2pt}{10pt} ARDNet \cite{ardnet} & Faster R-CNN R-101 & \multicolumn{1}{c}{33.4}  & \multicolumn{1}{c}{50.5}  & \multicolumn{1}{c}{57.2}  & \multicolumn{1}{c}{59.0}  & 65.4   & \multicolumn{1}{c}{28.0}  & \multicolumn{1}{c}{47.6}  & \multicolumn{1}{c}{53.9}  & \multicolumn{1}{c}{56.4}  & 59.4  & \multicolumn{1}{c}{27.9}  & \multicolumn{1}{c}{39.6}  & \multicolumn{1}{c}{41.7}  & \multicolumn{1}{c}{43.7}  &  49.8       \\

\rowcolor{lightmintbg} \rule{-2pt}{10pt} Zhu \textit{et al.} \cite{data_augmentation_and_distribution_calibration} & Faster R-CNN R-101& \multicolumn{1}{c}{48.1}  & \multicolumn{1}{c}{54.2}  & \multicolumn{1}{c}{56.1}  & \multicolumn{1}{c}{62.1}  &  65.9  & \multicolumn{1}{c}{38.8}  & \multicolumn{1}{c}{40.9}  & \multicolumn{1}{c}{46.8}  & \multicolumn{1}{c}{49.2}  &  54.5 & \multicolumn{1}{c}{41.1}  & \multicolumn{1}{c}{47.2}  & \multicolumn{1}{c}{52.1}  & \multicolumn{1}{c}{57.8}  &  60.2   \\

\rule{-2pt}{10pt} Yan \textit{et al.} \cite{unp} & Faster R-CNN R-101& \multicolumn{1}{c}{43.7}  & \multicolumn{1}{c}{58.3}  & \multicolumn{1}{c}{59.8}  & \multicolumn{1}{c}{63.7}  &  64.2  & \multicolumn{1}{c}{28.1}  & \multicolumn{1}{c}{42.8}  & \multicolumn{1}{c}{47.7}  & \multicolumn{1}{c}{49.5}  &  50.3 & \multicolumn{1}{c}{38.4}  & \multicolumn{1}{c}{49.3}  & \multicolumn{1}{c}{53.8}  & \multicolumn{1}{c}{57.7}  &  58.7 \\

\rowcolor{lightmintbg} \rule{-2pt}{10pt} CKDT \cite{ckdt} & Faster R-CNN R-101 & \multicolumn{1}{c}{48.6}  & \multicolumn{1}{c}{60.6}  & \multicolumn{1}{c}{64.3}  & \multicolumn{1}{c}{69.0}  &  70.8  & \multicolumn{1}{c}{33.0}  & \multicolumn{1}{c}{42.1}  & \multicolumn{1}{c}{46.6}  & \multicolumn{1}{c}{52.4}  &  53.3 & \multicolumn{1}{c}{40.2}  & \multicolumn{1}{c}{52.9}  & \multicolumn{1}{c}{55.2}  & \multicolumn{1}{c}{61.6}  &  63.7 \\

\rule{-2pt}{10pt} CRTED \cite{crted} & Faster R-CNN R-101 & \multicolumn{1}{c}{69.8}  & \multicolumn{1}{c}{70.2}  & \multicolumn{1}{c}{72.0}  & \multicolumn{1}{c}{75.4}  & 76.5  & \multicolumn{1}{c}{67.7}  & \multicolumn{1}{c}{68.0}  & \multicolumn{1}{c}{71.3}  & \multicolumn{1}{c}{71.8}  & 73.7 & \multicolumn{1}{c}{68.6}  & \multicolumn{1}{c}{70.4}  & \multicolumn{1}{c}{72.7}  & \multicolumn{1}{c}{73.9}  & 75.1 \\

\rowcolor{lightmintbg} \rule{-2pt}{10pt} SNIDA-MFD \cite{snida} & Faster R-CNN R-101& \multicolumn{1}{c}{64.9}  & \multicolumn{1}{c}{67.9}  & \multicolumn{1}{c}{69.7}  & \multicolumn{1}{c}{71.4}  &  70.5  & \multicolumn{1}{c}{42.2}  & \multicolumn{1}{c}{47.8}  & \multicolumn{1}{c}{54.5}  & \multicolumn{1}{c}{56.6}  & 54.9 & \multicolumn{1}{c}{58.1}  & \multicolumn{1}{c}{61.3}  & \multicolumn{1}{c}{60.7}  & \multicolumn{1}{c}{63.6}  &  66.0 \\

\rule{-2pt}{10pt} FSNA \cite{fsna} & Faster R-CNN R-101& \multicolumn{1}{c}{43.8}  & \multicolumn{1}{c}{47.7}  & \multicolumn{1}{c}{50.8}  & \multicolumn{1}{c}{57.4}  &  60.3 & \multicolumn{1}{c}{23.9}  & \multicolumn{1}{c}{32.3}  & \multicolumn{1}{c}{37.9}  & \multicolumn{1}{c}{40.2}  &  41.8  & \multicolumn{1}{c}{34.0}  & \multicolumn{1}{c}{40.7}  & \multicolumn{1}{c}{45.5}  & \multicolumn{1}{c}{52.3}  & 54.0 \\

\rowcolor{lightmintbg} \rule{-2pt}{10pt} FPD \cite{fpd} & Faster R-CNN R-101& \multicolumn{1}{c}{41.5}  & \multicolumn{1}{c}{52.8}  & \multicolumn{1}{c}{58.4}  & \multicolumn{1}{c}{64.9}  &  67.1 & \multicolumn{1}{c}{28.2}  & \multicolumn{1}{c}{38.7}  & \multicolumn{1}{c}{43.8}  & \multicolumn{1}{c}{50.3}  & 53.6 & \multicolumn{1}{c}{34.9}  & \multicolumn{1}{c}{48.6}  & \multicolumn{1}{c}{54.0}  & \multicolumn{1}{c}{58.4}  & 61.5  \\

\rule{-2pt}{10pt} SparseCT \cite{sparse_ct} & SSD & \multicolumn{1}{c}{37.9}  & \multicolumn{1}{c}{40.8}  & \multicolumn{1}{c}{41.2}  & \multicolumn{1}{c}{44.9}  &  45.5 & \multicolumn{1}{c}{32.1}  & \multicolumn{1}{c}{33.1}  & \multicolumn{1}{c}{32.7}  & \multicolumn{1}{c}{36.3}  &  36.4 & \multicolumn{1}{c}{33.7}  & \multicolumn{1}{c}{31.8}  & \multicolumn{1}{c}{35.4}  & \multicolumn{1}{c}{39.1}  & 43.2 \\

\rowcolor{lightmintbg} \rule{-2pt}{10pt} FM-FSOD \cite{fm_fsod} & DINOv2 & \multicolumn{1}{c}{40.1}  & \multicolumn{1}{c}{53.5}  & \multicolumn{1}{c}{57.0}  & \multicolumn{1}{c}{68.6}  &  72 & \multicolumn{1}{c}{33.1}  & \multicolumn{1}{c}{36.3}  & \multicolumn{1}{c}{48.8}  & \multicolumn{1}{c}{54.8}  &  64.7 & \multicolumn{1}{c}{39.2}  & \multicolumn{1}{c}{50.2}  & \multicolumn{1}{c}{55.7}  & \multicolumn{1}{c}{63.4}  & 68.1 \\
\hline \hline
\end{tabular}
\vspace{-10em}
\end{adjustbox}
\end{table}

\subsection{Discussion on results of G-FSOD methods} \label{subsec:discussion-GFSOD}
G-FSOD works, similar to standard FSOD, have been evaluated on the COCO and Pascal VOC datasets. G-FSOD methods access their works on both novel and base class images. 
Table~\ref{Tab:GFSOD-COCO} presents the result comparison for the COCO dataset. It can be observed that the UPPR \cite{uppr} performs best on novel classes with a gain of 0-3.5\% over the second-best CFA \cite{cfa} on 10 and 30 shot, as well as on the base classes with a gain of 6.6-6.9\% over the second best performing DiGeo \cite{digeo}.
For the PASCAL VOC benchmark, as shown in Table \ref{Tab:GFSOD-Pascal}, UPPR \cite{uppr} and NIFF \cite{niff} outperform other methods in most settings when evaluating novel sets. On the other hand, TFA \cite{tfa} performs best on the base class evaluation sets, achieving a gain of 3.8-23.9\% across all shot numbers (1, 2, 3, 5, 10). On all classes (i.e., base classes + novel classes) evaluation, UPPR \cite{uppr} outperforms all other methods on all shot numbers, gaining 0.1-2.7\% over the second-best performing NIFF \cite{niff}, except shot 3 of split 2, where NIFF \cite{niff} gets the highest AP score of 74.5.

\begin{table}[t]
\caption{Result comparison between G-FSOD methods on the novel and base classes of the COCO dataset with shots \textbf{K}=1, 10, 30. The metric of comparison is AP.} \label{Tab:GFSOD-COCO}
\vspace{-1em}
\centering
\begin{adjustbox}{max width=0.8\linewidth}
\begin{tabular}{lrl|ccc|ccc}
\hline \hline
\multirow{2}{*}{\textbf{Method}} & \multirow{2}{*}{\textbf{Publication}} & \multirow{2}{*}{\textbf{Model}} & \multicolumn{3}{c|}{\rule{-2pt}{10pt}\textbf{Novel classes }}  & \multicolumn{3}{c}{\rule{-2pt}{10pt}\textbf{Base classes }}  \\ \cline{4-9}                         
                        &                              &                        & \multicolumn{1}{c}{\rule{-2pt}{10pt}\textbf{K = 1}}   & \multicolumn{1}{c}{\textbf{10}} &  \multicolumn{1}{c|}{\textbf{30}}     & \multicolumn{1}{c}{\rule{-2pt}{10pt}\textbf{1}}   & \multicolumn{1}{c}{\textbf{10}} &  \multicolumn{1}{c}{\textbf{30}}  
                             \\\hline \hline
\rowcolor{lightmintbg} \rule{-2pt}{10pt} FRCN+ft-full~\cite{tfa}                & ICML 2020             & Faster R-CNN R-101   & \multicolumn{1}{c}{1.8}  & \multicolumn{1}{c}{9.2}  & \multicolumn{1}{c|}{12.5}& \multicolumn{1}{c}{24.8}  & \multicolumn{1}{c}{21.0}  & \multicolumn{1}{c}{20.6}   \\
\rule{-2pt}{10pt} FRCN+ft-full avg~\cite{tfa}                & ICML 2020             & Faster R-CNN R-101   & \multicolumn{1}{c}{1.7}  & \multicolumn{1}{c}{5.5}  & \multicolumn{1}{c|}{7.4}& \multicolumn{1}{c}{21.0}  & \multicolumn{1}{c}{16.1}  & \multicolumn{1}{c}{15.6}   \\  
\rowcolor{lightmintbg} \rule{-2pt}{10pt} TFA~\cite{tfa}                & ICML 2020             & Faster R-CNN R-101   & \multicolumn{1}{c}{3.4}  & \multicolumn{1}{c}{10.0}  & \multicolumn{1}{c|}{13.7}& \multicolumn{1}{c}{34.1}  & \multicolumn{1}{c}{35.0}  & \multicolumn{1}{c}{35.8}   \\
\rule{-2pt}{10pt} TFA avg~\cite{tfa}                & ICML 2020             & Faster R-CNN R-101   & \multicolumn{1}{c}{1.9}  & \multicolumn{1}{c}{9.1}  & \multicolumn{1}{c|}{12.1}& \multicolumn{1}{c}{31.9}  & \multicolumn{1}{c}{32.4}  & \multicolumn{1}{c}{34.2}   \\
\rowcolor{lightmintbg} \rule{-2pt}{10pt} Retentive R-CNN~\cite{retentive_rcnn}                & CVPR 2021             & Faster R-CNN R-101   & \multicolumn{1}{c}{-}  & \multicolumn{1}{c}{10.5}  & \multicolumn{1}{c|}{13.8}& \multicolumn{1}{c}{-}  & \multicolumn{1}{c}{39.2}  & \multicolumn{1}{c}{39.3}   \\
\rule{-2pt}{10pt} CenterNet-ft-full~\cite{pnpdet}                 & WACV 2021                   & DLA34   & \multicolumn{1}{c}{-}  & \multicolumn{1}{c}{1.4}  & \multicolumn{1}{c|}{ - }& \multicolumn{1}{c}{-}  & \multicolumn{1}{c}{20.7}  & \multicolumn{1}{c}{ - }   \\
 
\rowcolor{lightmintbg} \rule{-2pt}{10pt} PNPDet~\cite{pnpdet}                    & WACV 2021                   & DLA34   & \multicolumn{1}{c}{-}  & \multicolumn{1}{c}{5.5}  & \multicolumn{1}{c|}{ - }& \multicolumn{1}{c}{-}  & \multicolumn{1}{c}{25.8}  & \multicolumn{1}{c}{ - }   \\
\rule{-2pt}{10pt} CFA  \cite{cfa}                  & CVPRW 2022                   &  Faster R-CNN R-101  & \multicolumn{1}{c}{-}  & \multicolumn{1}{c}{19.1}  & \multicolumn{1}{c|}{23.0}& \multicolumn{1}{c}{-}  & \multicolumn{1}{c}{35.5}  & \multicolumn{1}{c}{35.0}   \\
\rowcolor{lightmintbg} \rule{-2pt}{10pt} NIFF   \cite{niff}                 & CVPR 2023                   &  Faster R-CNN R-101  & \multicolumn{1}{c}{-}  & \multicolumn{1}{c}{18.8}  & \multicolumn{1}{c|}{ 20.9 }& \multicolumn{1}{c}{-}  & \multicolumn{1}{c}{39.0}  & \multicolumn{1}{c}{39.0}   \\

\rule{-2pt}{10pt} DiGeo  \cite{digeo}                & CVPR 2023                   &  Faster R-CNN R-101  & \multicolumn{1}{c}{-}  & \multicolumn{1}{c}{10.3}  & \multicolumn{1}{c|}{ 14.2 }& \multicolumn{1}{c}{-}  & \multicolumn{1}{c}{39.2}  & \multicolumn{1}{c}{39.4}   \\

\rowcolor{lightmintbg} \rule{-2pt}{10pt} UPPR  \cite{uppr}                & CVPR 2024    &  Cascade R-CNN R-101  & \multicolumn{1}{c}{-}  & \multicolumn{1}{c}{19.1}  & \multicolumn{1}{c|}{23.8}& \multicolumn{1}{c}{-}  & \multicolumn{1}{c}{41.9}  & \multicolumn{1}{c}{42.0}   \\
\hline \hline
\end{tabular}
\end{adjustbox}
\end{table}

\begin{table}[t!]
\caption{Result comparison between G-FSOD methods on the 3 splits of the Pascal VOC dataset on novel, base and all classes with shots \textit{K}=1, 2, 3, 5, 10. The metric of comparison is \textbf{AP}.} \label{Tab:GFSOD-Pascal}
\vspace{-1em}
\centering
\begin{adjustbox}{max width=\linewidth}
\begin{tabular}{lll|ccccc|ccccc|ccccc}
\hline \hline
\multirow{2}{*}{\textbf{Method}} & \multirow{2}{*}{\textbf{Publication}} & \multirow{2}{*}{\textbf{Model}} & \multicolumn{5}{c|}{\rule{-2pt}{10pt}\textbf{Novel set1}} & \multicolumn{5}{c|}{\textbf{Novel set 2}} & \multicolumn{5}{c}{\textbf{Novel set 3}}                                                    \\ \cline{4-18} 
                        &                              &                        & \multicolumn{1}{c}{\rule{-2pt}{10pt}\textbf{K = 1}} & \multicolumn{1}{c}{\textbf{2}} & \multicolumn{1}{c}{\textbf{3}} & \multicolumn{1}{c}{\textbf{5}} & \multicolumn{1}{c|}{\textbf{10}} &\multicolumn{1}{c}{\textbf{1}} & \multicolumn{1}{c}{\textbf{2}} & \multicolumn{1}{c}{\textbf{3}} & \multicolumn{1}{c}{\textbf{5}}& \multicolumn{1}{c|}{\textbf{10}} &\multicolumn{1}{c}{\textbf{1}} & \multicolumn{1}{c}{\textbf{2}} & \multicolumn{1}{c}{\textbf{3}} & \multicolumn{1}{c}{\textbf{5}} & \multicolumn{1}{c}{\textbf{10}}  \\ \hline \hline

\rowcolor{lightmintbg} \rule{-2pt}{10pt} YOLO-FR \cite{YOLO-FR}    &  ICCV 2019          & YOLO v2   & \multicolumn{1}{c}{14.8}  & \multicolumn{1}{c}{-}  & \multicolumn{1}{c}{26.7}  & \multicolumn{1}{c}{-}  &  47.2  &\multicolumn{1}{c}{15.7}  & \multicolumn{1}{c}{-}  & \multicolumn{1}{c}{22.7}  & \multicolumn{1}{c}{-}  &  40.5 & \multicolumn{1}{c}{21.3}  & \multicolumn{1}{c}{-}  & \multicolumn{1}{c}{28.4}  & \multicolumn{1}{c}{-}  & 45.9     \\ 

\rule{-2pt}{10pt} FRCN+ft-full~\cite{tfa}                & ICML 2020             & Faster R-CNN R-101   & \multicolumn{1}{c}{15.2}  & \multicolumn{1}{c}{20.3}  & \multicolumn{1}{c}{29.0}  & \multicolumn{1}{c}{40.1}  &  45.5  &\multicolumn{1}{c}{13.4}  & \multicolumn{1}{c}{20.6}  & \multicolumn{1}{c}{28.6}  & \multicolumn{1}{c}{32.4}  &  38.8  & \multicolumn{1}{c}{19.6}  & \multicolumn{1}{c}{20.8}  & \multicolumn{1}{c}{28.7}  & \multicolumn{1}{c}{42.2}  & 42.1  \\ 

\rowcolor{lightmintbg} \rule{-2pt}{10pt} FRCN+ft-full avg~\cite{tfa}                & ICML 2020             & Faster R-CNN R-101   & \multicolumn{1}{c}{9.9}  & \multicolumn{1}{c}{15.6}  & \multicolumn{1}{c}{21.6}  & \multicolumn{1}{c}{28.0}  &  35.6  &\multicolumn{1}{c}{9.4}  & \multicolumn{1}{c}{13.8}  & \multicolumn{1}{c}{17.4}  & \multicolumn{1}{c}{21.9}  &  29.8  & \multicolumn{1}{c}{8.1}  & \multicolumn{1}{c}{13.9}  & \multicolumn{1}{c}{19.0}  & \multicolumn{1}{c}{23.9}  & 31.0 \\ 

\rule{-2pt}{10pt} TFA~\cite{tfa}                & ICML 2020             & Faster R-CNN R-101   & \multicolumn{1}{c}{39.8}  & \multicolumn{1}{c}{36.1}  & \multicolumn{1}{c}{44.7}  & \multicolumn{1}{c}{55.7}  &  56.0 &\multicolumn{1}{c}{23.5}  & \multicolumn{1}{c}{26.9}  & \multicolumn{1}{c}{34.1}  & \multicolumn{1}{c}{35.1}  &  39.1 & \multicolumn{1}{c}{30.8}  & \multicolumn{1}{c}{34.8}  & \multicolumn{1}{c}{42.8}  & \multicolumn{1}{c}{49.5}  & 49.8  \\

\rowcolor{lightmintbg} \rule{-2pt}{10pt} TFA- avg~\cite{tfa}                & ICML 2020             & Faster R-CNN R-101   & \multicolumn{1}{c}{25.3}  & \multicolumn{1}{c}{36.4}  & \multicolumn{1}{c}{42.1}  & \multicolumn{1}{c}{47.9}  &  52.9  &\multicolumn{1}{c}{18.3}  & \multicolumn{1}{c}{27.5}  & \multicolumn{1}{c}{30.9}  & \multicolumn{1}{c}{34.1}  &  39.5 & \multicolumn{1}{c}{17.9}  & \multicolumn{1}{c}{27.2}  & \multicolumn{1}{c}{34.3}  & \multicolumn{1}{c}{40.8}  & 45.6    \\

\rule{-2pt}{10pt} CenterNet-ft-full~\cite{pnpdet}                 & WACV 2021             & DLA34   & \multicolumn{1}{c}{8.5}  & \multicolumn{1}{c}{-}  & \multicolumn{1}{c}{14.4}  & \multicolumn{1}{c}{-}  &  32.5  &\multicolumn{1}{c}{9.0}  & \multicolumn{1}{c}{-}  & \multicolumn{1}{c}{11.6}  & \multicolumn{1}{c}{-}  &  32.9  & \multicolumn{1}{c}{9.0}  & \multicolumn{1}{c}{-}  & \multicolumn{1}{c}{14.0}  & \multicolumn{1}{c}{-}  & 26.4    \\ 

\rowcolor{lightmintbg} \rule{-2pt}{10pt} PNPDet~\cite{pnpdet}                & WACV 2021             & DLA34   & \multicolumn{1}{c}{18.2}  & \multicolumn{1}{c}{-}  & \multicolumn{1}{c}{27.3}  & \multicolumn{1}{c}{-}  &  41.0  &\multicolumn{1}{c}{16.6}  & \multicolumn{1}{c}{-}  & \multicolumn{1}{c}{26.5}  & \multicolumn{1}{c}{-}  &  36.4 & \multicolumn{1}{c}{18.9}  & \multicolumn{1}{c}{-}  & \multicolumn{1}{c}{27.2}  & \multicolumn{1}{c}{-}  & 36.2  \\ 

\rule{-2pt}{10pt} Retentive R-CNN~\cite{retentive_rcnn}                & CVPR 2021             & Faster R-CNN R-101   & \multicolumn{1}{c}{42.4}  & \multicolumn{1}{c}{45.8}  & \multicolumn{1}{c}{45.9}  & \multicolumn{1}{c}{53.7}  &  56.1 &\multicolumn{1}{c}{21.7}  & \multicolumn{1}{c}{27.8}  & \multicolumn{1}{c}{35.2}  & \multicolumn{1}{c}{37.0}  &  40.3  & \multicolumn{1}{c}{30.2}  & \multicolumn{1}{c}{37.6}  & \multicolumn{1}{c}{43.0}  & \multicolumn{1}{c}{49.7}  & 50.1  \\

\rowcolor{lightmintbg} \rule{-2pt}{10pt} CFA~\cite{cfa} & CVPRW 2022             & Faster R-CNN R-101   & \multicolumn{1}{c}{59.0}  & \multicolumn{1}{c}{63.5}  & \multicolumn{1}{c}{66.4}  & \multicolumn{1}{c}{68.4}  & 68.3 &\multicolumn{1}{c}{37.0}  & \multicolumn{1}{c}{45.8}  & \multicolumn{1}{c}{50.0}  & \multicolumn{1}{c}{54.2}  &  52.5 & \multicolumn{1}{c}{54.8}  & \multicolumn{1}{c}{58.5}  & \multicolumn{1}{c}{56.5}  & \multicolumn{1}{c}{61.3}  & 63.5  \\

\rule{-2pt}{10pt} NIFF~\cite{niff} & CVPR 2023             & Faster R-CNN R-101   & \multicolumn{1}{c}{63.5}  & \multicolumn{1}{c}{67.2}  & \multicolumn{1}{c}{68.3}  & \multicolumn{1}{c}{71.1}  & 69.3 &\multicolumn{1}{c}{37.8}  & \multicolumn{1}{c}{41.9}  & \multicolumn{1}{c}{53.4}  & \multicolumn{1}{c}{56.0}  &  53.5 & \multicolumn{1}{c}{55.3}  & \multicolumn{1}{c}{60.5}  & \multicolumn{1}{c}{61.1}  & \multicolumn{1}{c}{63.7}  &  63.9 \\

\rowcolor{lightmintbg} \rule{-2pt}{10pt} DiGeo~\cite{digeo} & CVPR 2023             & Faster R-CNN R-101   & \multicolumn{1}{c}{37.9}  & \multicolumn{1}{c}{39.4}  & \multicolumn{1}{c}{48.5}  & \multicolumn{1}{c}{58.6}  & 61.5 &\multicolumn{1}{c}{26.6}  & \multicolumn{1}{c}{28.9}  & \multicolumn{1}{c}{41.9}  & \multicolumn{1}{c}{42.1}  &  49.1 & \multicolumn{1}{c}{30.4}  & \multicolumn{1}{c}{40.1}  & \multicolumn{1}{c}{46.9}  & \multicolumn{1}{c}{52.7}  &  54.7 \\

\rule{-2pt}{10pt} UPPR~\cite{uppr}  & CVPR 2024             & Cascade R-CNN R-101   & \multicolumn{1}{c}{61.0}  & \multicolumn{1}{c}{64.5}  & \multicolumn{1}{c}{67.8}  & \multicolumn{1}{c}{69.7}  & 69.0 &\multicolumn{1}{c}{38.5}  & \multicolumn{1}{c}{46.9}  & \multicolumn{1}{c}{51.4}  & \multicolumn{1}{c}{55.9}  & 53.6  & \multicolumn{1}{c}{55.3}  & \multicolumn{1}{c}{59.4}  & \multicolumn{1}{c}{57.5}  & \multicolumn{1}{c}{62.8}  & 64.1  \\
\end{tabular}
\end{adjustbox}
\begin{adjustbox}{max width=\linewidth}
\begin{tabular}{lll|ccccc|ccccc|ccccc}
\hline \hline
\multirow{2}{*}{\textbf{Method}} & \multirow{2}{*}{\textbf{Publication}} & \multirow{2}{*}{\textbf{Model}} & \multicolumn{5}{c|}{\rule{-2pt}{10pt}\textbf{Base set1}} & \multicolumn{5}{c|}{\textbf{Base set 2}} & \multicolumn{5}{c}{\textbf{Base set 3}}    

\\ \cline{4-18} 
                        &                              &                        & \multicolumn{1}{c}{\rule{-2pt}{10pt}\textbf{K = 1}} & \multicolumn{1}{c}{\textbf{2}} & \multicolumn{1}{c}{\textbf{3}} & \multicolumn{1}{c}{\textbf{5}} & \multicolumn{1}{c|}{\textbf{10}} &\multicolumn{1}{c}{\textbf{1}} & \multicolumn{1}{c}{\textbf{2}} & \multicolumn{1}{c}{\textbf{3}} & \multicolumn{1}{c}{\textbf{5}}& \multicolumn{1}{c|}{\textbf{10}} &\multicolumn{1}{c}{\textbf{1}} & \multicolumn{1}{c}{\textbf{2}} & \multicolumn{1}{c}{\textbf{3}} & \multicolumn{1}{c}{\textbf{5}} & \multicolumn{1}{c}{\textbf{10}} \\ \hline \hline

\rowcolor{lightmintbg} \rule{-2pt}{10pt} YOLO-FR \cite{YOLO-FR}    &  ICCV 2019          & YOLO v2   &\multicolumn{1}{c}{66.4}  & \multicolumn{1}{c}{-}  & \multicolumn{1}{c}{64.8}  & \multicolumn{1}{c}{-}  &  63.6 & \multicolumn{1}{c}{68.2}  & \multicolumn{1}{c}{-}  & \multicolumn{1}{c}{66.00}  & \multicolumn{1}{c}{-}  &  64.7 &\multicolumn{1}{c}{65.9}  & \multicolumn{1}{c}{-}  & \multicolumn{1}{c}{65.0}  & \multicolumn{1}{c}{-}  &  63.1   \\ 

\rule{-2pt}{10pt} FRCN+ft-full~\cite{tfa}                & ICML 2020             & Faster R-CNN R-101  \quad \quad &\multicolumn{1}{c}{68.9}  & \multicolumn{1}{c}{69.4}  & \multicolumn{1}{c}{66.1}  & \multicolumn{1}{c}{66.7}  &  66.0 & \multicolumn{1}{c}{62.4}  & \multicolumn{1}{c}{64.8}  & \multicolumn{1}{c}{62.0}  & \multicolumn{1}{c}{63.7}  &  61.0  &\multicolumn{1}{c}{71.4}  & \multicolumn{1}{c}{71.8}  & \multicolumn{1}{c}{68.7}  & \multicolumn{1}{c}{68.3}  &  67.0   \\ 

\rowcolor{lightmintbg} \rule{-2pt}{10pt} FRCN+ft-full avg~\cite{tfa}                & ICML 2020             & Faster R-CNN R-101   &\multicolumn{1}{c}{62.6}  & \multicolumn{1}{c}{60.7}  & \multicolumn{1}{c}{61.3}  & \multicolumn{1}{c}{60.6}  &  59.8 & \multicolumn{1}{c}{63.2}  & \multicolumn{1}{c}{61.6}  & \multicolumn{1}{c}{61.0}  & \multicolumn{1}{c}{60.4}  &  59.8 &\multicolumn{1}{c}{63.7}  & \multicolumn{1}{c}{62.4}  & \multicolumn{1}{c}{62.1}  & \multicolumn{1}{c}{61.0}  &  60.5   \\ 

\rule{-2pt}{10pt} TFA~\cite{tfa}                & ICML 2020             & Faster R-CNN R-101    &\multicolumn{1}{c}{79.6}  & \multicolumn{1}{c}{78.9}  & \multicolumn{1}{c}{79.1}  & \multicolumn{1}{c}{79.3}  &  78.4 & \multicolumn{1}{c}{79.5}  & \multicolumn{1}{c}{77.7}  & \multicolumn{1}{c}{78.8}  & \multicolumn{1}{c}{78.9}  &  78.5  &\multicolumn{1}{c}{80.3}  & \multicolumn{1}{c}{79.9}  & \multicolumn{1}{c}{80.4}  & \multicolumn{1}{c}{80.2}  &  79.9   \\

\rowcolor{lightmintbg} \rule{-2pt}{10pt} TFA- avg~\cite{tfa}                & ICML 2020             & Faster R-CNN R-101   &\multicolumn{1}{c}{77.6}  & \multicolumn{1}{c}{77.3}  & \multicolumn{1}{c}{77.3}  & \multicolumn{1}{c}{77.4}  &  77.5 & \multicolumn{1}{c}{73.8}  & \multicolumn{1}{c}{74.9}  & \multicolumn{1}{c}{75.6}  & \multicolumn{1}{c}{76.2}  &  76.9   &\multicolumn{1}{c}{78.7}  & \multicolumn{1}{c}{78.4}  & \multicolumn{1}{c}{78.6}  & \multicolumn{1}{c}{78.5}  &  78.6   \\

\rule{-2pt}{10pt} CenterNet-ft-full~\cite{pnpdet}                 & WACV 2021             & DLA34  &\multicolumn{1}{c}{68.2}  & \multicolumn{1}{c}{-}  & \multicolumn{1}{c}{65.0}  & \multicolumn{1}{c}{-}  &  59.8 & \multicolumn{1}{c}{66.0}  & \multicolumn{1}{c}{-}  & \multicolumn{1}{c}{66.2}  & \multicolumn{1}{c}{-}  &  61.0    &\multicolumn{1}{c}{66.6}  & \multicolumn{1}{c}{-}  & \multicolumn{1}{c}{62.9}  & \multicolumn{1}{c}{-}  &  58.3   \\ 

\rowcolor{lightmintbg} \rule{-2pt}{10pt} PNPDet~\cite{pnpdet}                & WACV 2021             & DLA34   &\multicolumn{1}{c}{75.5}  & \multicolumn{1}{c}{-}  & \multicolumn{1}{c}{75.5}  & \multicolumn{1}{c}{-}  &  75.5 & \multicolumn{1}{c}{73.1}  & \multicolumn{1}{c}{-}  & \multicolumn{1}{c}{73.1}  & \multicolumn{1}{c}{-}  &  73.1 &\multicolumn{1}{c}{74.6}  & \multicolumn{1}{c}{-}  & \multicolumn{1}{c}{74.6}  & \multicolumn{1}{c}{-}  &  74.6   \\ 

\end{tabular}
\end{adjustbox}
\begin{adjustbox}{max width=\linewidth}
\begin{tabular}{lll|ccccc|ccccc|ccccc}
\hline \hline
\multirow{2}{*}{\textbf{Method}} & \multirow{2}{*}{\textbf{Publication}} & \multirow{2}{*}{\textbf{Model}} & \multicolumn{5}{c|}{\rule{-2pt}{10pt}\textbf{All set1}} & \multicolumn{5}{c|}{\textbf{All set 2}} & \multicolumn{5}{c}{\textbf{All set 3}}    

\\ \cline{4-18} 
                        &                              &                        & \multicolumn{1}{c}{\rule{-2pt}{10pt}\textbf{K = 1}} & \multicolumn{1}{c}{\textbf{2}} & \multicolumn{1}{c}{\textbf{3}} & \multicolumn{1}{c}{\textbf{5}} & \multicolumn{1}{c|}{\textbf{10}} &\multicolumn{1}{c}{\textbf{1}} & \multicolumn{1}{c}{\textbf{2}} & \multicolumn{1}{c}{\textbf{3}} & \multicolumn{1}{c}{\textbf{5}}& \multicolumn{1}{c|}{\textbf{10}} &\multicolumn{1}{c}{\textbf{1}} & \multicolumn{1}{c}{\textbf{2}} & \multicolumn{1}{c}{\textbf{3}} & \multicolumn{1}{c}{\textbf{5}} & \multicolumn{1}{c}{\textbf{10}} \\ \hline \hline

\rowcolor{lightmintbg}  \rule{-2pt}{10pt} Retentive R-CNN~\cite{retentive_rcnn}                & CVPR 2021             & Faster R-CNN R-101   &\multicolumn{1}{c}{71.3}  & \multicolumn{1}{c}{72.3}  & \multicolumn{1}{c}{72.1}  & \multicolumn{1}{c}{74.0}  & 74.6 & \multicolumn{1}{c}{66.8}  & \multicolumn{1}{c}{68.4}  & \multicolumn{1}{c}{70.2}  & \multicolumn{1}{c}{70.7}  & 71.5 &\multicolumn{1}{c}{69.0}  & \multicolumn{1}{c}{70.9}  & \multicolumn{1}{c}{72.3}  & \multicolumn{1}{c}{73.9}  & 74.1   \\ 

\rule{-2pt}{10pt} CFA~\cite{cfa}  & CVPRW 2022             & Faster R-CNN R-101  \quad \quad &\multicolumn{1}{c}{75.0}  & \multicolumn{1}{c}{76.0}  & \multicolumn{1}{c}{76.8}  & \multicolumn{1}{c}{77.3}  & 77.3 & \multicolumn{1}{c}{70.4}  & \multicolumn{1}{c}{72.7}  & \multicolumn{1}{c}{73.7}  & \multicolumn{1}{c}{74.7}  &  74.2 &\multicolumn{1}{c}{74.7}  & \multicolumn{1}{c}{75.5}  & \multicolumn{1}{c}{75.0}  & \multicolumn{1}{c}{76.2}  &  76.6 \\ 

\rowcolor{lightmintbg}  \rule{-2pt}{10pt} NIFF~\cite{niff}                & CVPR 2023             & Faster R-CNN R-101   &\multicolumn{1}{c}{75.9}  & \multicolumn{1}{c}{76.9}  & \multicolumn{1}{c}{77.3}  & \multicolumn{1}{c}{77.9}  & 77.5 & \multicolumn{1}{c}{70.6}  & \multicolumn{1}{c}{71.6}  & \multicolumn{1}{c}{74.5}  & \multicolumn{1}{c}{75.1}  & 74.5 &\multicolumn{1}{c}{74.7}  & \multicolumn{1}{c}{76.0}  & \multicolumn{1}{c}{76.1}  & \multicolumn{1}{c}{76.8}  &  76.7 \\

\rule{-2pt}{10pt} DiGeo~\cite{digeo}  & CVPR 2023             & Faster R-CNN R-101   &\multicolumn{1}{c}{67.9}  & \multicolumn{1}{c}{70.6}  & \multicolumn{1}{c}{72.4}  & \multicolumn{1}{c}{75.4}  & 76.1 & \multicolumn{1}{c}{67.5}  & \multicolumn{1}{c}{68.4}  & \multicolumn{1}{c}{71.4}  & \multicolumn{1}{c}{71.6}  &  73.6 &\multicolumn{1}{c}{68.6}  & \multicolumn{1}{c}{70.9}  & \multicolumn{1}{c}{72.9}  & \multicolumn{1}{c}{74.4}  & 75.0  \\ 

\rowcolor{lightmintbg}  \rule{-2pt}{10pt} UPPR~\cite{uppr}     & CVPR 2024             & Cascade R-CNN R-101   &\multicolumn{1}{c}{76.1}  & \multicolumn{1}{c}{77.0}  & \multicolumn{1}{c}{77.9}  & \multicolumn{1}{c}{78.2}  & 78.4 & \multicolumn{1}{c}{71.3}  & \multicolumn{1}{c}{73.5}  & \multicolumn{1}{c}{74.4}  & \multicolumn{1}{c}{75.1}  & 75.2 &\multicolumn{1}{c}{75.1}  & \multicolumn{1}{c}{76.9}  & \multicolumn{1}{c}{76.2}  & \multicolumn{1}{c}{77.3}  &  77.5 \\

\hline \hline
\end{tabular}
\end{adjustbox} 
\end{table}

\subsection{Discussion on results of I-FSOD methods} \label{subsec:discussion-IFSOD}
I-FSOD methods mainly evaluate their performance on the MSCOCO dataset \cite{coco} using AP and AR metrics. These benchmarks are typically assessed under 1-shot, 5-shot, and 10-shot settings. Table \ref{Tab:IFSOD-COCO} illustrates the comparative performance of various existing I-FSOD methods on the COCO dataset. Notably, the SC-AWG \cite{sc_awg} method demonstrates superior performance in terms of the AR metric in most settings. Specifically, in the 1-shot setting, SC-AWG achieves the highest AR scores, surpassing the previous best results by 5.6\% to 60\%. Conversely, iFS-RCNN \cite{ifsrcnn} excels in the AP metric, securing the best scores across all settings except for the novel classes in the 10-shot setting. In this particular setting, iFS-RCNN achieves the second-best result, following i-DETR \cite{idetr}. Overall, iFS-RCNN's AP scores surpass the second-best precision scores by a margin of 2.3\% to 9.1\%.
\begin{table}[t!]
\caption{Result comparison between I-FSOD methods on novel, base \& all classes of the COCO dataset with shots \textbf{K}=1, 5, 10. The metrics of comparison are \textbf{AP}, \textbf{AR}.} \label{Tab:IFSOD-COCO}
\vspace{-1em}
\centering
\begin{adjustbox}{max width=\linewidth}
\begin{tabular}{lrl|cccccc|cccccc|cccccc}
\hline \hline
\multirow{3}{*}{\textbf{Method}} & \multirow{3}{*}{\textbf{Publication}} & \multirow{3}{*}{\textbf{Model}} & \multicolumn{6}{c|}{\rule{-2pt}{10pt} \textbf{K = 1}}                                                                                                                       & \multicolumn{6}{c|}{\textbf{K = 5}}                                                                                                                       & \multicolumn{6}{c}{\textbf{K = 10} }                                                                                                                       \\ \cline{4-21} 
                        &                              &                           & \multicolumn{2}{c}{\rule{-2pt}{10pt}\textbf{Novel Classes}}                   & \multicolumn{2}{c}{\textbf{Base Classes}}                     & \multicolumn{2}{c}{\textbf{All Classes}}  & \multicolumn{2}{|c}{\textbf{Novel Classes}}                   & \multicolumn{2}{c}{\textbf{Base Classes}}                     & \multicolumn{2}{c}{\textbf{All Classes}}  & \multicolumn{2}{|c}{\textbf{Novel Classes}}                     & \multicolumn{2}{c}{\textbf{Base Classes}}                     & \multicolumn{2}{c}{\textbf{All Classes}}  \\ \cline{4-21} 
                        &                              &                           & \multicolumn{1}{c}{\rule{-2pt}{10pt}\textbf{AP}}  & \multicolumn{1}{c}{\textbf{AR}}   & \multicolumn{1}{c}{\textbf{AP}}   & \multicolumn{1}{c}{\textbf{AR}}   & \multicolumn{1}{c}{\textbf{AP}}    & \textbf{AR}   & \multicolumn{1}{c}{\textbf{AP}}  & \multicolumn{1}{c}{\textbf{AR}}   & \multicolumn{1}{c}{\textbf{AP}}   & \multicolumn{1}{c}{\textbf{AR}}   & \multicolumn{1}{c}{\textbf{AP}}    & \textbf{AR}   & \multicolumn{1}{c}{\textbf{AP}}    & \multicolumn{1}{c}{\textbf{AR}}   & \multicolumn{1}{c}{\textbf{AP}}   & \multicolumn{1}{c}{\textbf{AR}}   & \multicolumn{1}{c}{\textbf{AP}}    & \textbf{AR}   \\ \hline \hline
 \rowcolor{lightmintbg} \rule{-2pt}{10pt}  Finetuning              & $-$                         &                           & \multicolumn{1}{c}{0.0} & \multicolumn{1}{c}{0.0}  & \multicolumn{1}{c}{1.1}  & \multicolumn{1}{c}{1.8}  & \multicolumn{1}{c}{0.8}   & 1.4  & \multicolumn{1}{c}{0.2} & \multicolumn{1}{c}{3.5}  & \multicolumn{1}{c}{2.6}  & \multicolumn{1}{c}{7.4}  & \multicolumn{1}{c}{2.0}   & 6.4  & \multicolumn{1}{c}{0.6}   & \multicolumn{1}{c}{4.2}  & \multicolumn{1}{c}{2.8}  & \multicolumn{1}{c}{8.0}  & \multicolumn{1}{c}{2.3}   & 7.0  \\ 

\rule{-2pt}{10pt} ONCE~\cite{once}                   & CVPR 2020                    & CenterNet R-50            & \multicolumn{1}{c}{0.7} & \multicolumn{1}{c}{6.3}  & \multicolumn{1}{c}{17.9} & \multicolumn{1}{c}{19.5} & \multicolumn{1}{c}{13.6}  & 16.2 & \multicolumn{1}{c}{1.0} & \multicolumn{1}{c}{7.4}  & \multicolumn{1}{c}{17.9} & \multicolumn{1}{c}{19.5} & \multicolumn{1}{c}{13.7}  & 16.4 & \multicolumn{1}{c}{1.2}   & \multicolumn{1}{c}{7.6}  & \multicolumn{1}{c}{17.9} & \multicolumn{1}{c}{19.5} & \multicolumn{1}{c}{13.7}  & 16.5 \\ 

 \rowcolor{lightmintbg} \rule{-2pt}{10pt} LEAST~\cite{least}             & Arxiv 2021                   & Faster RCNN R-101         & \multicolumn{1}{c}{4.4} & \multicolumn{1}{c}{21.6} & \multicolumn{1}{c}{24.6} & \multicolumn{1}{c}{35.8} & \multicolumn{1}{c}{7.5}   &$-$   & \multicolumn{1}{c}{9.4} & \multicolumn{1}{c}{27.6} & \multicolumn{1}{c}{25.2} & \multicolumn{1}{c}{36.4} & \multicolumn{1}{c}{13.70} & $-$     & \multicolumn{1}{c}{12.5}  & \multicolumn{1}{c}{30.3} & \multicolumn{1}{c}{23.1} & \multicolumn{1}{c}{34.2} & \multicolumn{1}{c}{16.2}  & $-$    \\ 

\rule{-2pt}{10pt} iMFTA~\cite{imfta}                & CVPR 2021                    & Mask RCNN R-50 + FPN      & \multicolumn{1}{c}{3.2} & \multicolumn{1}{c}{$-$ }    & \multicolumn{1}{c}{27.8} & \multicolumn{1}{c}{$-$ }    & \multicolumn{1}{c}{21.7}  & $-$     & \multicolumn{1}{c}{6.1} & \multicolumn{1}{c}{$-$ }    & \multicolumn{1}{c}{24.1} & \multicolumn{1}{c}{$-$ }    & \multicolumn{1}{c}{19.6}  & $-$     & \multicolumn{1}{c}{6.9}   & \multicolumn{1}{c}{$-$ }    & \multicolumn{1}{c}{23.4} & \multicolumn{1}{c}{$-$ }    & \multicolumn{1}{c}{19.3}  & $-$     \\ 
 
 \rowcolor{lightmintbg} \rule{-2pt}{10pt} Sylph~\cite{sylph}                   & CVPR 2022                    & Faster RCNN R-50 + FPN    & \multicolumn{1}{c}{0.9} & \multicolumn{1}{c}{$-$ }    & \multicolumn{1}{c}{17.9} & \multicolumn{1}{c}{$-$ }    & \multicolumn{1}{c}{$-$ }     & $-$     & \multicolumn{1}{c}{1.4} & \multicolumn{1}{c}{$-$ }    & \multicolumn{1}{c}{35.5} & \multicolumn{1}{c}{$-$ }    & \multicolumn{1}{c}{$-$ }     & -    & \multicolumn{1}{c}{1.6}   & \multicolumn{1}{c}{$-$ }    & \multicolumn{1}{c}{35.8} & \multicolumn{1}{c}{$-$ }    & \multicolumn{1}{c}{$-$ }     & $-$     \\ 

\rule{-2pt}{10pt}iFS-RCNN~\cite{ifsrcnn}               & CVPR 2022                    & Mask RCNN R-50 + FPN      & \multicolumn{1}{c}{4.5} & \multicolumn{1}{c}{$-$ }    & \multicolumn{1}{c}{40.1} & \multicolumn{1}{c}{$-$ }    & \multicolumn{1}{c}{31.2} & $-$     & \multicolumn{1}{c}{9.9} & \multicolumn{1}{c}{$-$ }    & \multicolumn{1}{c}{40.1} & \multicolumn{1}{c}{$-$ }    & \multicolumn{1}{c}{32.5} & $-$     & \multicolumn{1}{c}{12.5} & \multicolumn{1}{c}{$-$ }    & \multicolumn{1}{c}{40.0} & \multicolumn{1}{c}{$-$ }    & \multicolumn{1}{c}{33.0} & $-$     \\ 

 \rowcolor{lightmintbg} \rule{-2pt}{10pt} Meta-iFSOD~\cite{meta_ifsod}               & TCSVT 2022                    &CenterNet R-50       & \multicolumn{1}{c}{1.5} & \multicolumn{1}{c}{5.5}    & \multicolumn{1}{c}{30.7} & \multicolumn{1}{c}{27.6}    & \multicolumn{1}{c}{23.4} & 22     & \multicolumn{1}{c}{2.5} & \multicolumn{1}{c}{9.1}    & \multicolumn{1}{c}{33.3} & \multicolumn{1}{c}{29.1}    & \multicolumn{1}{c}{25.6} & 24.1     & \multicolumn{1}{c}{2.6} & \multicolumn{1}{c}{9.6 }    & \multicolumn{1}{c}{31.4} & \multicolumn{1}{c}{27.8 }    & \multicolumn{1}{c}{24.2} & 23.3     \\

\rule{-2pt}{10pt} i-DETR~\cite{idetr}                & AAAI 2023                    & DETR     & \multicolumn{1}{c}{1.9} & \multicolumn{1}{c}{$-$ }    & \multicolumn{1}{c}{29.4} & \multicolumn{1}{c}{$-$ }    & \multicolumn{1}{c}{22.5} & $-$     & \multicolumn{1}{c}{8.3} & \multicolumn{1}{c}{$-$ }    & \multicolumn{1}{c}{30.5} & \multicolumn{1}{c}{$-$ }    & \multicolumn{1}{c}{24.9} & $-$     & \multicolumn{1}{c}{14.4} & \multicolumn{1}{c}{$-$ }    & \multicolumn{1}{c}{27.3} & \multicolumn{1}{c}{$-$ }    & \multicolumn{1}{c}{24.1} & $-$     \\ 
 
 \rowcolor{lightmintbg} \rule{-2pt}{10pt} SC-AWG~\cite{sc_awg}               & CVIU 2023                    & FCOS     & \multicolumn{1}{c}{1.3} & \multicolumn{1}{c}{22.8 }    & \multicolumn{1}{c}{32.2} & \multicolumn{1}{c}{39.3 }    & \multicolumn{1}{c}{24.4} & 35.2     & \multicolumn{1}{c}{1.8} & \multicolumn{1}{c}{25.5 }    & \multicolumn{1}{c}{34.3} & \multicolumn{1}{c}{46 }    & \multicolumn{1}{c}{26.2} & 40.9     & \multicolumn{1}{c}{2.0} & \multicolumn{1}{c}{25.6 }    & \multicolumn{1}{c}{34.7} & \multicolumn{1}{c}{46.4 }    & \multicolumn{1}{c}{26.5} & 41.2     \\

\multirow{2}{*}{iTFA~\cite{ifsod_sft} }              & \multirow{2}{*}{ICRA 2023}                    & Faster RCNN R-50 + FPN      & \multicolumn{1}{c}{3.8} & \multicolumn{1}{c}{- }    & \multicolumn{1}{c}{35.7} & \multicolumn{1}{c}{- }    & \multicolumn{1}{c}{-} & -     & \multicolumn{1}{c}{8.3} & \multicolumn{1}{c}{- }    & \multicolumn{1}{c}{35.5} & \multicolumn{1}{c}{-}    & \multicolumn{1}{c}{-} & -    & \multicolumn{1}{c}{10.2} & \multicolumn{1}{c}{-}    & \multicolumn{1}{c}{35.5} & \multicolumn{1}{c}{- }    & \multicolumn{1}{c}{-} & - 
\\
              &                   & Faster RCNN R-101 + FPN      & \multicolumn{1}{c}{4.3} & \multicolumn{1}{c}{-}    & \multicolumn{1}{c}{37.4} & \multicolumn{1}{c}{-}    & \multicolumn{1}{c}{-} & -     & \multicolumn{1}{c}{9.9} & \multicolumn{1}{c}{-}    & \multicolumn{1}{c}{37.2} & \multicolumn{1}{c}{-}    & \multicolumn{1}{c}{-} & -     & \multicolumn{1}{c}{11.8} & \multicolumn{1}{c}{-}    & \multicolumn{1}{c}{37.2} & \multicolumn{1}{c}{-}    & \multicolumn{1}{c}{-} & - 
\\

 \rowcolor{lightmintbg} \rule{-2pt}{10pt} Tang \textit{et al.} \cite{non-registrable_weights}   & ICONIP 2023                  &   Faster R-CNN R50   & \multicolumn{1}{c}{2.8} & \multicolumn{1}{c}{-}    & \multicolumn{1}{c}{37.3} & \multicolumn{1}{c}{-}    & \multicolumn{1}{c}{28.7} & -     & \multicolumn{1}{c}{7.1} & \multicolumn{1}{c}{-}    & \multicolumn{1}{c}{37.3} & \multicolumn{1}{c}{-}    & \multicolumn{1}{c}{29.8} &   -   & \multicolumn{1}{c}{9.1} & \multicolumn{1}{c}{-}    & \multicolumn{1}{c}{37.3} & \multicolumn{1}{c}{-}    & \multicolumn{1}{c}{30.3} &   -   \\

 \rule{-2pt}{10pt} WS-iFSD \cite{ws_ifsd} & PMLR 2024 &    Base-iFSD  & \multicolumn{1}{c}{2.3} & \multicolumn{1}{c}{-}    & \multicolumn{1}{c}{38.4} & \multicolumn{1}{c}{-}    & \multicolumn{1}{c}{-} &   -   & \multicolumn{1}{c}{3.8} & \multicolumn{1}{c}{-}    & \multicolumn{1}{c}{38.4} & \multicolumn{1}{c}{-}    & \multicolumn{1}{c}{-} &   -   & \multicolumn{1}{c}{4.0} & \multicolumn{1}{c}{-}    & \multicolumn{1}{c}{38.4} & \multicolumn{1}{c}{-}    & \multicolumn{1}{c}{-} &   -   \\
\hline \hline
\end{tabular}
\end{adjustbox}
\end{table}

\subsection{Discussion on results of O-FSOD methods} \label{subsec:discussion-OFSOD}
O-FSOD methods evaluate their models on various datasets, including VOC (single dataset benchmark) and VOC-COCO (cross-dataset benchmark). The chosen metric for assessing the known/closed-set object detection performance is the mAP of known base and novel classes. Recall and average recall values are reported to evaluate the unknown/open-set detection performance. The single dataset benchmark, VOC, is divided into groups of 10, 5, and 5 classes for known base, known novel, and unknown sets, respectively. The results of this single dataset benchmark, referred to as VOC-10-5-5, are shown in Table \ref{Tab:OFSOD}. For the cross-dataset benchmark, VOC-COCO, 20 classes from PASCAL VOC and 20 non-VOC classes from COCO are used as the closed-set training data, while the remaining 40 classes from COCO are assigned as unknown classes. The results for this cross-dataset benchmark, i.e., VOC-COCO, are also presented in Table \ref{Tab:OFSOD}. In both benchmarks, Binyi \textit{et al.} \cite{hsic} outperforms FOOD \cite{food} across all metrics. Specifically, in the VOC-10-5-5 benchmark, they achieve a gain of 2.1-7.5\% in mAP for known base classes, 14.1-31.2\% in mAP for novel base classes, 35.6-37.5\% in recall for unknown classes, and 32.7-38.9\% in average recall for unknown classes. In the VOC-COCO benchmark, the gains are 9.9-17.1\% in mAP for known base classes, 22.1-91.6\% in mAP for novel base classes, 52.6-95.9\% in recall for unknown classes, and 50.7-96.3\% in average recall for unknown classes.

\begin{table*}
\caption{Result comparison between O-FSOD methods on the \textbf{VOC-10-5-5} and \textbf{VOC-COCO} benchmarks with shots \textbf{K}=1, 3, 5, 10. The metrics for comparison are $mAP_k$, $mAP_n$, $R_u$ \& $AR_u$ for mean average precision of known classes, mean average precision of novel classes, recall of unknown classes and average recall of unknown classes, respectively.} \label{Tab:OFSOD}
\vspace{-1em}
\resizebox{\textwidth}{!}{
\fontsize{20}{22}\selectfont
\begin{tabular}{c|ccccc|ccccc|ccccc}
\hline \hline
\multirow{2}{*}{\textbf{Method}} & \multicolumn{5}{c|}{\textbf{mAP\_k/mAP\_n}} &\multicolumn{5}{c|}{\textbf{R\_u}}       & \multicolumn{5}{c|}{\textbf{AR\_u}} \\ \cline{2-16} &   
                        \multicolumn{1}{c|}{k=1} & \multicolumn{1}{c|}{3} & \multicolumn{1}{c|}{5} & \multicolumn{1}{c|}{10} & Mean & \multicolumn{1}{c|}{1} & \multicolumn{1}{c|}{3} & \multicolumn{1}{c|}{5} & \multicolumn{1}{c|}{10} & Mean & \multicolumn{1}{c|}{1} & \multicolumn{1}{c|}{3} & \multicolumn{1}{c|}{5} & \multicolumn{1}{c|}{10} & Mean \\ \hline
\multicolumn{16}{l}{VOC-10-5-5 dataset setting} 
\\ \hline
\rowcolor{lightmintbg} FOOD \cite{g_food, food} & 43.97/8.95  & 48.48/16.83  & 50.18/23.10  & 53.23/28.60  &  48.97/19.37  & 43.72  & 44.52  & 45.65  & 45.84  &  44.93 & 23.51  & 23.58  & 23.61  & 23.86  &  23.64 \\
Binyi \textit{et al.} \cite{hsic} & 45.12/11.56 & 48.90/18.96 & 52.55/27.31 & 57.24/32.63 & 50.95/22.62 & 60.03 & 61.21 & 62.02 & 62.14 & 61.35 & 31.19 & 32.03 & 32.79 & 32.80 & 32.20\\  
\hline
\multicolumn{16}{l}{VOC-COCO dataset setting} 
\\ \hline
\rowcolor{lightmintbg} FOOD \cite{g_food, food} & 15.83/2.26 &   18.08/6.69 & 20.17/9.35 &  23.9/14.47 &  19.5/8.19 & 15.76 & 20.02 & 21.48 & 23.17 &  20.11 & 7.20 & 9.45 & 9.56 & 11.45 & 9.42 \\
Binyi \textit{et al.} \cite{hsic}& 18.54/4.33 & 19.88/11.95 & 22.64/13.82 & 23.71/17.67 & 21.19/11.94 & 30.87 & 32.53 & 32.78 & 35.74 & 32.98 & 14.13 & 15.74 & 16.52 & 17.26 & 15.91\\  
\hline \hline

\end{tabular}
}
\end{table*}

\begin{table}[t!]
\caption{Result comparison between FSDAOD methods on three benchmarks: Cityscapes $\rightarrow$ Foggy Cityscapes, Sim10K $\rightarrow$ Cityscapes, and KITTI $\rightarrow$ Cityscapes. The metric of comparison is \textbf{AP}.}
\label{Tab:FSDAOD}
\vspace{-1em}
\resizebox{\textwidth}{!}{
\footnotesize
\begin{tabular}{lcc|c|c|c}
\hline \hline
Method & Publication & Model & Cityscapes $\rightarrow$ Foggy Cityscapes & Sim10K $\rightarrow$ Cityscapes  & KITTI $\rightarrow$ Cityscapes \\ \hline \hline
\rowcolor{lightmintbg} \rule{-2pt}{10pt} Wang \textit{et al.} \cite{fs_adaptive_frcnn} & CVPR 2019 & Faster R-CNN & 31.3 & 41.2 $\pm$ 0.6 & - \\
\rule{-2pt}{10pt} Nakamura \textit{et al.} \cite{cutmix} & ACCV 2022 & Faster R-CNN VGG16 & - & 63.6 & - \\ 
\rowcolor{lightmintbg} \rule{-2pt}{10pt} PICA \cite{pica} & WACV 2022 & Faster R-CNN VGG16 & 32.2 $\pm$ 0.8 & 42.1 $\pm$ 0.7 & - \\
\rule{-2pt}{10pt} AcroFOD \cite{acrofod} & ECCV 2022 & YOLOv5 & 41.1 $\pm$ 0.8 & 62.5 $\pm$ 1.6 & 62.6 $\pm$ 2.1\\
\rowcolor{lightmintbg} \rule{-2pt}{10pt} AsyFOD \cite{asyfod} & CVPR 2023 & YOLOv5 & 44.3 $\pm$ 1.0 & 65.4 $\pm$ 0.9 & 64.1 $\pm$ 1.1 \\ 
\hline
\end{tabular}
}
\end{table}

\subsection{Discussion on results of FSDAOD methods} \label{subsec:discussion-FSDAOD}
FSDAOD methods have been evaluated on multiple source $\rightarrow$ target pairs, including Cityscapes $\rightarrow$ Foggy Cityscapes, Sim10k $\rightarrow$ Cityscapes and KITTI $\rightarrow$ Cityscapes, using the mAP metric. Table \ref{Tab:FSDAOD} presents a comparison of results from different FSDAOD approaches. It demonstrates that AsyFOD \cite{asyfod} outperforms other methods across all benchmarks. Specifically, it achieves a 7.8\% improvement on Cityscapes $\rightarrow$ Foggy Cityscapes benchmark, a 4.6\% improvement on Sim10K $\rightarrow$ Cityscapes benchmark and a 2.4\% improvement on KITTI $\rightarrow$ Cityscapes benchmark over AcroFOD \cite{acrofod} method. Nakamura \textit{et al.} \cite{cutmix} achieve the second best results in the Sim10k $\rightarrow$ Cityscapes benchmark behind AsyFOD \cite{asyfod} with a mAP score of 63.6. This represents a 1.8\% gain over the 62.5 mAP score achieved by AcroFOD \cite{acrofod}. 

\section{Challenges, Applications and Future Research Directions}
\label{sec:challenges-application-future}
\subsection{Challenges} \label{subsec:challenges}
Few-shot object detection (FSOD) faces numerous challenges. One primary issue is the scarcity of labeled data, which makes it difficult for models to learn effectively. Additionally, retaining knowledge of old classes while learning new ones without forgetting is a significant hurdle. FSOD must also manage domain shifts and environmental variations, which can affect model performance across different contexts. Another challenge is distinguishing between known and unknown classes, ensuring accurate detection without confusion. Moreover, ensuring model scalability and efficiency is crucial for practical applications. Finally, dealing with incomplete annotations adds another layer of complexity to training effective FSOD models. This section will delve into these challenges in detail.
\begin{itemize}[leftmargin=0.5cm,noitemsep,topsep=17pt,
                    before = \tablistcommand,
                    after  = \tablistcommand] 
    \item \textbf{Scarcity of Labeled Data:} 
    Due to limited labeled data, efficiently utilizing available examples while mitigating the risk of overfitting or poor generalization is a primary challenge in FSOD tasks. Given the few annotated instances per class, models can quickly memorize the few training examples, leading to overfitting. This results in poor performance when the model encounters new instances that slightly differ from the training examples. Strategies such as data augmentation, transfer learning, and leveraging synthetic data generation can help create a more diverse training set, improving the model’s generalization ability. Additionally, techniques like meta-learning train models to learn from limited data by focusing on learning how to learn, which enhances their performance on few-shot tasks.
    \item \textbf{Retaining Knowledge of Old Classes:} 
    Balancing the ability to learn new classes while not forgetting previously learned ones is a key challenge in FSOD methods. This dilemma, known as the stability-plasticity dilemma, highlights the need for models to adapt to new information (plasticity) while retaining old knowledge (stability). Strategies such as incremental learning, rehearsal techniques (where a portion of old class data is retained and replayed), and regularization terms can assist in achieving this balance. Nonetheless, ongoing research continues to focus on determining the most effective approach to minimize interference between old and new knowledge.
    \item \textbf{Handling Domain Shifts and Environmental Variations:} 
    Handling domain shifts and variations in environmental conditions poses a significant challenge for FSOD. These shifts and variations demand robustness and adaptability to diverse domains and scenarios for effective FSOD. For instance, a model trained to detect objects in sunny, outdoor scenes may struggle to perform well in indoor or rainy environments. Domain adaptation techniques, such as unsupervised domain adaptation, can help by aligning feature distributions between the source and target domains. Additionally, data augmentation strategies, such as altering lighting conditions, backgrounds, and object appearances during training, can improve robustness. Transfer learning from pre-trained models on large, diverse datasets can also provide more generalized features that better handle domain shifts.
    \item \textbf{Class Imbalance and Semantic Gap:} 
    The class imbalance and semantic gaps are common challenges in object detection tasks. Class imbalance can significantly affect the performance of FSOD models, as they might become biased towards the more common classes, resulting in poor detection performance for the less common ones. Techniques such as re-sampling, synthetic data generation, and specialized loss functions targeting minority classes can help address class imbalances. On the other hand, the semantic gap refers to the difference in understanding between the training data and the target classes the model needs to generalize. This gap can impede the model’s ability to correctly identify and categorize objects not well-represented in the training data. To bridge semantic gaps, leveraging external knowledge sources or pre-trained models can provide a richer semantic understanding, facilitating better generalization to new classes.
    \item \textbf{Model Scalability and Efficiency:} One of the significant challenges in FSOD is ensuring model scalability and efficiency, particularly when dealing with a large number of classes. As the number of classes increases, the model must maintain high performance without a corresponding increase in computational resources. This scalability issue becomes especially pronounced in real-world applications where new object classes are frequently introduced. Efficiently managing this dynamic growth requires innovative approaches, such as hierarchical classification, feature reuse, and modular network designs that can adapt to an expanding set of classes without extensive retraining.
\end{itemize}

\subsection{Applications} \label{subsec:application}
This section outlines the diverse applications of few-shot object detection methods, showcasing their versatility and utility in addressing real-world challenges across various domains.
\begin{itemize}[leftmargin=0.5cm,noitemsep,topsep=17pt,
                    before = \tablistcommand,
                    after  = \tablistcommand]
    \item \textbf{Adaptability to Novel Environments}: 
    FSOD methods excel in adapting to novel environments or scenarios where labeled data is scarce or unavailable. This adaptability is particularly valuable in applications such as autonomous vehicles, environmental monitoring, and border security, where detecting unseen or unexpected objects is critical for safe and effective operation. For instance, in autonomous driving, FSOD systems can quickly learn to recognize new obstacles or traffic signs not present in the training data, enhancing the vehicle’s ability to navigate safely. Similarly, in environmental monitoring, FSOD can detect rare wildlife species or emerging environmental threats with minimal labeled data, facilitating timely and informed decision-making. In border security, FSOD enables the detection of unconventional or novel contraband items, enhancing security measures by identifying potential threats that traditional systems might miss.
    \item \textbf{Efficient Resource Utilization}: FSOD techniques optimize resource utilization by requiring only a limited amount of labeled data for training. This efficiency is advantageous in resource-constrained environments such as medical imaging facilities, where acquiring large amounts of labeled data may be impractical or costly. In medical imaging, FSOD can assist in diagnosing rare diseases by learning from a small number of annotated cases, thus reducing the burden on medical professionals and speeding up the diagnostic process. Similarly, in remote sensing applications, where collecting labeled data can be expensive and time-consuming, FSOD can help identify new land cover types or detect changes in the environment with minimal data.
    \item \textbf{Real-time Object Recognition:} The speed and accuracy of FSOD methods make them suitable for real-time object recognition tasks in applications such as augmented reality (AR), robotics, and surveillance. Their ability to quickly adapt to new objects or environments enables seamless integration into dynamic systems requiring rapid decision-making and response. In AR, FSOD can enhance the user experience by recognizing and interacting with new objects in the user’s environment in real-time. In robotics, FSOD allows robots to operate in unstructured environments, identifying and manipulating new objects without extensive retraining. For surveillance, FSOD provides the capability to detect suspicious activities or objects that were not previously known, thereby improving security and situational awareness in real-time.
    \item \textbf{Cross-Domain Generalization:} FSOD methods, especially FSDAOD-based methods, demonstrate strong generalization capabilities across different domains, enabling their deployment in diverse real-world scenarios. This cross-domain generalization is essential for applications spanning multiple industries, such as agriculture, retail, and logistics, where objects of interest vary widely in appearance and context. In agriculture, FSOD can be used to identify different types of crops, pests, or diseases with minimal labeled examples, improving crop management and yield. In retail, FSOD enables the detection of various products on shelves, even those not seen during training, facilitating inventory management and enhancing customer experience. In logistics, FSOD can help in sorting and recognizing packages or items that change frequently, improving efficiency and accuracy in supply chain operations.
    \item \textbf{Rare Disease Detection:} FSOD is particularly valuable in medical applications, especially for detecting rare diseases and conditions. In many cases, obtaining a large dataset of labeled medical images for rare diseases is challenging due to the infrequency of these conditions. FSOD can train models with only a few annotated examples, enabling the detection of rare abnormalities in medical imaging, such as unusual tumors, rare genetic conditions, or atypical presentations of common diseases. This capability can enhance diagnostic accuracy, aid in early detection, and improve patient outcomes by ensuring that even rare conditions are considered during medical examinations.
    \item \textbf{Industry Quality Control: } FSOD can significantly enhance industrial quality control processes. In manufacturing, products often need to be inspected for defects, which can vary widely and occur infrequently. FSOD allows for the detection of these rare defects with minimal labeled examples, improving the efficiency and accuracy of quality control systems. By quickly identifying defects such as cracks, deformations, or misalignments, FSOD helps maintain high product standards and reduces waste, ensuring that only high-quality products reach the market.
    \item \textbf{Traffic Monitoring and Accident Prevention: } FSOD can greatly enhance traffic safety by enabling the rapid detection of unexpected hazards on the road with minimal labeled data. This includes identifying debris, stray animals, or stalled vehicles that pose risks to drivers. FSOD's ability to quickly learn from a few examples allows traffic monitoring systems to adapt swiftly to new threats, improving real-time responses and accident prevention. Integrating FSOD into traffic surveillance infrastructure can thus significantly reduce accidents and enhance overall traffic management efficiency.
    \item \textbf{Cyber-security: } FSOD can be effectively applied in cyber-security, particularly in detecting novel malware or phishing attempts. Traditional cyber-security systems often rely on extensive databases of known threats to identify and mitigate risks. However, FSOD can enhance these systems by enabling the rapid identification of new and emerging threats with minimal labeled examples. For instance, an FSOD model could be trained to recognize subtle variations in phishing emails or new malware signatures after encountering only a few instances. This capability allows for quicker responses to evolving cyber threats, improving overall security posture by identifying and neutralizing attacks that traditional systems might miss.
    \item \textbf{Remote Sensing: } Few-shot object detection (FSOD) has significant applications in remote sensing \cite{fsod4rsi, retentive_com_remote_sensing, remote_sensing_label_cons}, particularly in addressing the challenges of limited labeled data for training robust models. In the realm of environmental monitoring, FSOD identifies rare and endangered species and detects changes in small-scale ecosystems, even with minimal annotated examples. In agriculture, FSOD can be used to monitor crop health, detect pest infestations, and assess soil conditions using only a limited number of labeled samples, thereby enhancing precision agriculture practices. For disaster management, FSOD can rapidly detect and map affected areas during events like floods, earthquakes, or wildfires, even with scarce pre-labeled disaster imagery. In urban planning and development, FSOD enables the detection of new constructions, illegal developments, and changes in infrastructure using minimal labeled satellite or aerial images. Furthermore, in military and defense applications, FSOD can identify and track rare or newly emerged threats with limited prior data, thereby enhancing surveillance and reconnaissance operations. Overall, FSOD's ability to learn and generalize from few examples makes it a potent tool in the dynamic, data-scarce field of remote sensing.
\end{itemize}

\subsection{Future Research Directions} \label{subsec:future}
Several promising research directions emerge in FSOD, each offering unique opportunities to advance the field. In this subsection, we outline potential avenues for future investigations.
\begin{itemize}[leftmargin=0.5cm,noitemsep,topsep=17pt,
                    before = \tablistcommand,
                    after  = \tablistcommand]
    \item \textbf{Leveraging State-of-The-Art (SOTA) Object Detector Models:} One promising direction is the integration of integrating SOTA object detection models, such as DINO \cite{dino} into few-shot learning frameworks. DINO and similar models have demonstrated exceptional performance in conventional object detection tasks, achieving high accuracy and robustness. By leveraging these models' advanced feature extraction and object recognition capabilities, researchers may gain valuable insights into enhancing the effectiveness and efficiency of few-shot detection. This integration could potentially address the challenges of limited data availability and improve the generalization ability of few-shot learning models in detecting novel objects.
    \item \textbf{Unsupervised Incremental Object Detection:}  Inspired by unsupervised learning paradigms, researchers can delve into developing unsupervised FSOD techniques. Such methods could aim to discover and classify objects in an entirely data-driven manner without relying on explicit class annotations. Utilizing self-supervised or clustering-based strategies, these approaches may alleviate the need for extensive labeled datasets, making them highly applicable in real-world settings.
    \item \textbf{Improving Localization Quality}: In FSOD, improving object localization is essential but frequently neglected. Methods like Faster R-CNN \cite{faster_rcnn} produce strong candidate boxes, but misclassifications can arise from inadequate localization. Variations in the scale and features of candidate boxes can confuse the classifier. Furthermore, the limited samples in FSOD might miss parts of objects, resulting in incomplete localization. These localization challenges substantially affect detection accuracy. Therefore, future research should prioritize enhancing the precision and reliability of object localization to boost FSOD performance.
    \item \textbf{Leveraging Multimodal Data: } Integrating multimodal data, such as combining visual information with text, audio, or other sensor inputs, offers a promising approach to enhancing FSOD. Incorporating additional data types provides models with a more comprehensive understanding of the context and environment, leading to more accurate and robust object detection. For example, pairing images with descriptive text can help models distinguish between visually similar but semantically different objects. Future research should focus on effectively combining and leveraging these diverse data sources. This involves developing fusion techniques and architectures capable of handling and integrating multimodal information seamlessly, thereby improving the overall performance and applicability of FSOD systems.
    \item \textbf{Continuous Learning and Adaptation: } Continuous learning and adaptation in an open-world setting is a significant challenge and a promising research direction for FSOD. New object classes continuously emerge in real-world scenarios, and the environment can change dynamically. Developing FSOD models that can incrementally learn and adapt to these new classes without forgetting previously learned ones (known as catastrophic forgetting) is essential. This involves creating algorithms to efficiently update the model with new data while maintaining its performance on existing classes. Techniques such as lifelong learning, continual learning frameworks, and efficient memory management strategies can be explored to address this challenge, ensuring that FSOD systems remain effective over time as new data becomes available.
    \item \textbf{Harnessing Foundational Model Feature Extractors:} Building upon foundational models like CLIP (Contrastive Language-Image Pretraining) \cite{clip} and GLIP (Grounded Language-Image Pretraining) \cite{glip}, researchers can explore novel techniques to leverage the feature extractors of these models for improved FSOD. These models have demonstrated their efficacy in cross-modal tasks and could potentially enhance feature representations for object recognition.
    \item \textbf{Fusion of Large Language Models (LLMs):} Integrating LLMs like GPT-4 and BERT with FSOD frameworks represents a promising research direction. LLMs have shown remarkable capabilities in understanding and generating human language, which can be leveraged to enhance object detection tasks. For example, LLMs can generate informative descriptions or contextual information that aids in detecting and classifying objects, particularly in scenarios with limited labeled data. This integration could also lead to the creation of multimodal systems that utilize visual and textual data, thereby improving the overall performance and robustness of FSOD systems.
    \item \textbf{Exploring Self-Supervised Learning Techniques: } Self-supervised learning (SSL) has become a potent technique across various machine learning fields, utilizing unlabeled data to acquire valuable representations. In the context of FSOD, SSL can address the issue of limited labeled data by extracting robust feature representations from extensive unlabeled datasets. Future research could explore SSL methods designed explicitly for FSOD, including pretext tasks that are highly effective for object detection. Leveraging the abundance of unlabeled images can enhance the performance and generalization of FSOD models, making them more practical and efficient in real-world applications with limited labeled data.
\end{itemize}

\section{Conclusion} \label{sec:conclusion}
FSOD is a significant breakthrough in computer vision, tackling the challenge of identifying objects with minimal labeled data. This ability is crucial for applications where collecting large annotated datasets is impractical or impossible. Despite progress, FSOD faces several ongoing challenges, including the scarcity of labeled data, retaining knowledge of old classes while learning new ones, handling domain shifts and environmental variations, distinguishing between known and unknown classes, and addressing incomplete annotations and data augmentation issues. Overcoming these challenges is essential for improving the robustness and accuracy of FSOD models.

In this survey, we have conducted a comprehensive exploration of the rapidly evolving landscape of FSOD tasks. Our analysis has provided an in-depth understanding of different few-shot approaches: standard FSOD, generalized FSOD (G-FSOD), incremental FSOD (I-FSOD), open-set FSOD (O-FSOD), and few-shot domain adaptive object detection (FSDAOD). As object detection continues to evolve and faces challenges related to limited data availability, it is crucial to understand the nuances of these different approaches to effectively assess and contrast them with existing methods. Our goal with this survey is to empower researchers with a deep understanding of the state-of-the-art solutions, thereby advancing progress in the dynamic field of FSOD.

\bibliographystyle{ACM-Reference-Format}
\bibliography{reference}

\end{document}